\documentclass[pdflatex,sn-mathphys-num]{sn-jnl}

\usepackage{amsmath,amsfonts}
\usepackage{algorithmic}
\usepackage{algorithm}
\usepackage{array}
\usepackage[caption=false,font=normalsize,labelfont=sf,textfont=sf]{subfig}
\usepackage{textcomp}
\usepackage{stfloats}
\usepackage{url}
\usepackage{verbatim}
\usepackage{graphicx}
\usepackage[nocompress]{cite}

\usepackage{CJKutf8}
\usepackage[utf8]{inputenc} 
\usepackage[T1]{fontenc}
\usepackage{adjustbox}
\usepackage[most]{tcolorbox}
\usepackage[edges]{forest}
\usepackage[framemethod=tikz]{mdframed}
\usepackage{soul}
\usepackage{multirow}           
\usepackage{booktabs}       
\usepackage{amsfonts}       
\usepackage{nicefrac}      
\usepackage{microtype}     
\usepackage{xcolor}        
\usepackage{colortbl}
\usepackage{enumitem}
\usepackage{amssymb}
\usepackage{wrapfig}
\usepackage{courier}
\usepackage{tabularx}
\usepackage{anyfontsize}
\usepackage{makecell}
\usepackage{dsfont}
\usepackage{hyperref}
\usepackage{pifont}
\usepackage{wasysym}
\usepackage{adjustbox}
\usepackage{rotating}
\usepackage{longtable}

\newcommand{\XSolidBrush}{\ding{55}}

\definecolor{hidden-black}{RGB}{0,0,0}
\definecolor{darkgray}{RGB}{100,100,100}
\definecolor{hidden-blue}{RGB}{0,100,200}
\definecolor{CBBlue}{RGB}{207,234,241}
\definecolor{CBGreen}{RGB}{178,223,138}
\definecolor{CBOrange}{RGB}{251,154,153}
\definecolor{Blue}{HTML}{1F77B4}
\definecolor{Orange}{HTML}{FF7F0E}
\definecolor{Green}{HTML}{2CA02C}
\definecolor{Red}{HTML}{D62728}
\definecolor{Purple}{HTML}{9467BD}
\definecolor{Brown}{HTML}{8C564B}

\theoremstyle{thmstyleone}%
\theoremstyle{thmstyletwo}%

\theoremstyle{thmstylethree}%

\begin{document}

\title[Reward Models for LLM Reasoning]{Enhancing Large Language Model Reasoning with Reward Models: An Analytical Survey}


\author[1]{\fnm{Qiyuan} \sur{Liu}}\email{qiyualiu@cityu.edu.hk}
\equalcont{These authors contributed equally to this work.}
\author[2]{\fnm{Hao} \sur{Xu}}\email{xuhao12@lixiang.com}
\equalcont{These authors contributed equally to this work.}
\author[2]{\fnm{Xuhong} \sur{Chen}}\email{chenxuhong@lixiang.com}
\author[2]{\fnm{Wei} \sur{Chen}}\email{chenwei10@lixiang.com}
\author[3]{\fnm{Yee Whye} \sur{Teh}}\email{y.w.teh@stats.ox.ac.uk}
\author*[1]{\fnm{Ning} \sur{Miao}}\email{ningmiao@cityu.edu.hk}

\affil*[1]{\orgdiv{Hong Kong Institute of AI for Science and Department of Data Science}, \orgname{City University of Hong Kong}, \orgaddress{\street{Tat Chee Avenue, Kowloon}, \city{Hong Kong SAR}, \country{China}}}

\affil[2]{\orgname{Li Auto Inc.}, \orgaddress{\street{11 Wenliang Street, Shunyi District}, \city{Beijing}, \postcode{101399}, \country{China}}}

\affil[3]{\orgdiv{Department of Statistics}, \orgname{University of Oxford}, \orgaddress{\street{24-29 St Giles'}, \city{Oxford}, \postcode{OX1 3LB}, \country{United Kingdom}}}

\abstract{Reward models~(RMs) play a critical role in enhancing the reasoning performance of LLMs. 
For example, they can provide training signals to finetune LLMs during reinforcement learning~(RL) and help select the best answer from multiple candidates during inference. 
In this paper, we provide a systematic introduction to RMs, along with a comprehensive survey of their applications in LLM reasoning.
We first review fundamental concepts of RMs, including their architectures, training methodologies, and evaluation techniques. 
Then, we explore their key applications: (1) guiding generation and selecting optimal outputs during LLM inference, (2) facilitating data synthesis and iterative self-improvement for LLMs, and (3) providing training signals in RL-based finetuning. 
Finally, we discuss critical open questions regarding the selection, generalization, evaluation, and enhancement of RMs, based on existing research and our own empirical findings. Our analysis aims to provide actionable insights for the effective deployment and advancement of RMs for LLM reasoning.}




\maketitle

\setcounter{tocdepth}{2}
\tableofcontents
\clearpage

\section{Introduction}

Large Language Models (LLMs) have demonstrated remarkable capabilities, achieving human-level or even superhuman performance in diverse domains~\citep{bubeck2023sparksagi,liu2024surveymedicallargelanguage, zhang2024visionlanguagemodels, gain2025llmtranslation, yue2025surveyquestionanswering}. 
Nevertheless, pretrained LLMs frequently encounter challenges when addressing more complex tasks that require sophisticated multi-step reasoning ability, such as mathematical problem-solving and code generation. 
Prior research has primarily sought reasoning improvements through extending the reasoning trajectory, either by innovative prompting-based techniques~\citep{wei2023cot, yao2023tot} or through fine-tuning on enriched datasets~\citep{lewkowycz2022solvingquantitativereasoningproblems, liang2023mint, ma2024robustvqa, an2024learningmistakes}. However, the limited availability of high-quality reasoning data constrains the effectiveness of these approaches.

Recent advancements in o1-style models~\citep{openai2024openaio1, deepseekai2025deepseekr1} have highlighted the great potential of reinforcement learning (RL) in boosting the reasoning ability of LLMs. 
In verifiable reinforcement learning settings, the source of supervision is often provided the verifiable reward mechanism~(VRM~\citep{lambert2025tulu3}), instead of reference reasoning trajectories. VRMs are typically implemented with rule-based or hard-coded checkers that validate whether a model’s output satisfies a deterministic specification (e.g., unit tests for code generation or ground-truth solutions for math reasoning). 
This mechanism produces clear binary pass/fail judgments or numeric scores, eliminating reliance on subjective human evaluations~\citep{lee2024rlaifvsrlhf}, which can be costly and noisy. 
However, VRMs also have important limitations. They typically rely on problems with accessible ground-truth answers, which limits their applicability to domains where such specifications or answers are unavailable. Moreover, many VRMs provide only outcome-level feedback at the end of a reasoning trajectory, making them too sparse to effectively refine intermediate reasoning steps.

To mitigate the problems of VRMs, learnable reward models~(RMs) can be used to provide denser and more generally applicable feedback.
Specifically, given a prompt and a candidate response, an RM assigns a score that reflects the acceptability or quality of the output. Unlike VRMs, RMs can be applied to novel questions and to domains where reference answers are unavailable, expensive to obtain, or difficult to verify; they can also provide denser supervision, including step-level assessments of intermediate reasoning processes. 

The application of RMs in LLM training dates back to around 2022. Before o1-style reasoning models, reinforcement learning from human feedback (RLHF) \citep{ouyang2022rlhf} was widely used to align LLM outputs with human preferences.
Specifically, RMs are trained on pairs of LLM responses, which are annotated by human labelers, and are then used as reward functions for PPO~\citep{schulman2017ppo,bai2022traininghelpfulharmlessassistant} style LLM fine-tuning. 
In this survey, we will focus on RMs dedicated to improving the reasoning ability of LLMs. 
Compared with conventional RLHF-style reward models, reasoning RMs place greater emphasis on the correctness and plausibility of reasoning paths, as well as on fine-grained credit assignment, which are more important to elicit the strong reasoning power of LLMs.
In practice, they are not only used for RL-based finetuning, but also play important roles in test-time search and data construction. Although RLHF provides an important historical foundation for reward modeling in LLMs, our focus is on how such models can be adapted to provide denser and more correctness-oriented feedback, thereby more effectively enhancing the reasoning capabilities of LLMs.

\noindent\paragraph{Comparison with Existing Surveys}

\noindent As shown in Table \ref{tab:survey_comparison}, existing surveys have covered some aspects of research on RM. For example, surveys on RLHF \citep{kaufmann2025rlhfsurvey} and human preference learning \citep{jiang2024surveyhumanpreferencelearning} discuss the application of RMs in preference alignment. Surveys on LLM reasoning \citep{plaat2025multistepreasoningsurvey, xu2025largereasoningmodelssurvey,li2025system1system2} and test-time scaling \citep{zhang2025surveytesttimescaling, pan2023selfcorrectionsurvey} cover a broad range of reasoning techniques, but they do not focus on the role of RMs and their design, training, evaluation, and limitations. Surveys on LLM-as-a-judge \citep{gu2025surveyllmasajudge, li2024surveyllmsasjudge} emphasize generative judges, while leaving other RM paradigms less systematically discussed. Recent RM-related surveys \citep{zhong2025rmsurvey, wu2025sailingstarssurvey} provide valuable summaries of reward modeling, but they do not comprehensively analyze how different types of RMs are applied to test-time guidance, synthetic data curation, and online RL, nor do they deeply examine the issues of RMs in reasoning scenarios.

\noindent\paragraph{Survey Methodology}

\noindent To clarify the scope of this survey, we conducted a structured literature review of reward models for LLM reasoning. We searched major research repositories and conference venues, including arXiv, Google Scholar, Semantic Scholar, OpenReview, ACL Anthology, and proceedings of major machine learning and NLP conferences, using combinations of keywords. To ensure relevance and scholarly impact, we prioritized representative and influential studies, including highly cited papers, papers published or accepted at major machine learning and NLP venues, and recent works that introduce important methods, benchmarks, empirical findings, or conceptual frameworks with clear value for subsequent research. We included papers that develop, train, evaluate, or apply reward models to improve or assess LLM reasoning, especially in test-time guidance, synthetic data curation, self-improvement, and RL-based post-training. We excluded works that only use rule-based or verifiable rewards without parametric reward models, general RLHF or alignment studies without any reasoning focus, and general reasoning methods that do not involve reward modeling.

\noindent\paragraph{Survey Organization and Key Findings}

\noindent In the survey, we begin by classifying and summarizing the latest developments in RM architectures and training methodologies in Section \ref{sec: Categorization}. Specifically, we consider two major RM families. A \emph{discriminative RM} maps a query, reasoning trace pair to a scalar score, without generating other content. In contrast, a \emph{generative RM} performs reward-aware generation: conditioned on the query and reasoning trace, it produces explicit critiques which can encode the final reward. We also compare outcome reward models (ORMs) and process reward models (PRMs), which provide solution-level and step-level rewards, respectively.

Following this, we investigate three main RM applications in improving LLM reasoning, according to the stage at which they intervene in the LLM reasoning pipeline. (1) At inference time, RMs improve a fixed policy by allocating and guiding test-time computation; this motivates our discussion of test-time guidance in Section \ref{sec:testtimeguidance}. (2) During data construction, RMs serve as scalable quality filters or annotators for model-generated reasoning traces, which motivates our discussion of synthetic data curation and self-iteration in Section \ref{sec:datasynthesis}. (3) During online policy optimization, RMs directly instantiate or shape the reward function used to update the policy, which motivates our discussion of online reinforcement learning in Section \ref{sec:reinforcementlearning}. These three stages: deployment-time inference, offline data construction, and online policy optimization, cover the major ways in which RMs are currently used to enhance LLM reasoning.

Complementing our exploration of these applications, in Section \ref{sec:analysis}, we answer five key questions regarding the selection, generalization, evaluation, and future enhancement of RMs, integrating results from existing literature and our experimental findings. We summarize our key findings below.

\vspace{4pt}
\noindent\textbf{Q1: How to choose from different types of RMs?}

The existing direct comparisons show that in some reasoning scenarios, generative RMs typically outperform discriminative RMs with the same base model, especially in inference-time applications, such as Best-of-N (BoN) or search-based selection. 
PRMs are often more effective than ORMs for test-time selection or search in multi-step reasoning tasks, where intermediate-step errors matter and reliable step-level rewards are available. 
In contrast, the advantage of PRMs is less clear in online RL: outcome-level rewards can outperform PRMs, and PRMs may suffer from severe reward hacking due to their noisier reward signals, which might be a result of insufficient step-level annotations.
Therefore, the choice between PRMs and ORMs should depend on the downstream use case, the availability of step-level supervision, and the evaluation protocol. See Section~\ref{sec:rmcomparison}.

\vspace{4pt}
\noindent\textbf{Q2: How do RMs generalize to out-of-distribution scenarios?}

Many existing RMs, especially discriminative RMs, still struggle to generalize reliably to out-of-distribution (OOD) settings. Their performance can degrade under shifts in the domains or difficulty levels of queries, as well as in the formats of reasoning paths. It remains challenging to build reliable and broadly applicable RMs. See Section~\ref{sec:RMgeneralization}.

\vspace{4pt}
\noindent\textbf{Q3: How do LLMs’ general reasoning capabilities relate to their discriminative performance when prompted as generative RMs?}

For generative RMs, existing evidence suggests a close relationship between their discriminative ability as RMs and the reasoning performance of their base LLMs, especially on math and coding tasks. This suggests that improving the reasoning ability of the base LLM may also improve its effectiveness as a generative RM. Conversely, stronger RMs can support further improvements in LLM reasoning through data generation and online RL. See Section~\ref{sec:q3improve}.

\vspace{4pt}
\noindent\textbf{Q4: Do current RM evaluation metrics reflect their real-world performance?}

Current RM evaluation practices often emphasize discriminative accuracies. While useful, such metrics may NOT align with RMs' downstream performance in settings such as BoN selection, search-guided reasoning, or online RL. This motivates evaluation protocols that are more closely aligned with the target use case, such as BoN scores and search or RL-oriented metrics when applicable. See Section~\ref{sec:q4rmevaluation}.
\vspace{1pt}

\vspace{4pt}
\noindent\textbf{Q5: What are the practical deployment challenges of RMs?}

Although RMs can enhance LLM reasoning, their practical use can introduce additional costs. These costs arise from training data construction and annotation, RM training, reward scoring during inference or RL, and long-term maintenance after deployment. 
The overhead is particularly significant for PRMs and generative RMs: PRMs often require costly process-level data construction, while generative RMs can provide richer feedback but usually incur higher latency. 
Therefore, deploying RMs requires a trade-off among accuracy, latency and maintenance cost. In high-throughput settings, lightweight or cascaded verification pipelines may be more practical. See Section \ref{sec:practical_deployment}.

\begin{table}[t]
\centering
\footnotesize
\setlength{\tabcolsep}{2.2pt}
\renewcommand{\arraystretch}{1.15}
\caption{Comparison with existing surveys.
\checkmark: systematically covered; \LEFTcircle: partially covered; \XSolidBrush: not systematically covered.}
\label{tab:survey_comparison}
\begin{tabularx}{\columnwidth}{@{}>{\raggedright\arraybackslash}X*{7}{>{\centering\arraybackslash}c}@{}}
\toprule
\textbf{Survey}
& \makecell[c]{\textbf{LLM}\\\textbf{Reason.}}
& \makecell[c]{\textbf{RM}\\\textbf{Taxonomy}}
& \makecell[c]{\textbf{RM}\\\textbf{Train./Eval.}}
& \makecell[c]{\textbf{Test-time}\\\textbf{Guid.}}
& \makecell[c]{\textbf{Data}\\\textbf{Synth.}}
& \makecell[c]{\textbf{Online}\\\textbf{RL}}
& \makecell[c]{\textbf{RM-specific}\\\textbf{Issues}} \\
\midrule
\citet{plaat2025multistepreasoningsurvey}
& \checkmark & \XSolidBrush & \XSolidBrush & \LEFTcircle & \XSolidBrush & \XSolidBrush & \XSolidBrush \\

\citet{xu2025largereasoningmodelssurvey}
& \checkmark & \LEFTcircle & \LEFTcircle & \checkmark & \LEFTcircle & \checkmark & \LEFTcircle \\

\citet{li2025system1system2}
& \checkmark & \XSolidBrush & \XSolidBrush & \checkmark & \LEFTcircle & \LEFTcircle & \LEFTcircle \\

\citet{kaufmann2025rlhfsurvey}
& \XSolidBrush & \LEFTcircle & \LEFTcircle & \XSolidBrush & \XSolidBrush & \checkmark & \LEFTcircle \\

\citet{jiang2024surveyhumanpreferencelearning}
& \XSolidBrush & \LEFTcircle & \LEFTcircle & \XSolidBrush & \LEFTcircle & \LEFTcircle & \LEFTcircle \\

\citet{gu2025surveyllmasajudge}
& \LEFTcircle & \LEFTcircle & \LEFTcircle & \LEFTcircle & \LEFTcircle & \XSolidBrush & \LEFTcircle \\

\citet{li2024surveyllmsasjudge}
& \LEFTcircle & \LEFTcircle & \LEFTcircle & \LEFTcircle & \LEFTcircle & \XSolidBrush & \LEFTcircle \\

\citet{zhang2025surveytesttimescaling}
& \checkmark & \LEFTcircle & \XSolidBrush & \checkmark & \XSolidBrush & \XSolidBrush & \LEFTcircle \\

\citet{pan2023selfcorrectionsurvey}
& \LEFTcircle & \XSolidBrush & \XSolidBrush & \checkmark & \XSolidBrush & \XSolidBrush & \LEFTcircle \\

\citet{zhong2025rmsurvey}
& \LEFTcircle & \checkmark & \checkmark & \LEFTcircle & \LEFTcircle & \LEFTcircle & \LEFTcircle \\

\citet{wu2025sailingstarssurvey}
& \LEFTcircle & \checkmark & \checkmark & \checkmark & \LEFTcircle & \checkmark & \LEFTcircle \\

\midrule
\textbf{This survey}
& \checkmark & \checkmark & \checkmark & \checkmark & \checkmark & \checkmark & \checkmark \\
\bottomrule
\end{tabularx}
\end{table}

\begin{figure*}[t]
    \centering
    \includegraphics[width=1.0\textwidth]{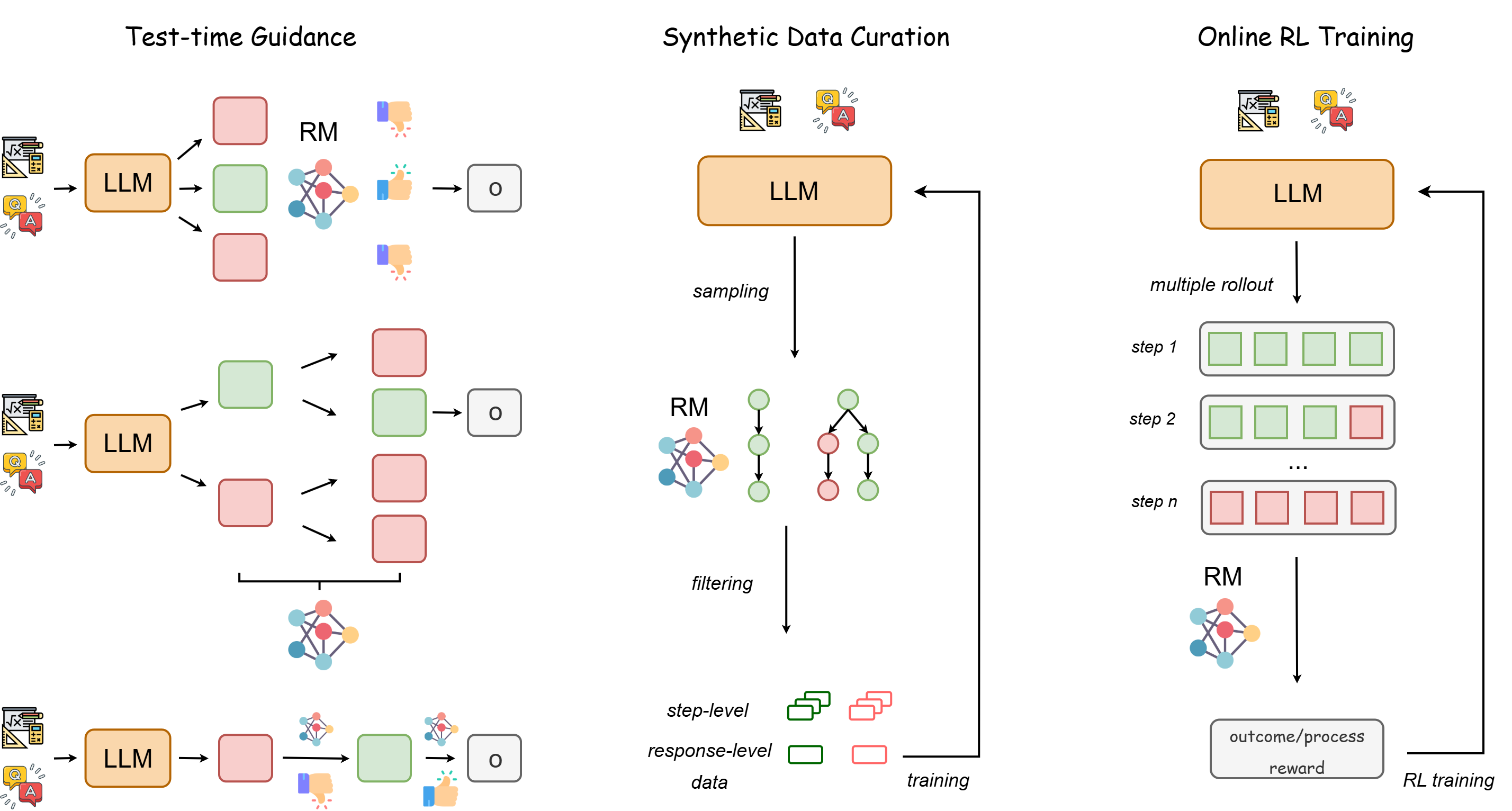}
    \caption{Illustration of three main applications of reward models in LLM reasoning.
    Green/red blocks denote higher/lower-quality candidates or intermediate steps; “o” denotes the final output. \textbf{Left}: test-time guidance. (Top) Sampling and selection: the LLM samples multiple answers and the RM selects the best one. (Middle) Search: a tree of steps is expanded; the RM scores nodes to guide expansion and chooses the terminal candidate. (Bottom) Refinement: failed steps are revised until an acceptable solution is produced. \textbf{Middle}: synthetic data curation. The LLM first samples raw examples step by step; the RM filters them at the response level or step level, and the accepted set is fed back for self-iteration. \textbf{Right}: online RL training. The LLM performs multi-step rollouts where each block denotes generated tokens; the RM supplies outcome or process rewards, based on which the LLM parameters are updated.}
    \label{fig:main}
\end{figure*}

\section{Reward Models: Categorization and Evaluation}
\label{sec: Categorization}

A fundamental challenge in enhancing LLMs is improving their multistep reasoning abilities. Given a problem $p \in \mathcal{P}$, an LLM $\pi$ generates a response with $n$ intermediate reasoning steps $\tau = (\tau_1, \dots, \tau_n)$, denoted $ \tau=\pi(p)$. 
Then a final answer $a \in \mathcal{A}$ can be extracted from $\tau$.
Effective reasoning requires $\tau$ to maintain coherence and logical validity throughout the inference process.

Reward models~(RMs), often referred to as \textit{verifier models}, provide a framework for evaluating $\tau$. Formally, an RM is a parameterized function $R_\theta: \mathcal{X} \rightarrow \mathbb{R}$, where the input $x$ consists of the problem statement $p$, the reasoning steps $\tau$, and other optional contextual information~$c$, such as the reference answer or the external knowledge base.
The output $r=R_\theta(x)$ is typically a scalar reward, which reflects the quality of the reasoning trajectory $\tau$.

In this section, we review reward models from both categorization and evaluation perspectives. We first distinguish RMs by reward granularity, including outcome reward models that assess complete reasoning trajectories and process reward models that provide step-level feedback. We then discuss their reward forms, covering discriminative models that directly output scalar scores and generative models that produce textual judgments or critiques before deriving rewards. We further introduce pointwise and pairwise reward formulations, and finally summarize representative benchmarks for evaluating different types of RMs across text-only and multimodal settings. See Table \ref{tab:representative_rms} and \ref{tab:representative_rms_training} for detailed comparisons of some representative RMs.

\subsection{Reward Granularity}
\label{sec:granularity}

\tikzstyle{my-box}=[
rectangle,
draw=hidden-black,
rounded corners,
text opacity=1,
minimum height=1.5em,
minimum width=5em,
inner sep=2pt,
align=center,
fill opacity=.5,
]
\tikzstyle{leaf}=[
my-box, 
minimum height=1.5em,
fill=CBBlue, 
text=black,
align=left,
font=\large,
inner xsep=5pt,
inner ysep=4pt,
align=left,
text width=45em,
]
\tikzstyle{leaf2}=[
my-box, 
minimum height=1.5em,
fill=purple!27, 
text=black,
align=left,
font=\large,
inner xsep=5pt,
inner ysep=4pt,
]
\tikzstyle{leaf3}=[
my-box, 
minimum height=1.5em,
fill=hidden-blue!57, 
text=black,
align=left,
font=\large,
inner xsep=5pt,
inner ysep=4pt,
]
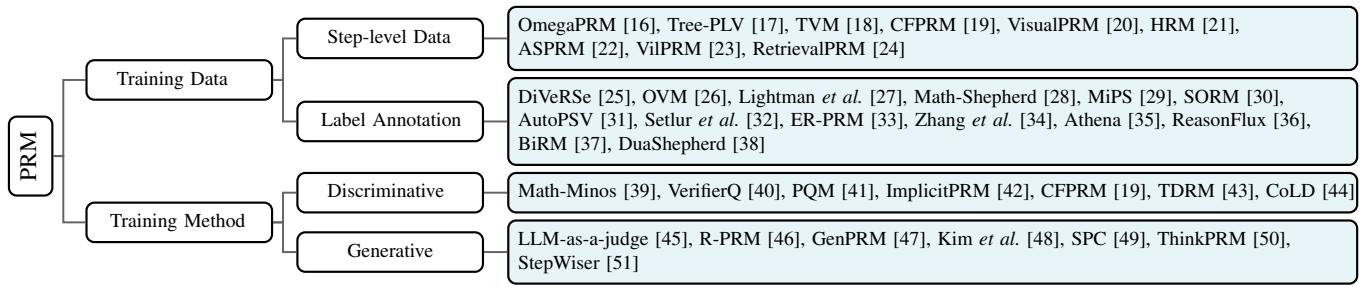
\begin{figure*}[t]
\vspace{-2mm}
\centering
\resizebox{\textwidth}{!}{
	\begin{forest}
		forked edges,
		for tree={
			grow=east,
			reversed=true,
			anchor=base west,
			parent anchor=east,
			child anchor=west,
			base=left,
			font=\large,
			rectangle,
			draw=hidden-black,
			rounded corners,
			align=left,
			minimum width=4em,
			edge+={darkgray, line width=1pt},
			s sep=18pt,
			inner xsep=2pt,
			inner ysep=4pt,
			line width=1.1pt,
			ver/.style={rotate=90, child anchor=north, parent anchor=south, anchor=center},
		},
		where level=1{text width=9em,font=\large,}{},
        where level=2{text width=9em,font=\large,}{},
        where level=3{text width=9.5em,font=\large,}{},
        where level=4{text width=52em,font=\large,}{},
[PRM, ver
		[Data Construction 
			[\ \ Step-level Data
				[OmegaPRM~\citep{luo2024omegaprm}{,}
                Tree-PLV~\citep{he2024treeplv}{,}
                TVM~\citep{lee2025tvm}{,}
                CFPRM~\citep{hu2025cfprm}{,}
                VisualPRM~\citep{wang2025visualprm}{,} \\
                HRM~\citep{wang2025hrm}{,}
                ASPRM~\citep{liu2025asprm}{,}
                VilPRM~\citep{tu2025vilbench}{,}
                RetrievalPRM~\citep{zhu2025retrievalprm}
                ,leaf, text width=42em]
			]
			[Label Annotation
				[DiVeRSe~\citep{li2023diverse}{,}
                OVM~\citep{yu2024ovm}{,}
                Lightman \emph{et al.}~\citep{lightman2023letsverifystepstep}{,} 
                Math-Shepherd~\citep{wang2024mathshepherd}{,}
                MiPS~\citep{wang2024mips}{,} \\
                SORM~\citep{havrilla2024glorewhenwhereimprove}{,} 
                AutoPSV~\citep{lu2024autopsv}{,} 
                Setlur \emph{et al.}~\citep{setlur2024rewardingprogress}{,}
                ER-PRM~\citep{zhang2024erprm}{,} 
                Zhang \emph{et al.}~\citep{zhang2025prmlessons}{,} \\
                Athena~\citep{wang2025athena}{,} 
                ReasonFlux~\citep{zou2025reasonfluxprm}{,} 
                BiRM~\citep{chen2025birm}{,} 
                DuaShepherd~\citep{wu2025duashepherd}
                , leaf, text width=42em]
			]
		]
		[\ Training Method
            [\ \ Discriminative
            [Math-Minos~\citep{gao2024mathminos}{,}
            VerifierQ~\citep{qi2024verifierq}{,}
            PQM~\citep{li2025pqm}{,} 
            ImplicitPRM~\citep{yuan2024implicitprm}{,}
            CFPRM~\citep{hu2025cfprm}{,} \\
            TDRM~\citep{zhang2025tdrm}{,} 
            CoLD~\citep{zheng2025cold}
            , leaf, text width=42em]
            ]
            [\ \ \ Generative
            [LLM-as-a-judge~\citep{zheng2023llmasajudge}{,}
            R-PRM~\citep{she2025rprm}{,}
            GenPRM~\citep{zhao2025genprm}{,} 
            Kim \emph{et al.}~\citep{kim2025reasoningmodelevaluator}{,}
            SPC~\citep{chen2025spc}{,} \\
            ThinkPRM\citep{khalifa2025thinkprm}{,} 
            StepWiser~\citep{xiong2025stepwiser}
            , leaf, text width=42em]
            ]
        ]
	]
]
	\end{forest}
}
\caption{Taxonomy of current research on process reward models}
\label{fig:prmsummary}

\end{figure*}

From the perspective of input granularity, reward models can be categorized into two types: outcome reward models (ORMs), which evaluate the whole response $\tau$, and process reward models (PRMs), which perform separate evaluation for each reasoning step $\tau_i$~\citep{uesato2022solvingmathwordproblems}.

\subsubsection{Outcome Reward Model (ORM)}

An outcome reward model assigns feedback to a complete response rather than to individual intermediate reasoning steps. Given a problem $p$ and a generated reasoning trajectory $\tau=(\tau_1,\ldots,\tau_n)$, an ORM estimates a response-level reward
\[
r = R_\theta(p,\tau),
\]
where $r$ may represent final-answer correctness, human preference, task-specific quality, or a scalar score extracted from a generated judgment. Early verifier studies for mathematical reasoning trained outcome-level models to score complete solutions for reranking or filtering~\citep{cobbe2021trainingverifiers,uesato2022solvingmathwordproblems}. In RLHF, response-level RMs are also commonly trained from preferences over complete answers and then used as reward functions for policy optimization~\citep{ouyang2022rlhf}.

\paragraph{Training methods.}
ORMs can be trained with different objectives depending on the available supervision. First, when each complete trajectory is annotated with a binary outcome label, ORMs are commonly trained as binary classifiers. This setting is typical in verifier-based reasoning tasks, where the label indicates whether the final answer is correct~\citep{cobbe2021trainingverifiers,uesato2022solvingmathwordproblems}. Given a label $\hat{y}\in\{0,1\}$ and a probabilistic reward $r_\theta=R_\theta(p,\tau)\in[0,1]$, the binary cross-entropy objective is
\[
\mathcal{L}_{\mathrm{CE}}
= -\mathbb{E}_{(p,\tau,\hat{y})}
\bigl[
\hat{y}\log r_\theta + (1-\hat{y})\log(1-r_\theta)
\bigr].
\]

Second, when supervision is provided as pairwise or listwise preferences among candidate responses, ORMs can be trained to learn a relative ranking. A standard choice is the Bradley--Terry objective~\citep{bradleyterry,ouyang2022rlhf}, which encourages a preferred response $\tau^+$ to receive a higher reward than a rejected response $\tau^-$:
\[
\mathcal{L}_{\mathrm{BT}}
= -\mathbb{E}_{(p,\tau^+,\tau^-)}
\log \sigma\bigl(R_\theta(p,\tau^+) - R_\theta(p,\tau^-)\bigr).
\]
This objective only constrains relative reward differences and does not require absolute scores. It is widely used in RLHF-style reward modeling and can be extended to reasoning-oriented RMs~\citep{yuan2024eurusrm,liu2025acemath}.

Third, when supervision consists of absolute scalar rating scores or multi-attribute scores, ORMs may be trained with regression-style objectives. Given a target score $\hat{s}$ and a predicted score $s_\theta=R_\theta(p,\tau)$, a typical mean-squared-error objective is
\[
\mathcal{L}_{\mathrm{MSE}}
= \mathbb{E}_{(p,\tau,\hat{s})}
\bigl[
(s_\theta-\hat{s})^2
\bigr].
\]
For multi-attribute annotations, the same objective can be applied to each attribute before aggregating or scalarizing the predicted scores for downstream use. This formulation is adopted in rating-based and multi-objective reward modeling~\citep{wang2023helpsteer,wang2024helpsteer2,wang2024armorm}.

\subsubsection{Process Reward Model (PRM)}
Contrastingly, PRMs perform fine-grained evaluations, assigning rewards to each reasoning step $\tau_i$:
\[
r_i \;=\; R_\theta\bigl(p,\tau_{1:{i-1}}, \tau_i\bigr).
\]

The central advantage of PRMs is dense supervision: instead of judging only whether a final answer is correct, they reveal where a reasoning trajectory starts to succeed or fail. This fine-grained signal, however, makes PRM construction more demanding than ORM construction, because it requires both reliable step-level labels and a clear definition of what constitutes a reasoning step. Therefore, recent PRM research (See Figure \ref{fig:prmsummary}) can be organized around three design choices: constructing step-level data, annotating the meaning of each step, and choosing an appropriate training paradigm.

\paragraph{Training Data construction.} 

The first challenge in PRM data construction is coverage: a PRM must observe not only complete solutions, but also diverse intermediate states where reasoning may branch, succeed, or fail. PRM training data consists of problems, step-by-step solutions, and associated labels. For the step-level data generation, common approaches generate multi-step chain-of-thought (CoT) reasoning via base-model sampling.
Recent advances further target step-level dataset expansion by leveraging reasoning trees, which allow for reusing and analyzing intermediate steps. For example, OmegaPRM~\citep{luo2024omegaprm} maintains an MCTS tree and employs binary search to efficiently locate the first error, while Tree-PLV~\citep{he2024treeplv} similarly constructs reasoning trees to facilitate the collection of preference data.

Moreover, defining an atomic "step" in the response is another challenge in PRM training due to the diverse generation styles of LLMs. 
For instance, TVM~\citep{lee2025tvm} assigns token-level values to accommodate tree search at inference time, whereas CFPRM~\citep{hu2025cfprm} and HRM~\citep{wang2025hrm} merge adjacent steps during data collection and training. ASPRM~\citep{liu2025asprm} introduces an automated partitioning strategy based on model confidence scores.
Furthermore, for multimodal reasoning tasks, methods such as VisualPRM~\citep{wang2025visualprm} and VilPRM~\citep{tu2025vilbench} extend PRMs to vision–language scenarios, enabling step-level reward assignment across both textual and visual modalities. Overall, PRM data construction involves a coverage–granularity trade-off: reasoning trees provide richer supervision, while carefully chosen step units make the resulting labels more consistent.

The next challenge, label annotation, is not only about who provides the label, but also about what the PRM is trained to predict. Regarding the first question, label annotation for PRM training involves both human expert annotations~\citep{lightman2023letsverifystepstep} and various automated labeling methods. 
Early automated efforts relied on semantic similarity~\citep{li2023diverse} or directly utilized outcome labels for intermediate steps~\citep{yu2024ovm}.
As for the second question, a PRM can function either as a value model or as a reward model. The key distinction is whether a step label represents its expected contribution to the final answer, or its local correctness.
Recent methodologies generally fall into two categories, depending on whether their labels reflect the `value' or `correctness' of the corresponding steps.

\textbf{Value-based methods} estimate the probability that a given step will lead to the correct final answer, commonly utilizing Monte Carlo estimation techniques~\citep{wang2024mathshepherd, wang2024mips, havrilla2024glorewhenwhereimprove, wang2025visualprm}. To more accurately predict the value of each step, Zhang \emph{et al.}~\citep{zhang2025prmlessons} make annotations based on the consensus of LLM-as-a-judge and MC estimation. 
Similarly, Athena~\citep{wang2025athena} employs a consensus of weak and strong completers. Other approaches, such as Setlur \emph{et al.}~\citep{setlur2024rewardingprogress} measure step-wise progress using the advantage, and ER-PRM~\citep{zhang2024erprm} computes the step-level values under entropy regularization. Thus, value-based PRMs emphasize downstream potential rather than immediate logical correctness.

\textbf{Correctness-based methods}, conversely, directly evaluate step correctness, typically assigning explicit rewards such as +1 for correct steps and -1 for incorrect steps. The logically correct step may not necessarily have a higher probability of leading to a correct final answer. Notable examples include human-annotated datasets like PRM800K~\citep{lightman2023letsverifystepstep}, and AutoPSV~\citep{lu2024autopsv}, which determine step correctness based on changes in verifier confidence scores.

\textbf{Hybrid methods} have been introduced by recent studies to combine the strengths of both value-based and correctness-based approaches.
For example, to better evaluate the prolonged thinking trajectories in reasoning models such as Deepseek-R1~\citep{deepseekai2025deepseekr1},
ReasonFlux~\citep{zou2025reasonfluxprm} annotates these trajectory-response pairs through a dual-reward system: step-level rewards assess individual steps via their quality, coherence, and alignment, while trajectory-level rewards evaluate the overall strategy through template-guided verification from a policy model \footnote{In this paper, ``policy model'' refers to the model used to generate responses, as opposed to the verifier or reward model.}. 
Meanwhile, BiRM~\citep{chen2025birm} and DuaShepherd~\citep{wu2025duashepherd} employ two different output heads to simultaneously predict step correctness and the potential of each step to achieve a correct final solution. These methods suggest that PRM labels need not be limited to a single scalar meaning. Separating local and global supervision may provide more informative feedback for long-chain reasoning.

\paragraph{Training methods.}
Once step-level data and labels are defined, the remaining question is how much reasoning the PRM should perform before producing a reward. The training procedures for discriminative PRMs, which directly output the reward $r$, and generative PRMs, which reason in natural language before determining the reward $r$, differ substantially and are therefore discussed separately.

Discriminative PRMs are typically trained as binary classifiers using cross-entropy loss. Given the ground truth label $\hat{y}_i\in\{0,1\}$ and the step reward \(r_i \in [0,1]\) for each step $\tau_i$, PRMs can be trained to minimize a similar cross-entropy loss as ORMs:

\[
\mathcal{L}_{\mathrm{PRM}}
= -\,\mathbb{E}_{(p,\tau,\hat{y})}
\Biggl[
  \sum_{i=1}^{n}
    \Bigl(
      \hat{y}_i\,\log r_i
      + (1-\hat{y}_i)\,\log\bigl(1 - r_i\bigr)
    \Bigr)
\Biggr].
\]
Some implementations employ multi-stage training strategies: Math-minos~\citep{gao2024mathminos} incorporates natural language feedback pre-training, while VerifierQ~\citep{qi2024verifierq} utilizes offline Q-learning. PQM~\citep{li2025pqm} optimizes Q-value rankings through comparative loss functions.
TDRM~\citep{zhang2025tdrm} trains PRM via temporal difference learning with cosine reward shaping.
CoLD~\citep{zheng2025cold} corrects the pervasive length bias in PRMs by adding an explicit length penalty, learning a bias estimator, and jointly training to enforce length-invariant rewards.
Notably, ImplicitPRM~\citep{yuan2024implicitprm} demonstrates that by parameterizing outcome rewards as log-likelihood ratios, an ORM trained on response-level labels implicitly learns process-level rewards without requiring step-by-step annotations.
Overall, discriminative PRMs are attractive when inference efficiency is important, but recent work mainly focuses on making their scalar rewards less biased, better ranked, or less dependent on costly step-level annotations.

Generative PRMs typically incorporate detailed reasoning processes to analyze intermediate steps, framing the training task as generation rather than classification. Some approaches, such as LLM-as-a-judge~\citep{zheng2023llmasajudge, kim2025reasoningmodelevaluator}, leverage off-the-shelf models without additional training. 
Other generative PRMs employ specialized fine-tuning data or training frameworks.
For example, R-PRM~\citep{she2025rprm} uses natural language judgments from a stronger LLM for further SFT or DPO~\citep{rafailov2024dpo}.
ThinkPRM~\citep{khalifa2025thinkprm} fine-tunes on long CoT reasoning data. GenPRM~\citep{zhao2025genprm} further enhances accuracy by incorporating code verification data and executing the code at test time. SPC~\citep{chen2025spc} adopts a self-play framework, iteratively improving a critic model through an adversarial game with a generator.
StepWiser~\citep{xiong2025stepwiser} is trained via reinforcement learning using reward labels from Monte-Carlo rollouts that estimate Q-values and score each chunk by the relative change in success rate. Thus, generative PRMs are most useful when interpretability and fine-grained diagnostic feedback are important, although they typically introduce higher training or inference cost.

\begin{table}[t]
\centering
\caption{Representative reward models: basic model information comparisons. N.R. means not reported.}
\label{tab:representative_rms}
\footnotesize
\setlength{\tabcolsep}{3pt}
\renewcommand{\arraystretch}{0.98}
\begin{tabular}{@{}>{\raggedright\arraybackslash}p{0.245\linewidth}>{\raggedright\arraybackslash}p{0.10\linewidth}>{\raggedright\arraybackslash}p{0.220\linewidth}>{\raggedright\arraybackslash}p{0.130\linewidth}>{\raggedright\arraybackslash}p{0.105\linewidth}>{\raggedright\arraybackslash}p{0.105\linewidth}@{}}
\toprule
RM & Year & Domain & Reward output & Scoring form & Open access \\
\midrule
\multicolumn{6}{l}{\textit{Text-only outcome reward models}} \\
\midrule
Eurus-RM~\citep{yuan2024eurusrm} & 2024 & reasoning & Discriminative & Pointwise & Yes \\
ArmoRM~\citep{wang2024armorm} & 2024 & general & Discriminative & Pointwise & Yes \\
HelpSteer RMs~\citep{wang2023helpsteer,wang2024helpsteer2,wang2025helpsteer3} & 2023--25 & chat & Discriminative & Pointwise & Yes \\
Generative Verifiers~\citep{zhang2024genrmcot} & 2025 & math, algorithms & Generative & Pointwise & Data \\
PairJudge RM~\citep{liu2025pairjudgerm} & 2025 & math & Generative & Pairwise & Yes \\
DeepSeek-GRM~\citep{liu2025deepseekgrm} & 2025 & general & Generative & Pointwise & Yes \\
AceMath RM~\citep{liu2025acemath} & 2025 & math & Discriminative & Pointwise & Yes \\
RRM~\citep{guo2025rrm} & 2025 & general & Generative & Pairwise & Yes \\
Skywork-Reward~\citep{liu2024skyworkreward,liu2026skyworkrewardv2} & 2024-25 & general & Discriminative & Pointwise & Yes \\
Rubric-RM~\citep{liu2026openrubrics} & 2025 & general, biomedical & Generative & Pairwise & Yes \\
AgentV-RL~\citep{zhang2026agentvrl} & 2026 & agentic verification & Generative & Pointwise & Code \\
RationaleRM~\citep{wang2026outcomeaccuracy} & 2026 & general, code, writing & Generative & Pairwise & Yes \\
\midrule
\multicolumn{6}{l}{\textit{Text-only process reward models}} \\
\midrule
PRM800K~\citep{lightman2023letsverifystepstep} & 2023 & math & Discriminative & Pointwise & Data \\
OmegaPRM~\citep{luo2024omegaprm} & 2024 & math & Discriminative & Pointwise & N.R. \\
AutoPSV~\citep{lu2024autopsv} & 2024 & math, commonsense & Discriminative & Pointwise & Code \\
ImplicitPRM~\citep{yuan2024implicitprm} & 2024 & math & Discriminative & Pointwise & Yes \\
ThinkPRM~\citep{khalifa2025thinkprm} & 2025 & math & Generative & Pointwise & Yes \\
StepWiser~\citep{xiong2025stepwiser} & 2025 & math & Generative & Pointwise & N.R. \\
ReasonFlux-PRM~\citep{zou2025reasonfluxprm} & 2025 & math, science & Discriminative & Pointwise & Yes \\
AgentPRM~\citep{xi2025agentprm} & 2025 & LLM agents & Discriminative & Pointwise & N.R. \\
\midrule
\multicolumn{6}{l}{\textit{Multimodal outcome reward models}} \\
\midrule
ImageReward~\citep{xu2023imagereward} & 2023 & text-to-image & Discriminative & Pointwise & Yes \\
PickScore~\citep{kirstain2023pickscore} & 2023 & text-to-image & Discriminative & Pointwise & Yes \\
UnifiedReward-Think~\citep{wang2025unifiedrewardthink} & 2025 & image/video reward & Generative & Pairwise & Yes \\
R1-Reward~\citep{zhang2025r1reward} & 2025 & vision-language & Generative & Pairwise & Yes \\
VideoScore2~\citep{he2025videoscore2} & 2025 & text-to-video & Generative & Pointwise & Yes \\
Omni-Reward~\citep{jin2025omnireward} & 2025 & omni-modal & Both & Pairwise & Yes \\
EditReward~\citep{wu2026editreward} & 2026 & image editing & Discriminative & Pointwise & Yes \\
\midrule
\multicolumn{6}{l}{\textit{Multimodal process reward models}} \\
\midrule
VisualPRM~\citep{wang2025visualprm} & 2025 & visual reasoning & Discriminative & Pointwise & Yes \\
ViLPRM~\citep{tu2025vilbench} & 2025 & vision-language reasoning & Discriminative & Pointwise & Code \\
Athena-PRM~\citep{wang2025athena} & 2025 & multimodal reasoning & Discriminative & Pointwise & Code \\
\bottomrule
\end{tabular}
\end{table}

\begin{table}[!htbp]

\centering

\caption{Representative reward models: training backbones, data, methods, and platforms. N.R. means not reported.}

\label{tab:representative_rms_training}

\tiny

\setlength{\tabcolsep}{1pt}

\renewcommand{\arraystretch}{0.95}

\begin{tabular}{@{}>{\raggedright\arraybackslash}p{0.160\linewidth}>{\raggedright\arraybackslash}p{0.145\linewidth}>{\raggedright\arraybackslash}p{0.365\linewidth}>{\raggedright\arraybackslash}p{0.160\linewidth}>{\raggedright\arraybackslash}p{0.130\linewidth}@{}}

\toprule

RM & Backbone & Training data & Training method & Training platform \\

\midrule

\multicolumn{5}{l}{\textit{Text-only outcome reward models}} \\

\midrule

Eurus-RM~\citep{yuan2024eurusrm} & Mistral-7B & UltraInteract; UltraFeedback; UltraSafety & pairwise ranking & N.R. \\

ArmoRM~\citep{wang2024armorm} & Llama-3-8B & HelpSteer; UltraFeedback; BeaverTails; SHP; HH-RLHF; PKU-SafeRLHF & multi-objective regression + MoE & HuggingFace Transformers; Scikit-learn \\

HelpSteer RMs~\citep{wang2023helpsteer,wang2024helpsteer2,wang2025helpsteer3} & Llama/Nemotron & HelpSteer; HelpSteer2; HelpSteer2-Preference; HelpSteer3-Preference & regression or pairwise ranking & NeMo-Aligner \\

Generative Verifiers~\citep{zhang2024genrmcot} & Gemma family & GSM8K; synthetic rationales; LastLetterConcat; BBH WordSorting & generative SFT & N.R. \\

PairJudge RM~\citep{liu2025pairjudgerm} & Qwen2.5-7B-Instruct & PairJudge-432K (NuminaMath-CoT solutions annotated by Gemini-1.5-Flash) & generative SFT & N.R. \\

DeepSeek-GRM~\citep{liu2025deepseekgrm} & Gemma-2-27B & in-house data; MATH; UltraFeedback; OffsetBias; Skywork-Reward-Preference; HelpSteer2-Preference & rejective fine-tuning + GRPO & Fire-Flyer \\

AceMath RM~\citep{liu2025acemath} & Qwen2.5-Math & NuminaMath; OrcaMathWordProblems; MathInstruct; MetaMathQA; AceMath samples & listwise ranking & PyTorch \\

Reward Reasoning Model~\citep{guo2025rrm} & DeepSeek-R1-Distill-Qwen & Skywork-Reward; Tulu 3; WebInstruct-verified; Skywork-OR1; Big-Math-RL; DAPO-Math & GRPO & verl \\

Skywork-Reward~\citep{liu2024skyworkreward, liu2026skyworkrewardv2} & Llama-3.1/Qwen3 & SynPref-40M, curated into 26M preference pairs & pairwise ranking & DeepSpeed ZeRO-1 \\

Rubric-RM~\citep{liu2026openrubrics} & Qwen3-4B/8B & OpenRubrics (from UltraFeedback, Magpie, Skywork-Preference, Synthetic-IF, MegaScience, Medical-o1) & generative SFT  & LLaMA-Factory \\

AgentV-RL~\citep{zhang2026agentvrl} & Qwen3-4B & Polaris; DeepScaleR-40K; AReaL-boba-106K; synthetic verifier trajectories & rejection fine-tuning + GRPO & verl \\

RationaleRM~\citep{wang2026outcomeaccuracy} & Qwen3-14B/30B-A3B & HelpSteer3 rationales converted to atomic checklists & GRPO & N.R. \\

\midrule

\multicolumn{5}{l}{\textit{Text-only process reward models}} \\

\midrule

Lightman et al.~\citep{lightman2023letsverifystepstep} & GPT-4 base & PRM800K (MATH solutions with human step labels) & step classification & N.R. \\

OmegaPRM~\citep{luo2024omegaprm} & Gemini Pro/Gemma-2-27B & MATH-derived process supervision with OmegaPRM MCTS annotation & Step-value regression & N.R. \\

AutoPSV~\citep{lu2024autopsv} & Mistral-7B/Phi-2 & GSM8K; MATH; HellaSwag; Winogrande; ANLI; WizardLM prompts & Step-value regression & PyTorch; HuggingFace Transformers \\

ImplicitPRM~\citep{yuan2024implicitprm} & Llama-3.1-8B-Instruct & UltraInteract math; optional UltraFeedback and UltraInteract code & DPO/KTO /NCA/CE & OpenRLHF; DeepSpeed \\

ThinkPRM~\citep{khalifa2025thinkprm} & R1-Distill-Qwen & PRM800K; MATH prefixes; QwQ-32B-Preview synthetic verification chains & Generative SFT & N.R. \\

StepWiser~\citep{xiong2025stepwiser} & Qwen2.5-7B-Instruct & NuminaMath-CoT; Math-Verify labels; Llama-3.1-70B step segmentation & GRPO & Axolotl; verl \\

ReasonFlux-PRM~\citep{zou2025reasonfluxprm} & Qwen2.5-7B-Instruct & OpenThoughts-114K (from NuminaMath-CoT); Code Contests; camel-ai chemistry & Joint step/trajectory regression & verl \\

AgentPRM~\citep{xi2025agentprm} & Qwen2.5/Llama-3.1 & AgentGym trajectories from WebShop; BabyAI; TextCraft & Q-value and advantage regression & AgentGym \\

\midrule

\multicolumn{5}{l}{\textit{Multimodal outcome reward models}} \\

\midrule

ImageReward~\citep{xu2023imagereward} & BLIP & ImageRewardDB (137K expert comparisons) & pairwise ranking & PyTorch \\

PickScore~\citep{kirstain2023pickscore} & CLIP-H & Pick-a-Pic user preference data & pairwise preference learning & DeepSpeed \\

UnifiedReward-Think~\citep{wang2025unifiedrewardthink} & UnifiedReward & HPD; Open-Image-Preferences; EvalMuse; VideoDPO; LLaVA-Critic; ShareGPTVideo-DPO & SFT + GRPO & LLaMA-Factory; EasyR1 \\

R1-Reward~\citep{zhang2025r1reward} & Qwen2.5-VL-7B-Instruct & R1-Reward-200K (from RLAIF-V, VL-Feedback, POVID, WildVision-Battle, MM-RLHF) & SFT + StableReinforce & OpenRLHF \\

VideoScore2~\citep{he2025videoscore2} & Qwen2.5-VL-7B  & VideoFeedback2 (from VidProM and Koala-36M) & SFT + GRPO & LLaMA-Factory; Video-R1 \\

Omni-Reward~\citep{jin2025omnireward} & MiniCPM-o-2.6/Qwen2.5-VL & Omni-RewardData (from Skywork-Reward-Preference, RLAIF-V, OmniAlign-V-DPO, HPDv2, EvalMuse, VideoDPO) & pairwise ranking + GRPO & LLaMA-Factory; EasyR1 \\

EditReward~\citep{wu2026editreward} & Qwen2.5-VL-7B/MiMo-VL-7B & EditReward-Data (200K expert human preference pairs) & uncertainty-aware pairwise ranking with tie loss & N.R. \\

\midrule

\multicolumn{5}{l}{\textit{Multimodal process reward models}} \\

\midrule

VisualPRM~\citep{wang2025visualprm} & InternVL2.5-8B & VisualPRM400K (from MMPR v1.1 and InternVL2.5 sampled solutions) & step classification & N.R. \\

ViLPRM~\citep{tu2025vilbench} & Qwen2.5-VL-3B & ViLReward-73K (from MAVIS-Geometry, A-OKVQA, GeoQA170K, CLEVR-Math, ScienceQA) & step-value regression & HuggingFace Transformers; DeepSpeed \\

Athena-PRM~\citep{wang2025athena} & Qwen2.5-VL-7B & Athena-5K (from MathV360K; UniGeo; Geometry3K; CLEVR-Math; ScienceQA; DocVQA; GSM8K; MATH; NuminaMath) & step classification & LLaMA-Factory; DeepSpeed ZeRO-2 \\


\bottomrule

\end{tabular}

\end{table}

\subsection{Form of Rewards}

The final rewards from an RM need to be numerical for efficient use in downstream tasks.  
However, RMs may generate intermediate natural language outputs before producing the final reward value. These additional texts may include rubrics~\citep{liu2025deepseekgrm, chen2025rmr1, cook2024tick}, detailed verification~\citep{zhang2024genrmcot, ankner2024cloud} or factuality~\citep{chen2025learningreasonfactuality}, which can encode the final reward. 
Accordingly, RMs can be categorized into discriminative and generative RMs based on their reward generation paradigm. Scalar RMs directly output a numerical value, while generative RMs additionally provide textual critiques.

\subsubsection{Discriminative RMs}

Discriminative RMs~\citep{cai2024internlm2, yuan2024eurusrm, wang2024armorm}
refer to reward models that output only scalar values. Discriminative RMs can be further divided into explicit and implicit RMs. 
\textbf{Explicit RMs}~\citep{cobbe2021trainingverifiers, cai2024internlm2, yuan2024eurusrm, wang2024armorm, jiang2023pairrm, wang2025nemotronreward} are generally implemented by replacing the token-prediction head of an LLM with a linear head, thereby enabling the model to produce a scalar reward directly. 
Differently, \textbf{implicit RMs} such as~\citep{rafailov2024dpo, ethayarajh2024kto, chen2025dice} bypass supervised reward labeling and instead derive a reward signal directly from the model’s likelihood ratios before and after optimization.
For example, DPO itself induces an implicit RM given by
\[
r(p,\tau_{1:{i}}) = \beta \log \frac{\pi_\theta(\tau_{i}\mid p, \tau_{<i})}{\pi_{\mathrm{ref}}(\tau_{i}\mid p, \tau_{<i})},
\]
where \(\beta\) is a scaling constant and \(\pi_{\mathrm{ref}}\) and \(\pi_\theta\) denote the reference and optimized policies, respectively.

\subsubsection{Generative RMs}

Generative RMs output rewards solely in textual form, with the final scores extracted from the generated text. LLM-as-a-judge~\citep{zheng2023llmasajudge} is the most common generative RM, capable of adapting to a wide range of evaluation tasks. They can be further enhanced on additional multiple-domain data to specialize in evaluating LLM responses~\citep{li2024autoj, kim2024prometheus, vu2024flame, cao2024compassjudger1, ye2025conj, alexandru2025atlaselenemini}. 
Zhao \emph{et al.}~\citep{zhao2025onetokenfoolllmasajudge} reduce the false-positive rate via a simple, effective data-augmentation strategy. Furthermore, large reasoning models~\citep{chen2025judgelrm} are employed to produce long CoT for deeper reasoning. 
UnifiedReward-Think~\citep{wang2025unifiedrewardthink} trains a multimodal reasoning model via reinforcement learning across vision tasks. 

\textbf{Generative RMs with scalar outputs.} A special type of generative RMs is generative RMs with scalar outputs.
They generate intermediate texts as critiques alongside a final scalar output, harnessing the language generation abilities of LLMs to support reward justification.
This design serves as an intermediate paradigm between discriminative and generative RMs.
Compared with discriminative RMs whose reasoning remains implicit, the explicit critique can improve interpretability and may confer additional robustness.
For instance, GenRM~\citep{zhang2024genrmcot} first produces CoT-based reasoning to verify math answers step-wise and then computes the token probabilities for keywords (e.g., Yes/No) to extract the reward.
Mahan \emph{et al.}~\citep{mahan2024genrm} similarly trains a GenRM with CoT but uses majority voting to select a superior response from two candidates. CLoud~\citep{ankner2024cloud} features both a language modeling head that generates critiques and a reward head that outputs a scalar score.

\textbf{Rubric-based generative RMs.} A recent trend in generative reward modeling is to make the evaluation criteria explicit before producing the final reward judgment. In addition to a prompt and a candidate response, a rubric-based RM takes a set of criteria as an additional input, or first constructs such criteria as intermediate artifacts. These criteria may be written by humans or users, generated by another model, or generated by the RM itself. The RM then produces critiques or judgments according to these criteria, either criterion by criterion or in a single evaluation pass, and aggregates the resulting judgments into a final reward. Since many tasks do not come with explicit evaluation standards, the key value of rubric-based RMs is to make implicit standards explicit, thereby providing reward signals that better match human expectations.

This design is particularly useful for open-ended tasks, where response quality cannot be measured by a single fixed standard. For example, the same answer may need to be evaluated according to correctness, usefulness, faithfulness, conciseness, safety, or user preference, depending on the task. Making these criteria explicit improves interpretability and makes the reward mechanism easier to audit. However, this paradigm also introduces new failure modes. If the generated rubric misses an important requirement or includes irrelevant evaluation dimensions, the final reward may still be misleading even when the subsequent feedback is fluent \citep{zhang2026rubricbench, ma2026personalizedrewardbench}. Recent work on reward hacking in rubric-based RL further shows that policies may exploit incomplete rubrics or imperfect verifier judgments, so that higher rubric scores do not necessarily indicate better overall quality \citep{mahmoud2026rewardhackingrubric}. Therefore, the quality of rubric generation, selection, and verification has become an important research direction.

Existing rubric-based RM studies mainly differ in how the evaluation criteria are obtained. One line of work relies on human-written or user-provided criteria. For example, Prometheus 2 \citep{kim2024prometheus} supports both direct assessment and pairwise ranking under user-defined evaluation criteria, making the evaluation process more controllable. QA-LIGN \citep{Dineen2025qalign} follows a related idea, decomposing constitutional principles into principle-specific natural-language QA checks and aggregating the resulting judgments as interpretable reward components. RewardAnything \citep{yu2025rewardanything} trains RMs to follow natural-language principles specified at inference time, thereby improving task adaptation without retraining the RM for each new criterion.

Another line of work automatically derives criteria from the given instruction. TICK \citep{cook2024tick} decomposes an instruction into a checklist of yes/no questions, so that the judge can evaluate whether the response satisfies each requirement. Other methods train the RM to generate criteria by itself. DeepSeek-GRM \citep{liu2025deepseekgrm} adopts Self-Principled Critique Tuning (SPCT), encouraging the generative RM to produce adaptive principles before writing critiques. RM-R1 \citep{chen2025rmr1} follows a similar reasoning-oriented direction and applies self-generated rubrics during inference, making reward modeling a multi-step reasoning process rather than a direct score prediction.

Finally, some studies focus on constructing reusable rubric libraries instead of generating criteria from scratch for every evaluation. For example, Reinforcement Learning with Rubric Anchors \citep{huang2025reinforcementlearningrubricanchors} builds a large-scale rubric reward system whose rubrics come from humans, LLMs, or human--LLM collaboration. OpenRubrics \citep{liu2026openrubrics} and RubricHub \citep{li2026rubrichub} further study scalable synthetic rubric generation and large-scale rubric dataset construction. These works shift part of the research focus from only judging responses to also building, filtering, and maintaining high-quality rubrics.

\subsection{Pointwise vs. Pairwise Rewards}
\label{sec:pointwise_vs_pairwise}

RMs may also be classified according to their output format as pointwise or pairwise RMs~\citep{liu2025deepseekgrm}, depending on whether the model assigns independent scores to individual reasoning trajectories or computes relative preferences between multiple trajectories.

\textbf{Pointwise RMs}~\citep{yuan2024eurusrm, wang2024armorm} assign independent quality scores to each response. Given a prompt $p \in \mathcal{P}$ and a candidate response $\tau$, the scoring function is:

$$R_\theta^{\text{point}}\left(p, \tau\right) = r.$$

\textbf{Pairwise RMs}~\citep{jiang2023pairrm, liu2025pairjudgerm} compare two responses and output a preferred candidate. The selection function is:

$$R_\theta^{\text{pair}}\left(p, \tau_1, \tau_2\right) = \tau^*,$$
where $\tau^*$ denotes the higher-quality response according to the RM's learned preference criteria.
The pairwise paradigm can be used to construct a ranking via repeated pairwise comparisons.
Overall, pointwise RMs are more suitable for scenarios requiring an independent and specific score for each candidate, while pairwise RMs are better suited to preference comparison and ranking construction when relative quality differences are more important than absolute scores.

\subsection{Evaluations of Reward Models}
\label{sec:evaluation}

\newcolumntype{L}[1]{>{\raggedright\arraybackslash}p{#1}}
\newcolumntype{Y}{>{\raggedright\arraybackslash}X}

\begin{table}[t]
\footnotesize
\centering
\caption{Benchmarks for different types of reward models}
\label{tab:rm_benchmarks}
\setlength{\tabcolsep}{4pt}
\renewcommand{\arraystretch}{1.8}
\begin{tabularx}{\linewidth}{@{} L{0.18\linewidth} L{0.27\linewidth} Y @{}}
\toprule
\textbf{Category} & \textbf{Evaluation Focus} & \textbf{Benchmarks} \\
\midrule
\multirow[t]{4}{=}{\textbf{Text-only ORM}} & General Preference and Judge Evaluation & 
      MT-Bench~\citep{zheng2023llmasajudge},
      RewardBench~\citep{lambert2024rewardbench},
      RewardBench 2~\citep{malik2025rewardbench2},
      RMB~\citep{zhou2025rmb},
      PPE~\citep{frick2024ppe},
      JudgeBench~\citep{tan2025judgebench}\\
& Domain-Specific Judge Evaluation & 
      RAG-RewardBench~\citep{jin2024ragrewardbench},
      AceMath-RewardBench~\citep{liu2025acemath},
      M-RewardBench~\citep{gureja2024mrewardbench},
      LCB-RB~\citep{fan2025posteriorgrpo},
      Libra Bench~\citep{zhou2025libra}\\
& Context and Criteria Conditioned Evaluation & 
      RABench~\citep{yu2025rewardanything},
      IF-RewardBench~\citep{wen2026ifrewardbench},
      RubricBench~\citep{zhang2026rubricbench},
      Personalized RewardBench~\citep{ma2026personalizedrewardbench},
      Long-RewardBench~\citep{tang2025longrm},
      Long-form RewardBench~\citep{huang2026longformrewardbench}
\\
& Robustness and Bias Diagnostics & 
      RM-Bench~\citep{liu2024rmbench},
      reWordBench~\citep{wu2025rewordbench}\\
\addlinespace[0.35em]
\multirow[t]{4}{=}{\textbf{Text-only PRM}} & Step-level Error Localization & 
ProcessBench~\citep{zheng2024processbench}, 
PRMBench~\citep{song2025prmbench}, 
MR-GSM8K~\citep{zeng2024mrgsm8k}, 
MR-Ben~\citep{zeng2024mrben} \\
& General Reasoning Process Diagnostics & 
UniversalBench~\citep{tan2025aurora},
Socratic-PRMBench~\citep{li2025socraticprmbench}, 
GR-Ben~\citep{sun2026grben} \\
& Domain-Specific Process Supervision & 
MedPRMBench~\citep{wu2026medprmbench}, 
SCIPRM-Bench~\citep{zhao2026sciprm} \\
& PRM Robustness and Bias Diagnostics & 
JETTS~\citep{zhou2025jetts}, 
PRM-BiasBench~\citep{tiwari2026prmbiasbench} \\
\addlinespace[0.35em]
\multirow[t]{4}{=}{\textbf{Multimodal ORM}} & General VLM Response Evaluation & 
VL-RewardBench~\citep{li2024vlrewardbench}, 
Multimodal RewardBench~\citep{yasunaga2025multimodalrewardbench}, 
MM-RLHF-RewardBench~\citep{zhang2025mmrlhf}, 
M-JudgeBench~\citep{chen2026mjudgebench} \\
& Image and Vision-Language Generation Evaluation & 
MJ-Bench~\citep{chen2024mjbench}, 
GenAI-Bench~\citep{li2024genaibench}, 
MMRB2~\citep{hu2026mmrb2}, 
MR2Bench-Image~\citep{zhou2025mr2bench}\\
& Video Reward Evaluation & 
VideoRewardBench~\citep{zhang2025videorewardbench}, 
VideoGen-RewardBench~\citep{liu2025videogenrewardbench}, 
MJ-Bench-Video~\citep{chen2024mjbench}, 
VURB~\citep{wei2026videounderstandingrewardmodeling}, 
MR2Bench-Video~\citep{zhou2025mr2bench} \\
& Specialized Modality/Domain Evaluation & 
JudgeAnything~\citep{pu2025judgeanything}, 
Omni-RewardBench~\citep{jin2025omnireward}, 
CMI-RewardBench~\citep{ma2026cmirewardbench}, 
Med-RewardBench~\citep{ding2025medrewardbench} \\
\addlinespace[0.35em]
\multirow[t]{3}{=}{\textbf{Multimodal PRM}} & Vision-Language Stepwise Reasoning & 
ViLBench~\citep{tu2025vilbench}, 
VisualProcessBench~\citep{wang2025visualprm}, 
VLRMBench~\citep{ruan2025vlrmbench} \\
& Text-Image Process Reasoning & 
MPBench~\citep{xu2025mpbench}, 
ThinkWithImages-PRMBench~\citep{zhou2026thinkwithimages} \\
& Specialized Modality Process Reasoning & 
AudioProcessBench~\citep{zhao2026audioprocessbench} \\
\bottomrule
\end{tabularx}
\end{table}

To rigorously assess the capabilities of various reward models, a multitude of benchmarks have been developed, each tailored to specific modalities, reward formulations, and evaluation methodologies (see Table \ref{tab:rm_benchmarks}). 

A practical evaluation protocol should first identify the RM type under consideration, including its reward granularity (ORM or PRM), reward form (discriminative or generative), and modality (text-only or multimodal). The primary benchmark should then match both this RM type and the target application domain. For example, PRMs should be tested on step-level diagnostics, while domain-specific RMs should be evaluated on corresponding math, code, RAG, medical, or multimodal benchmarks. To assess generalization, the evaluation should further include OOD or perturbation-based benchmarks, which can reveal stylistic, positional, or domain biases, robustness failures, and possible losses of general reward-modeling ability. Finally, benchmark selection should be tied to the downstream role of the RM. As discussed in Section \ref{sec:q4rmevaluation}, isolated pairwise accuracy or correctness scores may not fully reflect practical utility. Therefore, RMs used for BoN selection, search guidance, online RL, or data filtering should be evaluated with metrics that directly measure, or at least reliably proxy, their corresponding downstream performance.

\subsubsection{Benchmarks for Text-only RMs}

\textbf{Text-only ORMs.} In the domain of text-only ORMs, one major line of benchmarks focuses on general preference modeling and judge evaluation, aiming to measure whether reward models can reliably distinguish response quality across diverse instructions, dialogue settings, and human preference scenarios. MT-Bench~\citep{zheng2023llmasajudge} proposes a multi-turn dialogue benchmark to evaluate LLM-as-a-judge models against human-like preferences. RewardBench~\citep{lambert2024rewardbench} is the first comprehensive benchmark for evaluating RMs, covering chat, reasoning, and safety with prompt-chosen-rejected trios to assess subtle preference distinctions. RewardBench 2~\citep{malik2025rewardbench2} extends this with new and more challenging data. RMB~\citep{zhou2025rmb} comprehensively covers a broad range of fine-grained real-world scenarios and introduces Best-of-N evaluation. PPE~\citep{frick2024ppe} serves as a cost-effective proxy for RLHF performance, incorporating human preference and verifiable correctness datasets. JudgeBench~\citep{tan2025judgebench} evaluates LLM-based judges on challenging response pairs spanning knowledge, reasoning, math, and coding.

Beyond general-purpose evaluation, another line of work targets domain-specific judge evaluation, where reward models are tested under more specialized settings such as retrieval-augmented generation, mathematical reasoning, multilingual scenarios, coding, and complex reasoning. RAG-RewardBench~\citep{jin2024ragrewardbench} evaluates RMs in retrieval-augmented generation (RAG) settings, covering RAG-specific challenges such as multi-hop reasoning, citation quality, abstention, and conflict robustness. AceMath-RewardBench~\citep{liu2025acemath} focuses on evaluating math reward models across diverse problem sources and difficulty levels. M-RewardBench~\citep{gureja2024mrewardbench} spans diverse linguistic contexts, testing chat, safety, reasoning, and translation capabilities. LCB-RB~\citep{fan2025posteriorgrpo} constructs preference pairs of textual reasoning for coding tasks, targeting ORMs' ability to assess reasoning quality in code generation. Libra Bench~\citep{zhou2025libra} is a reasoning-oriented benchmark constructed from challenging mathematical problems and advanced reasoning models, evaluating RMs' correctness judgment in complex reasoning scenarios.

A further group of benchmarks emphasizes context- and criteria-conditioned evaluation, requiring reward models to judge outputs according to explicit principles, rubrics, user preferences, or long-context information rather than relying on a fixed notion of response quality. RABench~\citep{yu2025rewardanything} is designed to assess RMs' ability to dynamically adapt evaluation criteria based on explicit natural language principles. IF-RewardBench~\citep{wen2026ifrewardbench} evaluates judge models for instruction-following, covering diverse instruction and constraint types and using preference graphs to support listwise ranking evaluation. RubricBench~\citep{zhang2026rubricbench} evaluates rubric-guided reward models with expert-annotated atomic rubrics and hard pairwise comparisons. Personalized RewardBench~\citep{ma2026personalizedrewardbench} assesses whether RMs can capture individual user preferences through user profiles and personalized rubrics. Long-RewardBench~\citep{tang2025longrm} evaluates reward models in long-context scenarios, testing whether they can handle context lengths far beyond standard RM settings. Long-form RewardBench~\citep{huang2026longformrewardbench} focuses on reward modeling for long-form generation, covering QA, RAG, chat, writing, and reasoning tasks.

In addition, robustness and bias diagnostics have become increasingly important for assessing whether RMs are sensitive to superficial perturbations, minor factual or reasoning errors, and undesirable stylistic or positional biases. RM-Bench~\citep{liu2024rmbench} complements RewardBench by evaluating RM sensitivity to minor errors and stylistic biases. reWordBench~\citep{wu2025rewordbench} systematically transforms reward model inputs in meaning- or ranking-preserving ways to test whether RMs remain robust under paraphrases and other input perturbations.

\textbf{Text-only PRMs.} Shifting from outcome-based to process-based rewards, text-only PRM benchmarks emphasize the correctness of intermediate reasoning steps and the ability to diagnose errors throughout a solution trajectory. A central evaluation focus is step-level error localization, where models are required to identify incorrect intermediate steps rather than merely judge the final answer. ProcessBench~\citep{zheng2024processbench} tasks models with identifying the first erroneous step in math solutions produced by various LLMs. PRMBench~\citep{song2025prmbench} includes fine-grained types of errors to assess models' ability to locate stepwise faults. MR-GSM8K~\citep{zeng2024mrgsm8k} evaluates LLM-as-a-judge's capability to detect and explain reasoning errors in GSM8K-style mathematical solutions. MR-Ben~\citep{zeng2024mrben} extends meta-reasoning evaluation across multiple domains, requiring models to locate, analyze, and explain errors in automatically generated reasoning steps.

Beyond localized error detection, general reasoning process diagnostics assess the quality of entire reasoning trajectories, including long chain-of-thought outputs, diverse policy-generated solutions, and broader forms of process supervision. UniversalBench~\citep{tan2025aurora} includes long CoT output from diverse policy distributions, requiring predictions over entire reasoning trajectories rather than just the first error. Socratic-PRMBench~\citep{li2025socraticprmbench} evaluates PRMs under systematic reasoning patterns, including transformation, decomposition, regathering, deduction, verification, and integration. GR-Ben~\citep{sun2026grben} broadens PRM evaluation beyond mathematical reasoning by assessing process-level error detection across science and logic domains.

Similar to text-only ORMs, PRM evaluation has also expanded to domain-specific process supervision. MedPRMBench~\citep{wu2026medprmbench} evaluates process-level reward models in medical reasoning, covering clinically relevant error types and severity levels. SCIPRM-Bench~\citep{zhao2026sciprm} targets scientific reasoning with tool-integrated trajectories, measuring factual grounding, multi-hop reasoning, and correctness in web-search and Python-based tasks.

Finally, robustness and bias diagnostics are also critical for PRMs, since step-level supervision can be affected by spurious cues, test-time artifacts, and systematic biases in reasoning traces. JETTS~\citep{zhou2025jetts} focuses on test-time tasks, including response reranking, step-level beam search, and critique-based response refinement. PRM-BiasBench~\citep{tiwari2026prmbiasbench} extends process-level evaluation with controlled perturbations, diagnosing whether PRMs are robust to transformations and adversarially exploitable biases.

\subsubsection{Benchmarks for Multimodal RMs}

\textbf{Multimodal ORMs.} As reward modeling extends from text-only settings to multimodal inputs and outputs, multimodal ORM benchmarks evaluate whether models can judge response quality while grounding their decisions in visual, linguistic, or cross-modal evidence. A first group focuses on general VLM response evaluation, covering tasks such as multimodal question answering, visual reasoning, hallucination detection, safety, and human preference alignment. VL-RewardBench~\citep{li2024vlrewardbench} challenges vision-language generative RMs on tasks of general multimodal queries, visual hallucination detection, and complex reasoning. Multimodal RewardBench~\citep{yasunaga2025multimodalrewardbench} provides a more holistic evaluation across six key domains, including general correctness, preference, knowledge, reasoning, safety, and visual question-answering. MM-RLHF-RewardBench~\citep{zhang2025mmrlhf} evaluates multimodal reward models in the MM-RLHF alignment setting, using fine-grained human preference annotations over image, multi-image, and video interactions. M-JudgeBench~\citep{chen2026mjudgebench} evaluates MLLM-as-a-judge systems through a capability-oriented design, covering pairwise CoT comparison, length bias avoidance, and process error detection.

Another important direction concerns image and vision-language generation evaluation, where reward models act as judges for generated images or multimodal responses with respect to alignment, perceptual quality, safety, and preference. MJ-Bench~\citep{chen2024mjbench} assesses multimodal RMs as judges for text-to-image generation, covering alignment, safety, quality, and bias. GenAI-Bench~\citep{li2024genaibench} evaluates compositional text-to-visual generation, emphasizing attributes, relationships, counting, comparison, logic, and other higher-order prompt requirements. MMRB2~\citep{hu2026mmrb2} evaluates omni reward models on multimodal understanding and interleaved image-text generation, spanning text-to-image, image editing, interleaved generation, and multimodal reasoning. MR2Bench-Image~\citep{zhou2025mr2bench} introduces explicit multi-response rankings for image-related reward evaluation, supporting listwise reranking beyond binary preference pairs.

Beyond static images, video reward evaluation introduces additional challenges involving temporal consistency, dynamic visual understanding, and instruction-video alignment. VideoRewardBench~\citep{zhang2025videorewardbench} evaluates multimodal reward models for video understanding across perception, knowledge, reasoning, and safety. VideoGen-RewardBench~\citep{liu2025videogenrewardbench} assesses video reward models for text-to-video generation using large-scale human preference annotations over prompt-video pairs. MJ-Bench-Video~\citep{chen2024mjbench} extends fine-grained multimodal judging to video generation, evaluating video preference assessment under multiple quality and alignment criteria. VURB~\citep{wei2026videounderstandingrewardmodeling} evaluates video understanding reward models with long chain-of-thought reasoning traces and majority-voting protocols across general, long, and reasoning-oriented video tasks. MR2Bench-Video~\citep{zhou2025mr2bench} extends multi-response reward evaluation to video, enabling N-way ranking rather than only pairwise comparison.

Finally, specialized modality and domain evaluation further broadens multimodal reward modeling beyond standard image-text or video-text settings. JudgeAnything~\citep{pu2025judgeanything} evaluates MLLM-as-a-judge capabilities across any-to-any modality tasks, including pairwise comparison and score-based evaluation with human judgments and rubrics. Omni-RewardBench~\citep{jin2025omnireward} targets generalist omni-modal reward modeling with free-form preferences across text, image, video, audio, and 3D tasks. CMI-RewardBench~\citep{ma2026cmirewardbench} evaluates music reward models under compositional multimodal instructions, covering musicality, text-music alignment, and instruction-level alignment with text, lyrics, and reference audio. Med-RewardBench~\citep{ding2025medrewardbench} evaluates reward models and judges for medical multimodal LLMs, emphasizing diagnostic accuracy, clinical relevance, and expert-aligned medical judgment.

\textbf{Multimodal PRMs.} Mirroring the distinction in text-only settings, multimodal PRM benchmarks focus on stepwise reasoning and process supervision in multimodal contexts. One major line of work evaluates vision-language stepwise reasoning, where models must assess intermediate reasoning steps grounded in visual evidence. ViLBench~\citep{tu2025vilbench} employs a Best-of-N selection accuracy metric for vision-language PRM evaluation and requires intensive step-wise reward signals. VisualProcessBench~\citep{wang2025visualprm} uses human-annotated, stepwise labels to assess reasoning correctness in multimodal reasoning tasks. VLRMBench~\citep{ruan2025vlrmbench} further broadens the scope, introducing more challenging, diverse tasks and fine-grained step-level evaluation for multimodal reasoning, outcome judgment, and critique generation.

Another line focuses on text-image process reasoning, where reasoning unfolds through both textual explanations and visual information. MPBench~\citep{xu2025mpbench} evaluates multimodal PRMs through step correctness, answer aggregation, and reasoning process search, thereby testing how PRMs support different stages of multimodal reasoning. ThinkWithImages-PRMBench~\citep{zhou2026thinkwithimages} evaluates PRMs' ability to detect errors in interleaved text-image reasoning trajectories under the ``thinking with images'' paradigm.

Finally, specialized modality process reasoning extends PRM evaluation beyond vision-language tasks. AudioProcessBench~\citep{zhao2026audioprocessbench} evaluates step-level process error identification in audio-grounded reasoning, covering step correctness, audio-specific error types, and chain-level aggregation.\\

\tikzstyle{my-box}=[
rectangle,
draw=hidden-black,
rounded corners,
text opacity=1,
minimum height=1.5em,
minimum width=5em,
inner sep=2pt,
align=center,
fill opacity=.5,
]
\tikzstyle{leaf}=[
my-box, 
minimum height=1.5em,
fill=yellow!20, 
text=black,
align=left,
font=\normalsize,
inner xsep=5pt,
inner ysep=4pt,
align=left,
text width=45em,
]
\tikzstyle{leaf2}=[
my-box, 
minimum height=1.5em,
fill=purple!20, 
text=black,
align=left,
font=\normalsize,
inner xsep=5pt,
inner ysep=4pt,
]
\tikzstyle{leaf3}=[
my-box, 
minimum height=1.5em,
fill=hidden-blue!40, 
text=black,
align=left,
font=\normalsize,
inner xsep=5pt,
inner ysep=4pt,
]
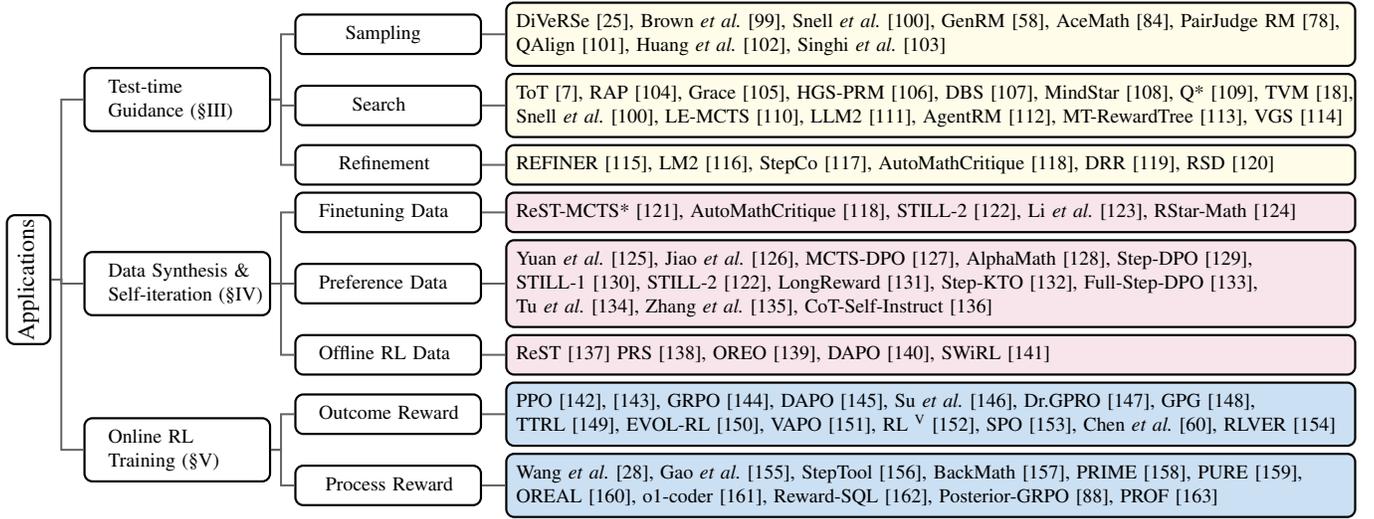
\begin{figure*}[t]
\vspace{-2mm}
\centering
\resizebox{\textwidth}{!}{
	\begin{forest}
		forked edges,
		for tree={
			grow=east,
			reversed=true,
			anchor=base west,
			parent anchor=east,
			child anchor=west,
			base=left,
			font=\large,
			rectangle,
			draw=hidden-black,
			rounded corners,
			align=left,
			minimum width=4em,
			edge+={darkgray, line width=1pt},
			s sep=18pt,
			inner xsep=2pt,
			inner ysep=4pt,
			line width=1.1pt,
			ver/.style={rotate=90, child anchor=north, parent anchor=south, anchor=center},
		},
		where level=1{text width=9em,font=\normalsize,}{},
        where level=2{text width=9em,font=\normalsize,}{},
        where level=3{text width=9.5em,font=\normalsize,}{},
        where level=4{text width=52em,font=\normalsize,}{},
[Applications, ver
		[\ \ \ Test-time \\\ \ \ Guidance~(\S\ref{sec:testtimeguidance})
			[\ \ \ \ \ \ \ Sampling 
				[DiVeRSe~\citep{li2023diverse}{,} 
                Brown \emph{et al.}~\citep{brown2024largelanguagemonkeys}{,} 
                Snell \emph{et al.}~\citep{snell2024scalingllmtesttimecompute}{,} GenRM~\citep{zhang2024genrmcot}{,} 
                AceMath~\citep{liu2025acemath}{,} \\
                PairJudge RM~\citep{liu2025pairjudgerm}{,} 
                QAlign~\citep{faria2025qalign}{,}
                Huang \emph{et al.}~\citep{huang2025inferencetimepessimism}{,}
                Singhi \emph{et al.}~\citep{singhi2025solveverifycomputeoptimalproblem}{,}
                CRM~\citep{zhang2026crm}{,} \\
                TrajSelector~\citep{yu2025trajselector}{,}
                Tang \emph{et al.}~\citep{tang2026exploringgenerativeprocessreward}{,}
                AggLM~\citep{zhao2025agglm}{,}
                Best-of-Majority~\citep{di2025bestofmajority}{,} \\
                Majority-of-the-bests~\citep{rakhsha2025majoritybests}{,}
                RoBoN~\citep{geuter2025robon}{,}
                DAJ~\citep{qin2026daj}{,}
                CodeScaler~\citep{zhu2026codescaler}{,}
                MSV~\citep{kim2026msv}{,} \\
                Sriraman and Block~\citep{sriraman2026revisitingsuboptimality}{,}
                Caution~\citep{yu2026curiositycaution}{,}
                Lu \emph{et al.}~\citep{lu2026robustrewardmodeling}{,}
                PRISM~\citep{agrawal2026hiddenbiasprocessreward}
                ,leaf, text width=42em]
			]
			[\ \ \ \ \ \ \ \  Search
				[ToT~\citep{yao2023tot}{,}
                RAP~\citep{hao2023rap}{,}
                Grace~\citep{khalifa2023grace}{,}
                HGS-PRM~\citep{ma2023hgsprm}{,}
                DBS~\citep{zhu2024dbs}{,} 
                MindStar~\citep{kang2024mindstar}{,}
                Q*~\citep{wang2024qstar}{,} \\
                TVM~\citep{lee2025tvm}{,} 
                Snell \emph{et al.}~\citep{snell2024scalingllmtesttimecompute}{,}
                LE-MCTS~\citep{park2024lemcts}{,}
                LLM2~\citep{yang2025llm2}{,}
                AgentRM~\citep{xia2025agentrm}{,} \\
                MT-RewardTree~\citep{feng2025mtrewardtree}{,}
                VGS~\citep{wang2025vgs}{,}
                ToolPRM~\citep{lin2026toolprm}{,}
                GroundedPRM~\citep{zhang2025groundedprm}{,}
                MASPRM~\citep{yazdani2026masprm}{,}\\
                AgentPRM~\citep{xi2025agentprm}{,}
                PUM~\citep{zhou2026pum}{,}
                PROPA~\citep{jiang2025propa}{,}
                Opedal \emph{et al.}~\citep{opedal2026learningreasonefficientlya}{,}
                AgentV-RL~\citep{zhang2026agentvrl}
                , leaf, text width=42em]
			]
			[\ \ \ \ \ \ Refinement
				[REFINER~\citep{pau2024refiner}{,}
                LM2~\citep{juneja2024lm2}{,}
                StepCo~\citep{wu2024stepco}{,} 
                AutoMathCritique~\citep{xi2024automathcritique}{,}
                DRR~\citep{yang2024drr}{,}
                RSD~\citep{liao2025rsd}{,}\\
                SWE-PRM~\citep{gandhi2025sweprm}{,}
                SMRC~\citep{zeng2025smrc}{,}
                CAMEL~\citep{zhu2026camel}{,}
                DeepTool~\citep{he2026deeptool}{,}
                SRaR~\citep{xie2026srar}{,}
                EVOLVE~\citep{zeng2025evolve}{,}\\
                Critique-GRPO~\citep{zhang2026critiquegrpo}
                , leaf, text width=42em]
			]
		]
		[\ \ \ Data Synthesis \& \\ \ \ \ Self-iteration~(\S\ref{sec:datasynthesis})
			[\ \ \ Finetuning Data 
				[ReST-MCTS*~\citep{zhang2024restmcts}{,}
                AutoMathCritique~\citep{xi2024automathcritique}{,}
                STILL-2~\citep{min2024imitateexploreselfimprovereproduction}{,}
                Li \emph{et al.}~\citep{li2025mctsgeneration}{,}
                RStar-Math\citep{guan2025rstarmath}{,}\\
                MathSE~\citep{chen2025mathse}{,}
                WebSTAR~\citep{he2026webstar}{,}
                AutoTraj~\citep{gong2026autotraj}{,}
                RPM-MCTS~\citep{lin2025rpmmcts}{,}
                StepPRM-RTL~\citep{vijayaraghavan2026stepprmrtl}{,}\\
                SWE-Trace~\citep{han2026swetrace}{,}
                StepORLM~\citep{zhou2025steporlm}{,}
                GroundedPRM~\citep{zhang2025groundedprm}{,}
                , leaf2, text width=42em]
			]
			[\ \ \ Preference Data
				[Yuan \emph{et al.}~\citep{yuan2025selfrewardinglanguagemodels}{,}
                Jiao \emph{et al.}~\citep{jiao2024processrewardsysthesizing}{,}
                MCTS-DPO~\citep{xie2024mctsdpo}{,}
                AlphaMath~\citep{chen2024alphamath}{,}
                Step-DPO~\citep{lai2024stepdpo}{,} \\
                STILL-1~\citep{jiang2024still1}{,}               
                STILL-2~\citep{min2024imitateexploreselfimprovereproduction}{,} 
                LongReward~\citep{zhang2024longreward}{,}
                Step-KTO~\citep{lin2025stepkto}{,}
                Full-Step-DPO~\citep{xu2025fullstepdpo}{,} \\
                Tu \emph{et al.}~\citep{tu2025iterativedpo}{,}
                Zhang \emph{et al.}~\citep{zhang2025processbasedselfrewardinglanguagemodels}{,}
                CoT-Self-Instruct~\citep{yu2025cotselfinstruct}{,}
                DPPrefSyn~\citep{gao2026dpprefsyn}{,}
                SAO~\citep{yin2025sao}{,} \\
                MADPO~\citep{rho2026madpo}{,}
                Lu \emph{et al.}~\citep{lu2026hardnegativesample}{,}
                CSO~\citep{li2026cso}{,}
                EvoLM~\citep{li2026evolm}{,}
                CausalFlow~\citep{bonagiri2026causalflow}{,}
                RoRo~\citep{ye2026roro}{,} \\
                OrchRM~\citep{tsang2026orchrm}{,}
                ARCO~\citep{tian2026arco}{,}
                TGPO~\citep{chen2025tgpo}
                , leaf2, text width=42em]
			]
                [\ \ \ Offline RL Data
                [ReST~\citep{gulcehre2023rest}
                PRS~\citep{ye2024prs}{,}
                OREO~\citep{wang2024oreo}{,}
                DAPO~\citep{liu2024dapo}{,}
                SWiRL~\citep{goldie2025swirl}{,}
                DQO~\citep{ji2025dqo}{,}
                BCPG-NSA~\citep{yang2025BCPG-NSA}{,} \\
                SPARE~\citep{rizvi2026spare}
                , leaf2, text width=42em]
                ]
	]
		[\ \ \ Online RL \\\ \ \ Training~(\S\ref{sec:reinforcementlearning})
			[\ \ \ Outcome Reward
				[PPO~\citep{schulman2017ppo, ouyang2022rlhf}{,}
                GRPO~\citep{shao2024grpodeepseekmath}{,}
                DAPO~\citep{yu2025dapo}{,}
                Su \emph{et al.}~\citep{su2025crossingrewardbridge}{,}
                Dr.GPRO~\citep{liu2025drgrpo}{,}
                GPG~\citep{chu2025gpg}{,} \\
                TTRL~\citep{zuo2025ttrl}{,}
                EVOL-RL~\citep{zhou2025evolrl}{,}
                VAPO~\citep{yue2025vapo}{,}
                RL \textsuperscript{V}~\citep{sareen2025rlv}{,}
                SPO~\citep{guo2025spo}{,}
                Chen \emph{et al.}~\citep{chen2025learningreasonfactuality}{,} \\
                RLVER~\citep{wang2025rlver}{,}
                CE-RM~\citep{hu2026cerm}{,}
                RLCS~\citep{li2026rlcs}{,}
                GRRM~\citep{yang2026grrm}{,}
                RubricARM~\citep{xu2026rubricarm}{,}
                CausalRM~\citep{wang2026causalrm}{,} \\
                Asghari \emph{et al.}~\citep{asghari2026efficientexplorationscale}{,}
                GPRL~\citep{umer2026gprl}{,}
                UARM~\citep{pan2026uarm}{,}
                EvoRubrics~\citep{ding2026evorubrics}{,}
                Lin \emph{et al.}~\citep{lin2026betterliterarytranslation}
                , leaf3, text width=42em]
			]
			[\ \ \ \ Process Reward
				[Wang \emph{et al.}~\citep{wang2024mathshepherd}{,}
                Gao \emph{et al.}~\citep{gao2024designingeffectiverlreward}{,}
                StepTool~\citep{yu2025steptool}{,}
                BackMath~\citep{zhang2025backmath}{,}
                PRIME~\citep{cui2025prime}{,}
                PURE~\citep{cheng2025pure}{,} \\
                OREAL~\citep{lyu2025oreal}{,}
                o1-coder~\citep{zhang2024o1coder}{,}
                Reward-SQL~\citep{zhang2025rewardsql}{,}
                Posterior-GRPO~\citep{fan2025posteriorgrpo}{,}
                PROF~\citep{ye2025prof}{,} \\
                PRPO~\citep{ding2026prpo}{,} 
                VPPO~\citep{liu2026vppo}{,}
                rePIPL~\citep{wu2026repirl}{,}
                FaithRL~\citep{nie2026faithrl}{,}
                ITPO~\citep{wang2026itpo}{,}
                HISR~\citep{lu2026hisr}{,}
                PAPO~\citep{tan2026papo}{,} \\
                uPRM~\citep{gadetsky2026uprm}{,}
                PUM~\citep{zhou2026pum}{,}
                ARCO~\citep{tian2026arco}{,}
                ARBOR~\citep{liu2026arbor}
                , leaf3, text width=42em]
			]
		]
]
	\end{forest}
}
\caption{Applications of RMs in LLM reasoning}
\label{fig:taxonomy}

\end{figure*}

\section{RM Application 1: Test-Time Guidance}
\label{sec:testtimeguidance}

Test-time scaling has emerged as a pivotal method for improving LLM reasoning. By dynamically allocating more computational effort during inference, models are enabled to `think harder', spending additional time on complex problems to enhance accuracy. Unlike traditional parameter updates, this approach optimizes performance by adjusting real-time computation without modifying model weights.

The most straightforward implementation involves repeated sampling, where language models generate multiple reasoning trajectories and the highest-reward solution is selected (Section \ref{sec:sampling}). To further optimize computational efficiency, more advanced approaches employ guided search techniques that selectively explore high-potential steps (Section \ref{sec:search}), or self-correction mechanisms that iteratively refine outputs (Section \ref{sec:refinement}). 
This section systematically examines how reward models can enhance three fundamental test-time computation strategies: selection, search, and refinement. We list brief summaries of related works in Table \ref{tab:rm-testtime-selection}, \ref{tab:rm-testtime-search} and \ref{tab:rm-testtime-refinement}.

\subsection{Sampling and Selection}
\label{sec:sampling}

By \emph{sampling}, multiple candidate solutions are drawn from a policy model. \emph{Selection} then chooses a single final solution from these candidates according to a decision rule. A lightweight baseline is \emph{self-consistency}~\citep{wang2023cotselfconsistency}, which samples many completions and returns the answer that appears most frequently (i.e., a majority vote over final answers) without an explicit verifier.
In contrast, the generator-verifier paradigm equips selection with reward scores from PRMs or ORMs to explicitly verify the correctness of each solution.
Whereas self-consistency may fail when the policy model has a higher probability of generating incorrect answers, selection methods with reward models can identify the correct solution regardless of its model generation probability.
The most prevalent model-based sampling and selection strategy is Best-of-N (BoN), which samples $N$ solutions for the same question and selects the one with the highest reward.
In practice, both ORMs and PRMs are widely adopted for solution selection.

\textbf{ORM-based selection}.
Early studies employed ORMs to assess candidate solutions holistically. 
For instance, Cobbe \emph{et al.}~\citep{cobbe2021trainingverifiers} and Uesato \emph{et al.}~\citep{uesato2022solvingmathwordproblems} both train ORMs for math problems and select the highest-ranked solutions as final answers. 
AceMath~\citep{liu2025acemath} attempts to push the envelope in the mathematical domain by systematically curating datasets for supervised fine-tuning, thereby facilitating the training of stronger policy models and ORMs.
Recent extensions improve outcome-level selection by changing the candidate pool or the aggregation rule. RoBoN~\citep{geuter2025robon} routes samples across multiple LLMs using reward scores and answer agreement, while AggLM~\citep{zhao2025agglm} trains an aggregation model to reconcile sampled solutions. Best-of-Majority~\citep{di2025bestofmajority} and Majority-of-the-Bests~\citep{rakhsha2025majoritybests} further modify BoN by combining reward scores, answer frequency, and bootstrapped selection.

Additionally, the utility of generative ORMs in BoN selection has been explored.
Zhang \emph{et al.}~\citep{zhang2024genrmcot} find that generative ORMs can improve BoN performance via CoT reasoning. 
PairjudgeRM~\citep{liu2025pairjudgerm} introduces an improved tournament-based mechanism that enhances BoN accuracy by training a generative RM to perform pairwise comparisons.
Mirror-Critique~\citep{yang2025critiqueverify} further improves generative verifier selection by training the verifier with critique signals and using multiple generated critiques for weighted voting or abstention.
In code generation, DAJ~\citep{qin2026daj} trains a data-reweighted LLM judge for BoN under distribution shift, while CodeScaler~\citep{zhu2026codescaler} provides an RM that reranks code candidates with much lower latency than unit-test-based checking.

To address the problem of imperfect RMs in solution selection, alternative sampling methods have been proposed. 
Specifically, to overcome the over-optimization of RMs in BoN as the number of samples $N$ increases, QAlign~\citep{faria2025qalign} leverages RMs to guide Markov Chain Monte Carlo (MCMC) sampling at test time,
enabling better-aligned outputs without model fine-tuning. 
Similarly, Huang \emph{et al.}~\citep{huang2025inferencetimepessimism} mitigates the reward hacking problem in BoN by applying inference-time pessimism with RM-based rejection sampling to conservatively downweight overconfident, high-uncertainty candidates and foster more consistent performance improvements. Caution~\citep{yu2026curiositycaution} reduces BoN reward hacking by lowering scores for uncertain candidates, while Lu \emph{et al.}~\citep{lu2026robustrewardmodeling} make RMs rely less on shortcuts such as response length, improving robust BoN selection.

\textbf{PRM-based selection}.
Some previous works find that ORMs may fail to detect detailed errors in responses; as a result, the assigned score often focuses exclusively on the final answer~\citep{yu2024ovm, zhang2025prmlessons}. 
Consequently, subsequent research has shifted toward leveraging step-wise scoring from PRMs, combined with various aggregation functions (e.g., final-value, minimum-value, product of process rewards), to compute an overall score for each solution. 
These approaches~\citep{lightman2023letsverifystepstep, wang2024mathshepherd, li2025pqm, yuan2024implicitprm} typically yield better performance than either ORMs or simple majority voting.
CRM~\citep{zhang2026crm} links step rewards to the final outcome, making it easier to compare across sampled trajectories in BoN.
PRISM~\citep{agrawal2026hiddenbiasprocessreward} shows that false-positive step rewards can mislead BoN selection and PRISM improves PRMs by learning clearer comparisons between correct and flawed steps.
TrajSelector~\citep{yu2025trajselector} reduces the cost of PRM-based BoN by using hidden states from the sampler model with a small verifier, while Tang \emph{et al.} \citep{tang2026exploringgenerativeprocessreward} show that generative PRMs can also support answer selection for table question answering.

Some methods further extend BoN with weighted voting.
For example, DiVeRSe~\citep{li2023diverse} performs weighted voting based on PRM final scores for output selection; Self-Check~\citep{miao2023selfcheck} first derives step-wise confidence scores through LLM self-validation and then aggregates via weighted voting.
Moreover, Zhang \emph{et al.}~\citep{zhang2025prmlessons} experimentally evaluate different aggregation functions for PRM scoring in BoN selection, finding that the optimal strategy may depend on the specific PRM design.

\textbf{Key findings}. 
Increasing the number of test-time samples and enhancing verifier accuracy are both promising directions.
MSV~\citep{kim2026msv} shows that verifiers can score a candidate set jointly rather than one response at a time, improving BoN selection and enabling early stopping.
Brown \emph{et al.}~\citep{brown2024largelanguagemonkeys} demonstrate that the probability of generating at least one correct solution follows a log-linear growth pattern with increased sampling iterations, and the verifier accuracy in identifying correct solutions is critical. Snell \emph{et al.}~\citep{snell2024scalingllmtesttimecompute} show that adaptively scaling test-time compute can sometimes outperform simply increasing model size.
However, Singhi \emph{et al.}~\citep{singhi2025solveverifycomputeoptimalproblem} suggest that generating more candidate solutions can be more computationally efficient than deploying a generative RM.
Sriraman and Block~\citep{sriraman2026revisitingsuboptimality} further analyze BoN under win-rate objectives, showing that tuned BoN can be optimal in some settings but still remains exposed to reward hacking.
Thus, the optimal allocation of computation between solution generation and verification is a crucial consideration, particularly in scenarios involving costly generative RMs.

\begin{table}[!t]
\centering
\caption{Summary of RM-based sampling and selection methods.}
\label{tab:rm-testtime-selection}
\tiny
\setlength{\tabcolsep}{2pt}
\renewcommand{\arraystretch}{0.86}
\begin{tabular}{@{}p{0.24\linewidth}p{0.18\linewidth}p{0.22\linewidth}p{0.26\linewidth}@{}}
\toprule
\textbf{RM / work} & \textbf{Domain} & \textbf{Test-time guidance} & \textbf{Core description} \\
\midrule
\multicolumn{4}{l}{\textit{Outcome-guided selection}} \\
\midrule
Self-Consistency~\citep{wang2023cotselfconsistency} & General reasoning & Majority voting & Aggregates final answers without an explicit verifier. \\
Cobbe et al.~\citep{cobbe2021trainingverifiers} & Math & ORM BoN reranking & Verifier selects the highest-ranked GSM8K solution. \\
Uesato et al.~\citep{uesato2022solvingmathwordproblems} & Math & ORM/PRM selection & Compares outcome and process feedback for GSM8K. \\
AceMath~\citep{liu2025acemath} & Math & ORM BoN reranking & Curates math data to strengthen policy and ORM. \\
Brown et al.~\citep{brown2024largelanguagemonkeys} & General reasoning & Repeated sampling & Studies scaling laws for finding correct samples. \\
Snell et al.~\citep{snell2024scalingllmtesttimecompute} & Math/general & Adaptive test-time compute & Allocates compute between sampling and verification. \\
GenRM~\citep{zhang2024genrmcot} & Math/algorithmic & Generative BoN verifier & Casts reward modeling as next-token verification. \\
PairJudge RM~\citep{liu2025pairjudgerm} & Math & Pairwise tournament & Replaces absolute scores with knockout comparisons. \\
Mirror-Critique~\citep{yang2025critiqueverify} & General reasoning & Critique-weighted selection & Uses multiple verifier critiques for voting/abstention. \\
QAlign~\citep{faria2025qalign} & Alignment/math & RM-guided MCMC & Samples from an aligned distribution without finetuning. \\
InferenceTimePessimism~\citep{huang2025inferencetimepessimism} & Alignment & Rejection sampling & Uses pessimism to avoid reward-hacked candidates. \\
Caution~\citep{yu2026curiositycaution} & Alignment & Pessimistic BoN & Penalizes uncertain/OOD candidates during selection. \\
Causal RM~\citep{wang2026causalrm} & General/alignment & Robust BoN & Suppresses shortcuts such as length bias. \\
AggLM~\citep{zhao2025agglm} & Math/general & Solution aggregation & Trains an aggregator to reconcile candidate answers. \\
Best-of-Majority~\citep{di2025bestofmajority} & Math/general & Reward-frequency selection & Restricts selection to frequent high-reward answers. \\
Majority-of-the-Bests~\citep{rakhsha2025majoritybests} & Math/general & Bootstrapped BoN & Chooses the mode of BoN output distribution. \\
RoBoN~\citep{geuter2025robon} & Math/general & Routed multi-LLM BoN & Routes samples using reward and answer agreement. \\
DAJ~\citep{qin2026daj} & Code & Judge-based BoN & Reweights judge training data for distribution shift. \\
CodeScaler~\citep{zhu2026codescaler} & Code & Execution-free reranking & Scores code candidates without running unit tests. \\
MSV~\citep{kim2026msv} & General reasoning & Setwise BoN verifier & Scores candidate sets jointly and enables early stop. \\
Singhi et al.~\citep{singhi2025solveverifycomputeoptimalproblem} & Math/general & Solve--verify budgeting & Analyzes when verification beats more generation. \\
Sriraman and Block~\citep{sriraman2026revisitingsuboptimality} & Alignment & BoN objective analysis & Characterizes optimality and reward-hacking limits. \\
\midrule
\multicolumn{4}{l}{\textit{Process-guided selection}} \\
\midrule
DiVeRSe~\citep{li2023diverse} & Math & PRM-weighted voting & Uses step-aware verifier scores for answer voting. \\
OVM~\citep{yu2024ovm} & Math & Step-value selection/search & Trains value estimates from outcome supervision. \\
Lightman et al.~\citep{lightman2023letsverifystepstep} & Math & PRM BoN reranking & Human step labels train a reliable PRM. \\
Math-Shepherd~\citep{wang2024mathshepherd} & Math & PRM reranking/RL & Builds step rewards automatically from rollouts. \\
PQM~\citep{li2025pqm} & Math & Q-value PRM ranking & Learns comparative Q-value rewards for steps. \\
ImplicitPRM~\citep{yuan2024implicitprm} & Math & Label-free PRM selection & Derives process rewards from outcome labels. \\
Zhang et al.~\citep{zhang2025prmlessons} & Math & PRM aggregation study & Compares final/min/product aggregation for BoN. \\
CRM~\citep{zhang2026crm} & Math & Conditional PRM BoN & Links step rewards to final outcome quality. \\
PRISM~\citep{agrawal2026hiddenbiasprocessreward} & Math/general & Robust PRM selection & Reduces false-positive step rewards via contrastive ranking. \\
TrajSelector~\citep{yu2025trajselector} & Math/general & Lightweight PRM BoN & Scores sampler hidden states to cut verifier cost. \\
Tang et al.~\citep{tang2026exploringgenerativeprocessreward} & Table QA & Generative PRM selection & Applies process judging to semi-structured reasoning. \\
SelfCheck~\citep{miao2023selfcheck} & General reasoning & Step-confidence voting & Self-validates steps and aggregates confidence. \\
\bottomrule
\end{tabular}
\end{table}

\subsection{Search}
\label{sec:search}

Different from the aforementioned selection methods, which select from a fixed set of generated candidates, test-time search mechanisms dynamically generate answers by actively exploring multiple reasoning paths during inference. 
A classical framework for test-time search is tree search, which constructs a tree of reasoning steps to find an optimal reasoning path. 
Tree-based methods vary in how they balance exploration (evaluating new paths) and exploitation (refining known paths), often differing in their use of heuristic evaluations and backtracking strategies.

For example, some of them prioritize efficiency by irreversibly pruning decisions. 
These methods expand nodes based on heuristic scores at each step without revisiting prior decisions, maintaining fixed search strategies. 
Typical implementations include greedy search and beam search. They typically rely on PRMs for step-wise guidance, and ORMs may not be applicable for such methods~\citep{zhou2025jetts}.
Greedy search selects the highest-scoring path at every step using an RM. For instance, GRACE~\citep{khalifa2023grace} employs a discriminative ORM as a PRM during test-time greedy search. HGS-PRM~\citep{ma2023hgsprm} employs trained PRM directly. MT-RewardTree~\citep{feng2025mtrewardtree} further demonstrates the efficacy of greedy decoding in machine translation tasks. 
Beam search retains a fixed number of top candidates at each step, where RM scores can improve the search quality. Methods such as DBS~\citep{zhu2024dbs}, MindStar~\citep{kang2024mindstar}, AgentRM~\citep{xia2025agentrm} and PUM~\citep{zhou2026pum} incorporate PRM to execute step-wise beam search. VGS~\citep{wang2025vgs} defines a fixed-length block as the atomic search unit. Token-level search or decoding with a token-wise RM can also be effective, as shown in ARGS~\citep{khanov2024args}, TVM~\citep{lee2025tvm}, and LLM2~\citep{yang2025llm2}.
Recent work further extends PRM-guided search to structured and agentic outputs: ToolPRM~\citep{lin2026toolprm} scores fine-grained function-calling decisions during beam search, while AgentPRM~\citep{xi2025agentprm} and MASPRM~\citep{yazdani2026masprm} score intermediate agent actions or multi-agent transcripts to guide test-time exploration.

In contrast, backtracking-enabled algorithms dynamically adjust their search paths dynamically during generation. 
These approaches iteratively refine node evaluations through backtracking or simulation. Notable examples include Monte-Carlo-Tree-Search (MCTS) methods and A*. Specifically, 
LATS~\citep{zhou2024lats} utilizes self-evaluated values to guide MCTS.
GroundedPRM~\citep{zhang2025groundedprm} and PROPA~\citep{jiang2025propa} use MCTS to obtain process-level supervision and then train PRMs to guide test-time MCTS search in mathematical or visual reasoning tasks.
TS-LLM~\citep{feng2024tsllm} trains an ORM to replace pretrained LLM as the value function. 
LE-MCTS~\citep{park2024lemcts} increases step diversity by ensembling multiple LLMs within the MCTS framework.
A* combines path cost and heuristic estimates to guide search, heavily relying on the quality of the heuristic function. Q*~\citep{wang2024qstar} further combines a trained Q-value model with PRM to guide A* search. More recently, A* post-training~\citep{opedal2026learningreasonefficientlya} uses A*-informed PRMs to train models toward reasoning paths that balance correctness and efficiency.
The choice between these strategies often depends on the characteristics of tasks. Static pruning is suitable for scenarios requiring low latency, while backtracking methods excel in complex, error-sensitive tasks.

\begin{table}[!t]
\centering
\caption{Summary of RM-based test-time search methods.}
\label{tab:rm-testtime-search}
\tiny
\setlength{\tabcolsep}{2pt}
\renewcommand{\arraystretch}{0.88}
\begin{tabular}{@{}p{0.24\linewidth}p{0.18\linewidth}p{0.22\linewidth}p{0.26\linewidth}@{}}
\toprule
\textbf{RM / work} & \textbf{Domain} & \textbf{Test-time guidance} & \textbf{Core description} \\
\midrule
\multicolumn{4}{l}{\textit{Outcome-guided search}} \\
\midrule
ToT~\citep{yao2023tot} & General reasoning & Tree search & Uses thought-level evaluation to explore alternatives. \\
RAP~\citep{hao2023rap} & Planning/reasoning & World-model tree search & Treats reasoning as planning with state evaluation. \\
LATS~\citep{zhou2024lats} & Agents/reasoning & MCTS & Combines self-evaluation with tree-search planning. \\
TS-LLM~\citep{feng2024tsllm} & Math/general & ORM-guided MCTS & Trains an ORM value model for tree decoding. \\
\midrule
\multicolumn{4}{l}{\textit{Process-guided search}} \\
\midrule
GRACE~\citep{khalifa2023grace} & Math & Greedy PRM search & Uses discriminator scores to guide CoT steps. \\
HGS-PRM~\citep{ma2023hgsprm} & Math & Greedy PRM search & Uses a step-level reward model as navigator. \\
DBS~\citep{zhu2024dbs} & Math & Deductive beam search & Keeps beam candidates with deducible rationales. \\
MindStar~\citep{kang2024mindstar} & Math & PRM-guided search & Enhances pretrained LLMs at inference time. \\
Q*~\citep{wang2024qstar} & Multi-step reasoning & A* / Q-value search & Combines Q-values and PRM for deliberative planning. \\
TVM~\citep{lee2025tvm} & Math & Token-level value search & Scores tokens/steps to improve problem solving. \\
LLM2~\citep{yang2025llm2} & General reasoning & System-2 decoding & Uses token/value feedback for deliberate decoding. \\
LE-MCTS~\citep{park2024lemcts} & Complex reasoning & Ensemble MCTS & Increases step diversity with multiple LLMs. \\
VGS~\citep{wang2025vgs} & CoT reasoning & Block-level value search & Searches fixed-length reasoning blocks efficiently. \\
AgentRM~\citep{xia2025agentrm} & Agents & PRM beam search & Scores intermediate agent decisions for exploration. \\
MT-RewardTree~\citep{feng2025mtrewardtree} & Machine translation & Reward-tree decoding & Guides translation with reward-model tree scoring. \\
ToolPRM~\citep{lin2026toolprm} & Function calling & Fine-grained beam search & Scores intra-call structured-output decisions. \\
AgentPRM~\citep{xi2025agentprm} & Agents & Step-wise PRM search & Measures promise/progress of agent actions. \\
MASPRM~\citep{yazdani2026masprm} & Multi-agent reasoning & Beam/MCTS control & Assigns per-agent, per-action process values. \\
PUM~\citep{zhou2026pum} & LLM reasoning & Prefix utility search & Evaluates prefixes by gain rather than correctness only. \\
GroundedPRM~\citep{zhang2025groundedprm} & Math & MCTS-guided PRM search & Fuses tree feedback with tool-grounded step checks. \\
PROPA~\citep{jiang2025propa} & Visual reasoning & MCTS-guided PRM search & Trains PRM from dense process-level visual rewards. \\
A* post-training~\citep{opedal2026learningreasonefficientlya} & Deductive reasoning & A*-informed PRM & Rewards proofs balancing correctness and efficiency. \\
AgentV-RL~\citep{zhang2026agentvrl} & Agents & Agentic verifier RM & Scales reward modeling with agentic verification. \\
\bottomrule
\end{tabular}
\end{table}

\subsection{Refinement}
\label{sec:refinement}

LLMs can further improve their outputs through iterative self-correction or refinement. Intrinsic self-correction operates without external feedback, relying instead on prompting the LLM to revise its own answers.
However, in tasks where external signals or rewards are available (e.g., code generation~\citep{chen2023teachingselfdebug} or tool use~\citep{gou2024toolselfcorrect}), LLMs can leverage such feedback for more effective refinement. Nonetheless, recent studies have found that some state-of-the-art LLMs still struggle with intrinsic self-correction, due to hallucinations, unreliable verification, or prompt misalignment~\citep{huang2024llmcannotselfcorrect, kamoi2024selfcorrectionsurvey}. Therefore, improving intrinsic self-correction ability is a critical direction.

\textbf{Training to improve self-refinement.} Approaches such as RISE~\citep{qu2024rise}, SCoRe~\citep{kumar2024score}, S$^2$R~\citep{ma2025s2r}, StepAMC~\citep{li2025stepamc}, Xiong \emph{et al.}~\citep{xiong2025selfrewardingcorrection}, PAG~\citep{jiang2025pag}, ThinkTwice~\citep{jiao2026thinktwice}, and PASR~\citep{han2025pasr} seek to improve intrinsic self-correction or self-refinement via supervised fine-tuning or reinforcement learning, often employing verifiable rewards or LLM-as-a-judge signals during training. Recent studies have also begun to incorporate RM-based feedback into training-time self-refinement. For example, EVOLVE~\citep{zeng2025evolve} iteratively collects self-refined responses and constructs preference data through rule-based or RM-based filtering to improve LLM self-correction capability, while Critique-GRPO~\citep{zhang2026critiquegrpo} combines scalar rewards with natural-language critiques from rule or model-based reward systems, enabling the policy model to learn from both initial responses and critique-guided refinements. Recent works also incorporate process rewards into self-refinement training: DeepTool~\citep{he2026deeptool} uses process rewards to supervise each think-act-observe turn in tool-integrated reasoning, and SRaR~\citep{xie2026srar} assigns rubric rewards to individual reasoning steps to reduce reward hacking from repeated self-correction.

\textbf{Self-refinement at test-time.}
At test time, refinement methods typically utilize RMs to guide corrective actions.
Discriminative RMs assist LLMs in identifying when to regenerate outputs and locating errors. For example, StepCo~\citep{wu2024stepco} leverages a PRM to identify errors in mathematical reasoning, prompting the policy model to correct specific mistakes.
SMRC~\citep{zeng2025smrc} treats revision as a sequential decision problem and uses search-derived process rewards to guide corrections of intermediate reasoning steps.
RSD~\citep{liao2025rsd} enhances generation efficiency by having a PRM assess steps produced by a weaker draft model, switching to a stronger target model when verification fails.
DRR~\citep{yang2024drr} injects ORM-based error detection feedback into the context for output refinement.
For software-engineering agents, SWE-PRM~\citep{gandhi2025sweprm} uses an inference-time PRM to detect trajectory-level errors such as redundant exploration and looping, providing lightweight feedback for correction.
Natural language feedback from generative models can provide richer, error-specific guidance when incorporated into the original prompt. Some methods employ self-evaluation feedback~\citep{madaan2023selfrefine, shinn2023reflexion}, while others train dedicated critic models for evaluation and refinement~\citep{pau2024refiner, juneja2024lm2, xi2024automathcritique}.
CAMEL~\citep{zhu2026camel} improves generative reward modeling by first making a lightweight preference judgment and invoking reflective reasoning only for low-confidence cases.

\begin{table}[!t]
\centering
\caption{Summary of RM-based self-refinement methods.}
\label{tab:rm-testtime-refinement}
\tiny
\setlength{\tabcolsep}{2pt}
\renewcommand{\arraystretch}{0.88}
\begin{tabular}{@{}p{0.24\linewidth}p{0.18\linewidth}p{0.22\linewidth}p{0.26\linewidth}@{}}
\toprule
\textbf{RM / work} & \textbf{Domain} & \textbf{Test-time guidance} & \textbf{Core description} \\
\midrule
\multicolumn{4}{l}{\textit{Outcome-guided refinement}} \\
\midrule
DRR~\citep{yang2024drr} & Math reasoning & ORM feedback refinement & Injects error-detection feedback into revision context. \\
CAMEL~\citep{zhu2026camel} & Reward modeling & Confidence-gated reflection & Reflects only when lightweight verdict confidence is low. \\
Self-Refine~\citep{madaan2023selfrefine} & General tasks & Self-feedback refinement & Iteratively revises outputs with natural-language feedback. \\
Reflexion~\citep{shinn2023reflexion} & Agents/code & Verbal feedback refinement & Stores verbal feedback to improve later attempts. \\
\midrule
\multicolumn{4}{l}{\textit{Process-guided refinement}} \\
\midrule
REFINER~\citep{pau2024refiner} & Reasoning & Critique-guided refinement & Gives feedback on intermediate representations. \\
LM2~\citep{juneja2024lm2} & Complex reasoning & Critic-agent refinement & Uses a society of models to critique and solve. \\
AutoMathCritique~\citep{xi2024automathcritique} & Math & Critique-model refinement & Provides test-time and training-time critiques. \\
StepCo~\citep{wu2024stepco} & Math & Stepwise correction & Locates erroneous steps and prompts targeted repair. \\
SMRC~\citep{zeng2025smrc} & Math & Reward-guided correction & Treats revision as sequential error correction. \\
RSD~\citep{liao2025rsd} & Reasoning & PRM-gated speculative decoding & Switches from draft to target model on failure. \\
SWE-PRM~\citep{gandhi2025sweprm} & Software agents & Trajectory correction & Detects loops and redundant exploration. \\
RISE~\citep{qu2024rise} & Agents & Self-improvement training & Teaches agents recursive introspection. \\
SCoRe~\citep{kumar2024score} & General reasoning & RL self-correction & Trains models to revise their own answers. \\
S$^2$R~\citep{ma2025s2r} & General reasoning & RL self-verification & Learns self-verify then self-correct behavior. \\
StepAMC~\citep{li2025stepamc} & Math & Step-level correction RL & Optimizes automatic math correction at step level. \\
Xiong et al.~\citep{xiong2025selfrewardingcorrection} & Math & Self-rewarding correction & Uses model-generated rewards for correction. \\
PAG~\citep{jiang2025pag} & General reasoning & Generative-verifier correction & Uses the policy as verifier across turns. \\
ThinkTwice~\citep{jiao2026thinktwice} & General reasoning & Joint refinement training & Optimizes reasoning and self-refinement together. \\
PASR~\citep{han2025pasr} & General reasoning & Proactive refinement & Revises early before errors compound. \\
EVOLVE~\citep{zeng2025evolve} & General reasoning & RM-filtered refinement data & Iteratively trains on self-refined responses. \\
Critique-GRPO~\citep{zhang2026critiquegrpo} & Math/STEM & Critique-guided RL & Combines scalar rewards with natural-language critiques. \\
DeepTool~\citep{he2026deeptool} & Tool reasoning & Process-supervised RL & Rewards each think--act--observe tool turn. \\
SRaR~\citep{xie2026srar} & Math & Step-wise rubric rewards & Attributes rubric criteria to individual steps. \\
\bottomrule
\end{tabular}
\end{table}

\section{RM Application 2: Synthetic Data Curation and Self-Iteration}
\label{sec:datasynthesis}

The quality of training data is crucial for the performance of LLMs, particularly during post-training stages. However, real-world datasets are often constrained by both quantity and diversity.
This limitation has driven increasing research attention towards synthetic data generation as a critical direction. 
The current trend in data synthesis is gradually shifting to an iterative self-improvement paradigm where LLMs first generate data by themselves, which is used to finetune the LLMs themselves after filtering.

In this pipeline, the effectiveness of data filtering is crucial to the eventual performance of LLMs after finetuning. In certain domains, ground-truth answers may be available, allowing the use of rule-based rewards or human annotators to assess data quality. Nevertheless, ground truth is frequently unavailable in most domains, and manual annotation can become expensive at scale, especially when step-level reward signals are required. To address these challenges, RMs are commonly adopted as automatic filters to curate higher-quality synthetic data. We organize this section according to the downstream training objective enabled by the curated data: single high-quality trajectories for supervised fine-tuning (Section \ref{sec:finetuning}), paired or ranked trajectories for preference optimization (Section \ref{sec:preference_data}), and trajectory-level or step-level state–action–reward data for offline reinforcement learning (Section \ref{sec:offline_rl}). This organization emphasizes how RMs transform raw synthetic generations into distinct forms of supervision for improving reasoning models. We list brief summaries of related works in Table \ref{tab:rm-finetuning-data-generation}, \ref{tab:rm-preference-data-generation} and \ref{tab:rm-offline-rl-data-generation}.

\subsection{Finetuning Data Generation}
\label{sec:finetuning}

\textbf{ORMs for data filtering}. One common usage of ORM is to select high-quality data for LLM finetuning. 
This approach is widely adopted in language model alignment tasks. For example, RAFT~\citep{dong2023raft} filters high-quality data by RM scores and uses them for SFT training, while 
RRHF~\citep{yuan2023rrhf} instead aligns LM with a proposed ranking loss.
For reasoning-related tasks, most methods like STaR~\citep{zelikman2022star}, RFT~\citep{yuan2023rft}, V-Star~\citep{hosseini2024vstar} and STILL-2~\citep{min2024imitateexploreselfimprovereproduction} leverage ground truth values for data filtering and self-iteration.

Recent works further extend this idea to RM-based filtering when explicit labels are limited.
MathSE~\citep{chen2025mathse} iteratively finetunes multimodal math models with correct reasoning paths from the previous round and reflections from a specialized ORM.
However, the Quality-Utility Paradox~\citep{qian2026qualityutilityparadox} shows that higher RM scores do not always imply better finetuning data, since traces refined by a stronger model can be less useful for small models than traces closer to the student's own style.

\textbf{PRMs for data generation}. Data generated using ORMs may result in reasoning trajectories with correct final answers but wrong intermediate steps. PRMs can mitigate this limitation. 
For example,
REST-MCTS*~\citep{zhang2024restmcts} and RStar-Math~\citep{guan2025rstarmath} use MCTS with the guidance of PRM to improve data quality. 
GroundedPRM~\citep{zhang2025groundedprm} constructs reasoning paths with MCTS and verifies intermediate steps with external tools to obtain automatic process supervision.
RPM-MCTS~\citep{lin2025rpmmcts} uses knowledge retrieval as a PRM within MCTS to score intermediate code steps, correct errors, and build data for full finetuning.
StepPRM-RTL~\citep{vijayaraghavan2026stepprmrtl} builds stepwise RTL reasoning trajectories, scores intermediate steps with a PRM, and uses MCTS to enrich high-quality finetuning data.
AutoMathCritique~\citep{xi2024automathcritique} generates critique data to fine-tune a critic model incorporating step-level feedback from annotator models such as LLM-as-a-judge. The critique model can be further utilized for training-time or test-time self-improvement.
Beyond math and code, WebSTAR~\citep{he2026webstar} converts noisy computer-use rollouts into SFT data through step-level filtering.
AutoTraj~\citep{gong2026autotraj} retains high-quality tool-use trajectories, repairs low-quality ones, and trains a trajectory-level RM from repaired and original trajectory pairs.
SWE-TRACE~\citep{han2026swetrace} uses stepwise oracle verification to distill a token-efficient SFT corpus for software engineering agents.

At the same time, RMs themselves can be iteratively improved during data collection, allowing the policy model and the RM to evolve simultaneously. For example, V-Star~\citep{hosseini2024vstar} updates its ORM via DPO using both correct and incorrect trajectories. 
REST-MCTS*~\citep{zhang2024restmcts} uses the per-step value in the tree as the value target for PRM training. 
To address the high variance in value estimation in MCTS, RStar-Math~\citep{guan2025rstarmath} selects trajectories with correct and incorrect final answers to construct preference pairs and refine the PRM. 
SER~\citep{huang2025ser} demonstrates that RMs can generate their own training data and iteratively improve through self-labelling. Training data is first annotated by the RM, then high-confidence self-labeled samples are selected for retraining. The refined RM can subsequently facilitate more effective RL training.
StepORLM~\citep{zhou2025steporlm} co-evolves a policy model and a generative PRM by combining solver-based outcome checks with process-level feedback.

\begin{table}[!t]
\centering
\caption{Summary of RM-based finetuning data generation methods.}
\label{tab:rm-finetuning-data-generation}
\tiny
\setlength{\tabcolsep}{2pt}
\renewcommand{\arraystretch}{0.84}
\begin{tabular}{@{}p{0.24\linewidth}p{0.16\linewidth}p{0.24\linewidth}p{0.26\linewidth}@{}}
\toprule
\textbf{RM / work} & \textbf{Domain} & \textbf{Data construction method} & \textbf{Core description} \\
\midrule
\multicolumn{4}{l}{\textit{Outcome-guided filtering and self-training}} \\
\midrule
RAFT~\citep{dong2023raft} & Alignment & RM-ranked SFT data & Keeps high-reward outputs for finetuning. \\
RRHF~\citep{yuan2023rrhf} & Alignment & Ranked-response loss & Learns from RM-ordered response lists. \\
STaR / RFT / STILL-2~\citep{zelikman2022star,yuan2023rft,min2024imitateexploreselfimprovereproduction} & Math & Answer-verified self-training & Iteratively retains correct generated rationales. \\
V-STaR~\citep{hosseini2024vstar} & Math & Verifier-filtered trajectories & Trains a verifier from correct and incorrect traces. \\
MathSE~\citep{chen2025mathse} & Multimodal math & ORM reflection loop & Uses correct paths and ORM reflections across rounds. \\
Quality-Utility Paradox~\citep{qian2026qualityutilityparadox} & Math distillation & Reward/style analysis & Shows high RM scores may hurt small-model utility. \\
SER~\citep{huang2025ser} & Reward learning & RM self-label filtering & Retrains RMs on high-confidence self-labeled data. \\
\midrule
\multicolumn{4}{l}{\textit{Process-guided trajectory construction}} \\
\midrule
ReST-MCTS$^*$~\citep{zhang2024restmcts} & Math reasoning & PRM-guided MCTS & Builds SFT data from high-value tree paths. \\
Li et al.~\citep{li2025mctsgeneration} & Math reasoning & MCTS process scores & Scores sampled steps and trains on weighted traces. \\
rStar-Math~\citep{guan2025rstarmath} & Math reasoning & MCTS preference data & Uses correct/incorrect paths to refine policy and PRM. \\
GroundedPRM~\citep{zhang2025groundedprm} & Math reasoning & Tool-grounded step checks & Verifies tree steps with external tools for process labels. \\
RPM-MCTS~\citep{lin2025rpmmcts} & Code & Retrieval PRM + execution & Scores code steps via retrieval and repairs by feedback. \\
StepPRM-RTL~\citep{vijayaraghavan2026stepprmrtl} & RTL code & PRM-guided MCTS trajectories & Enriches stepwise hardware-code finetuning data. \\
AutoMathCritique~\citep{xi2024automathcritique} & Math & Critique data synthesis & Distills judge feedback into a step-level critic. \\
WebSTAR~\citep{he2026webstar} & Web agents & Step-level rollout filtering & Converts noisy CUA rollouts into graded SFT steps. \\
AutoTraj~\citep{gong2026autotraj} & Tool reasoning & Trajectory repair + RM pairs & Repairs weak tool-use traces and trains a trajectory RM. \\
SWE-TRACE~\citep{han2026swetrace} & SWE agents & Oracle/rubric step filtering & Distills short-path SFT data and rubric PRM feedback. \\
StepORLM~\citep{zhou2025steporlm} & Operations research & Solver + GenPRM co-evolution & Co-trains policy and generative PRM with dual feedback. \\
\bottomrule
\end{tabular}
\end{table}

\subsection{Preference Data Generation}
\label{sec:preference_data}

Preference data with pairs of positive and negative samples can be used to align model outputs with human values and priorities (e.g., helpfulness, safety, coherence) by training models to distinguish and generate preferred responses over alternatives. Additionally, preference-based training can also enhance reasoning by refining the logical flow, relevance, and reliability of outputs, ensuring reasoned conclusions align with the facts.

\textbf{Preference data in outcome level}. 
Traditionally, outcome-level preference data relies on human annotations. 
Due to the substantial cost of human labeling~\citep{lee2024rlaifvsrlhf} and inherent variability among annotators~\citep{poddar2024personalizingrlhf}, scalable and automated solutions are desirable. 
An intuitive way for scaling preference data is the use of reward models.
Yuan \emph{et al.}~\citep{yuan2025selfrewardinglanguagemodels} use self-generated rewards from the LLM to obtain preference pairs and optimize the same LLM iteratively. 
Pang \emph{et al.}~\citep{pang2024irpo} extend this framework to reasoning tasks by generating CoT solutions.
LongReward~\citep{zhang2024longreward} uses LLM to provide reward on four human-designed dimensions: helpfulness, logicality, faithfulness, and completeness to enhance the reward reliability and effectively improve long-context SFT models.
DPPrefSyn~\citep{gao2026dpprefsyn} learns private preference models from human preference data and then synthesizes new preference pairs on public prompts under differential privacy guarantees.
SAO~\citep{yin2025sao} removes external annotators by letting the target LLM generate prompts, responses, and self-evaluated preferences for its own alignment.
EvoLM~\citep{li2026evolm} trains a rubric generator and a policy together, using rubric-conditioned scores on outputs from different policy checkpoints as self-generated preference signals.
Differently, Jiao \emph{et al.}~\citep{jiao2024processrewardsysthesizing} rank complete trajectories by their accumulated PRM scores. 
STILL-1~\citep{jiang2024still1} employs an ORM to select high-quality preference pairs for iterative DPO, further enhancing the ORM through active learning. Similarly, Tu \emph{et al.}~\citep{tu2025iterativedpo} perform multiple DPO rounds, but instead improve the PRM with annotations from a stronger model.
For general instruction-following, CoT-Self-Instruct~\citep{yu2025cotselfinstruct} focuses on generating high-quality instructions and constructs preference pairs for DPO by sampling $K$ responses for each candidate instruction, and using an RM to score them, taking the minimum as the instruction-level quality.
MADPO~\citep{rho2026madpo} trains an RM to estimate pairwise preference margins and uses these margins to weight DPO samples, so hard pairs receive stronger learning signals while easy pairs are still preserved.

\textbf{Preference data at the step level}.
Conventional DPO based on outcome-level preferences may be insufficient for multi-step reasoning, as it overlooks the credit assignment on individual steps~\citep{lai2024stepdpo, wang2024oreo}.
The emergence of PRMs enables the construction of step-level preference pairs.
Building on this, MCTS has become a common approach for assessing step-wise quality, as Q-values for each step node can be automatically learned or estimated using PRMs.
For instance, MCTS-DPO~\citep{xie2024mctsdpo} collects step-level DPO preference data via MCTS by selecting high or low Q-value nodes, where the Q-value is based on self-evaluation rewards. 
AlphaMath~\citep{chen2024alphamath} trains a value model as PRM from Q-values in MCTS and uses it to filter data for joint iterative training of value and policy models. 
Other approaches annotate step correctness with generative models. 
For example, Step-DPO~\citep{lai2024stepdpo} generates a large amount of step-wise chosen-reject pairs using the discriminative capability of stronger LLMs. 
Full-Step-DPO~\citep{xu2025fullstepdpo} further extends this by training with a step-wise DPO loss. Similarly, Zhang \emph{et al.}~\citep{zhang2025processbasedselfrewardinglanguagemodels} generate long CoT data, score steps using a step-wise LLM-as-a-judge, and optimize with step-wise DPO. Step-KTO~\citep{lin2025stepkto} optimizes the model by a step-wise KTO loss with both process and outcome feedback.
Hard Negative DPO~\citep{lu2026hardnegativesample} uses a compact math verifier to find near-correct but flawed reasoning traces and assigns higher weights to the most informative DPO pairs.

Recent agent-oriented works further extend step-level preference data from math reasoning to long-horizon decision making.
TGPO~\citep{chen2025tgpo} trains web agents with a PRM that assigns fine-grained rewards from subgoal progress, redundancy detection, and action verification.
CSO~\citep{li2026cso} uses a PRM to locate candidate critical steps in failed agent trajectories, verifies better alternatives by continued rollout, and uses the verified alternatives as DPO data.
CausalFlow~\citep{bonagiri2026causalflow} identifies failure-inducing steps by counterfactual intervention and converts them into validated wrong-corrected step pairs for offline preference optimization or RM training.
RoRo~\citep{ye2026roro} constructs routing preference pairs from outcome, cost, and process quality, then trains a rubric generator and judge to provide process rewards for stepwise model routing.
OrchRM~\citep{tsang2026orchrm} uses intermediate artifacts from multi-agent executions to construct win-lose pairs for Bradley-Terry RM training, enabling reward-guided orchestration without human annotations.

\begin{table}[!t]
\centering
\caption{Summary of RM-based preference data generation methods.}
\label{tab:rm-preference-data-generation}
\tiny
\setlength{\tabcolsep}{2pt}
\renewcommand{\arraystretch}{0.82}
\begin{tabular}{@{}p{0.24\linewidth}p{0.16\linewidth}p{0.24\linewidth}p{0.26\linewidth}@{}}
\toprule
\textbf{RM / work} & \textbf{Domain} & \textbf{Data construction method} & \textbf{Core description} \\
\midrule
\multicolumn{4}{l}{\textit{Outcome-guided preference construction}} \\
\midrule
Self-Rewarding LM~\citep{yuan2025selfrewardinglanguagemodels} & General alignment & Self-rewarded pairs & Uses the model's own reward judgments for iterative DPO. \\
IRPO~\citep{pang2024irpo} & Reasoning & CoT preference iteration & Extends self-rewarding pairs to reasoning traces. \\
LongReward~\citep{zhang2024longreward} & Long context & Multi-aspect judge scores & Scores helpfulness, logic, faithfulness, and completeness. \\
DPPrefSyn~\citep{gao2026dpprefsyn} & Private alignment & DP preference synthesis & Learns private preference models and samples public-prompt pairs. \\
SAO~\citep{yin2025sao} & General alignment & Self-synthetic preferences & Generates prompts, responses, and preferences internally. \\
EvoLM~\citep{li2026evolm} & General alignment & Rubric-conditioned contrasts & Uses checkpoint contrasts and co-evolved rubric scores. \\
STILL-1 / STILL-2~\citep{jiang2024still1,min2024imitateexploreselfimprovereproduction} & Math reasoning & ORM-selected DPO pairs & Filters tree/self-iteration outputs for preference training. \\
Tu et al.~\citep{tu2025iterativedpo} & Math reasoning & Iterative DPO annotation & Repeats DPO while improving the PRM with stronger labels. \\
CoT-Self-Instruct~\citep{yu2025cotselfinstruct} & Instruction following & RM-scored response sets & Scores $K$ responses to derive instruction-level pair quality. \\
MADPO~\citep{rho2026madpo} & General alignment & RM margin weighting & Weights DPO pairs by predicted preference margins. \\
\midrule
\multicolumn{4}{l}{\textit{Process-guided preference construction}} \\
\midrule
Jiao et al.~\citep{jiao2024processrewardsysthesizing} & Planning/math & Accumulated PRM ranking & Converts process scores into trajectory preferences. \\
MCTS-DPO~\citep{xie2024mctsdpo} & Math reasoning & Step Q-value pairs & Chooses high/low MCTS nodes for DPO. \\
AlphaMath~\citep{chen2024alphamath} & Math reasoning & Value-model filtering & Learns PRM-like values from MCTS Q-values. \\
Step-DPO~\citep{lai2024stepdpo} & Math reasoning & LLM-labeled step pairs & Builds chosen--rejected pairs at each reasoning step. \\
Full-Step-DPO~\citep{xu2025fullstepdpo} & Math reasoning & Step-wise DPO loss & Optimizes preferences over full step sequences. \\
Process-based SRLM~\citep{zhang2025processbasedselfrewardinglanguagemodels} & Long CoT & Step-judge preference data & Scores long-CoT steps with an LLM-as-judge. \\
Step-KTO~\citep{lin2025stepkto} & Math reasoning & Process/outcome binary labels & Applies KTO to step-level and final feedback. \\
Hard Negative DPO~\citep{lu2026hardnegativesample} & Math reasoning & Verifier-mined hard pairs & Upweights near-correct but flawed traces. \\
TGPO~\citep{chen2025tgpo} & Web agents & PRM-labeled tree decisions & Rewards subgoal progress, redundancy, and action validity. \\
CSO~\citep{li2026cso} & Agents & Verified critical-step pairs & Replaces failed critical steps with rollout-verified alternatives. \\
CausalFlow~\citep{bonagiri2026causalflow} & Agent tasks & Counterfactual repair pairs & Turns causal failure steps into wrong--corrected pairs. \\
RoRo~\citep{ye2026roro} & Model routing & Rubric routing preferences & Uses outcome, cost, and process quality for routing pairs. \\
OrchRM~\citep{tsang2026orchrm} & Multi-agent systems & Artifact win--lose pairs & Trains orchestration RMs from intermediate artifacts. \\
ARCO~\citep{tian2026arco} & Multi-step agents & Co-evolved step rubrics & Generates per-step criteria and rubric-conditioned rewards. \\
\bottomrule
\end{tabular}
\end{table}

\subsection{Offline RL Data Generation}
\label{sec:offline_rl}

Offline RL trains LLM policies using pre-collected interaction data, eliminating the need for further online interaction during training. Such datasets typically consist of fixed trajectories equipped with reward, value, or advantage labels for direct policy optimization, whereas preference data are based on pairwise comparisons or rankings of model outputs. In this setting, RMs serve as automatic teachers that score either complete reasoning trajectories or intermediate reasoning steps, making the fixed data usable for policy improvement. Existing methods can be broadly divided into reward-filtered trajectory selection, step- or advantage-labeled trajectory construction, and value/Q-function-based offline RL.

\textbf{Trajectory selection.} This line first samples complete trajectories and then uses an RM to select or emphasize trajectories with high estimated quality. ReST \citep{gulcehre2023rest} leverages a learned RM to filter high-quality data for further multi-turn offline RL training, iteratively aligning the policy with human preferences. PRS \citep{ye2024prs} constructs offline RL datasets by iteratively sampling responses via a tree-based framework integrated with an RM, then performs offline RL by repeatedly training on the highest-reward samples.

\textbf{Trajectory construction.} This line decomposes reasoning trajectories into smaller units and uses RMs, PRMs, or RM-like critics to provide more fine-grained learning signals. SWiRL \citep{goldie2025swirl} synthesizes multi-step reasoning trajectories, evaluates the quality of each step with a generative RM to filter low-quality segments, and then applies step-wise RL optimization on the curated sub-trajectories. DAPO \citep{liu2024dapo} trains a critic to estimate the advantage of each reasoning step, constructing offline datasets with state–action–advantage tuples to optimize policy performance. SPARE \citep{rizvi2026spare} aligns generated reasoning steps with reference solutions to produce automatic step-level reward annotations, which are used for both PRM training and offline RL finetuning. BCPG-NSA \citep{yang2025BCPG-NSA} segments long-CoT trajectories, uses both LLM and PRM judges to identify correct steps inside negative samples, and then performs behavior-constrained offline RL with these mined useful steps.

\textbf{Value/Q-function-based offline RL.} Instead of only filtering trajectories, this line learns value or Q-functions that score partial reasoning states or actions, which can be viewed as PRM-like dense reward estimators for policy optimization. OREO \citep{wang2024oreo} proposes a soft Q-learning algorithm under a maximum-entropy RL framework that jointly learns a policy network and an explicit value model, where the trained value model may work similarly as a PRM to generate data for further iterative training. DQO \citep{ji2025dqo} formulates chain-of-thought generation as a Markov decision process and directly optimizes an LM-parameterized Q-function with a soft actor-critic objective, enabling offline RL to exploit process-level reward signals for multi-step reasoning.

\begin{table}[!t]
\centering
\caption{Summary of RM-based offline RL data generation methods.}
\label{tab:rm-offline-rl-data-generation}
\tiny
\setlength{\tabcolsep}{2pt}
\renewcommand{\arraystretch}{0.88}
\begin{tabular}{@{}p{0.24\linewidth}p{0.16\linewidth}p{0.24\linewidth}p{0.26\linewidth}@{}}
\toprule
\textbf{RM / work} & \textbf{Domain} & \textbf{Data construction method} & \textbf{Core description} \\
\midrule
\multicolumn{4}{l}{\textit{Outcome-guided trajectory selection}} \\
\midrule
ReST~\citep{gulcehre2023rest} & General LM & RM-filtered rollouts & Iteratively trains on high-reward sampled trajectories. \\
PRS~\citep{ye2024prs} & Preference alignment & Reward-guided tree sampling & Builds offline RL data from highest-reward reflective samples. \\
\midrule
\multicolumn{4}{l}{\textit{Process-guided trajectory construction}} \\
\midrule
SWiRL~\citep{goldie2025swirl} & Reasoning/tool use & Step sub-trajectory filtering & Filters and optimizes decomposed multi-step rollouts. \\
DAPO~\citep{liu2024dapo} & Math/code & Step advantage tuples & Uses a critic to label state--action--advantage data. \\
SPARE~\citep{rizvi2026spare} & Reasoning & Reference-guided step labels & Aligns steps to references for PRM and offline RL data. \\
BCPG-NSA~\citep{yang2025BCPG-NSA} & Math reasoning & Negative-sample step mining & Extracts useful steps from flawed long-CoT trajectories. \\
\midrule
\multicolumn{4}{l}{\textit{Value/Q-function-guided offline RL}} \\
\midrule
OREO~\citep{wang2024oreo} & Multi-step reasoning & Soft value-function data & Learns policy and value model for dense credit assignment. \\
DQO~\citep{ji2025dqo} & Math reasoning & LM-parameterized Q-function & Optimizes CoT generation as an offline MDP. \\
\bottomrule
\end{tabular}
\end{table}
\section{RM Application 3: Online Reinforcement Learning}
\label{sec:reinforcementlearning}

Online reinforcement learning has been widely adopted to elicit the reasoning ability of LLMs.
By enabling LLMs to autonomously explore potential reasoning paths and receive feedback via reward models, online RL guides policy models toward desired behaviors without relying on offline datasets.
Early applications of online RL primarily focused on alignment techniques such as RLHF to enhance instruction-following and ensure the safety and consistency of model outputs. More recently, large-scale online RL has been applied to multi-step reasoning tasks. When trained on long CoT data, LLMs demonstrate strong reasoning capabilities and achieve remarkable performance across a variety of tasks. 

Reward signals are crucial in this optimization framework, providing external feedback to guide the refinement of the policy model.
Iterative optimization techniques are commonly employed to maximize cumulative rewards while avoiding excessive deviation from the model's initial capabilities. 
Verifiable rewards are frequently adopted in online RL, which rely on objective criteria or external ground-truth. For instance, in code generation tasks, rewards typically depend on whether the produced code passes predefined unit tests. Such rewards are transparent and reliable because they are based on measurable outcomes rather than subjective judgments. However, their applicability is restricted to tasks with clearly defined verification procedures. Despite considerable achievements obtained through verifiable or rule-based rewards~\citep{lambert2025tulu3, deepseekai2025deepseekr1, mistralai2025magistral}, we primarily focus on the role and impact of \textit{parametric} reward RMs. We structure this section around the key components required to understand RM-based online RL. Section \ref{sec:llm_as_rl} formulates LLM post-training as an RL problem and clarifies where RM rewards enter the objective. Section \ref{sec:common_rl} summarizes common RL algorithms, since different optimization algorithms convert the same RM signal into policy updates in different ways. Section \ref{sec:rewards_online_rl} discusses outcome-level and process-level rewards, which constitute the main design axis of RMs in online RL. Finally, Section \ref{sec:rewardhacking} examines reward hacking, a central failure mode that arises when a policy is optimized against an imperfect RM. We list brief summaries of related works in Table \ref{tab:rm-online-rl-outcome} and \ref{tab:rm-online-rl-process-hybrid}.

\subsection{Formulating LLM Post-training as an RL Problem}
\label{sec:llm_as_rl}

To connect reward models with RL optimization, we formulate LLM post-training as a sequential decision-making problem.
Given an input prompt $x$, the policy model $\pi_\theta$ generates a response $y=(y_1,\dots,y_T)$ token by token.
At decoding step $t$, the state is the prompt together with the generated prefix, i.e., $s_t=(x, y_{<t})$, and the action is the next token $a_t=y_t$.
The generated response therefore constitutes a trajectory $\tau=(s_1,a_1,\dots,s_T,a_T)$ sampled from the policy.

The RL objective is to maximize the expected return of generated trajectories:
\[
J(\theta)=\mathbb{E}_{y\sim\pi_\theta(\cdot|x)}\left[\sum_{t=1}^{T} r_t\right].
\]
In LLM post-training, this objective is often regularized to prevent the policy from drifting excessively away from the reference model:
\[
J(\theta)=\mathbb{E}_{y\sim\pi_\theta(\cdot|x)}\left[\sum_{t=1}^{T} r_t\right]
-\beta\,\mathrm{KL}\!\left(\pi_\theta(\cdot|x)\,\|\,\pi_{\mathrm{ref}}(\cdot|x)\right).
\]
Different online RL algorithms, such as REINFORCE, PPO, and GRPO, can be viewed as different strategies for optimizing this objective, mainly differing in how they estimate advantages, control policy updates, and trade off stability against efficiency.

Under this formulation, reward models define the reward function used by RL.
An ORM provides a terminal reward for the final response,
\[
r_t=0 \ \text{for } t<T,\qquad r_T = R_{\mathrm{out}}(x,y),
\]
which corresponds to sparse outcome-level supervision.
By contrast, a PRM provides intermediate rewards for partial reasoning steps or token prefixes,
\[
r_t = R_{\mathrm{proc}}(x, y_{\le t}),
\]
or more generally step-level rewards over a decomposed reasoning trajectory.
Therefore, the difference between ORM and PRM is not only the granularity of evaluation, but also the way rewards are assigned along the trajectory, which directly affects credit assignment in RL optimization.

\subsection{Common RL Algorithms}
\label{sec:common_rl}

Given the above formulation, commonly used online RL algorithms for LLM post-training can be understood as different methods for optimizing the same reward-maximization objective, while differing in their advantage estimation, regularization, and update constraints. 
Table~\ref{tab:online_rl_algorithms} summarizes the main objectives or update rules of representative algorithms.
According to whether they use a value model, group-based relative rewards, or sequence-level likelihood ratios, these algorithms can be roughly grouped into four categories:

1) \textbf{REINFORCE-style policy-gradient methods.}
\textit{REINFORCE}~\citep{sutton1999policygradient} is a foundational policy-gradient method that directly maximizes expected cumulative rewards via gradient ascent.
Although conceptually straightforward, REINFORCE suffers from high variance in gradient estimations, causing instability when applied to tasks such as LLM alignment.
Recent modifications, such as \textit{RLOO}~\citep{ahmadian2024rloo}, improve upon this by employing the mean reward of other responses from identical prompts as a leave-one-out baseline, reducing variance in advantage estimation without introducing an additional value model.

2) \textbf{Value-model-based PPO methods.}
\textit{Proximal Policy Optimization (PPO)}~\citep{schulman2017ppo} addresses training instability by using a clipped surrogate objective alongside a value-function baseline.
Despite improved stability, PPO necessitates maintaining four distinct models during training, including policy, reference, reward, and value, which imposes significant computational overhead.
\textit{VAPO}~\citep{yue2025vapo} further revisits value-model-based RL for long-CoT reasoning.
It improves PPO by using value pretraining, decoupled and length-adaptive GAE, and token-level loss normalization, showing that a well-trained value model can provide more fine-grained and stable advantage estimates for long reasoning trajectories.

3) \textbf{Critic-free group-relative methods.}
\textit{Group Relative Policy Optimization (GRPO)}~\citep{shao2024grpodeepseekmath} eliminates the value model required by PPO, adopting group-based relative advantage estimation from multiple responses to the same prompt.
This design significantly reduces memory usage and has become widely used in RL training for mathematical reasoning.
Subsequently, \textit{DAPO}~\citep{yu2025dapo} resolves several critical limitations of GRPO, including entropy collapse, inefficient gradient utilization, long-sequence biases, and reward noise.
It introduces techniques such as dynamic sampling, asymmetric clipping, and token-level loss normalization, thereby improving reasoning diversity and training efficiency.

4) \textbf{Sequence-level group optimization.}
Although \textit{GSPO}~\citep{zheng2025gspo} also uses group-based advantage estimation, its key distinction is that it changes the optimization unit from tokens to entire responses.
Instead of applying importance ratios and clipping at each token position as in GRPO, GSPO defines a sequence-level likelihood ratio and performs sequence-level clipping.
This better matches the common setting where rewards are assigned to complete responses, reduces token-level noise, and improves training stability, especially for long responses and large-scale LLM post-training.

\begin{table}[t]
\centering
\footnotesize
\setlength{\tabcolsep}{3pt}
\renewcommand{\arraystretch}{2.5}
\caption{Representative online RL algorithms for LLM post-training.
Here $x$ denotes a prompt, $y_i=(y_{i,1},\ldots,y_{i,|y_i|})$ denotes the $i$-th sampled response, $G$ is the group size, and $T$ is the response length.
$R_i=R(x,y_i)$ is the reward, $b(x)$ is a baseline, $A_t$ is a token-level advantage, and $\hat A_i=(R_i-\mu_R)/\sigma_R$ is the group-normalized advantage, where $\mu_R$ and $\sigma_R$ are computed over responses to the same prompt.
$r_t$ or $r_{i,t}$ denotes the token-level probability ratio
$\pi_\theta(y_{i,t}|x,y_{i,<t})/\pi_{\theta_{\rm old}}(y_{i,t}|x,y_{i,<t})$.
$\operatorname{clip}(z,l,u)$ clips $z$ into $[l,u]$; $\epsilon$, $\epsilon_{\rm low}$, and $\epsilon_{\rm high}$ are clipping bounds; $\beta$ controls KL regularization to the reference policy $\pi_{\rm ref}$.
In GRPO,$\widehat D_{{\rm KL},i,t}=q_{i,t}-\log q_{i,t}-1$, where$q_{i,t}=\pi_{\rm ref}(y_{i,t}|x,y_{i,<t})/\pi_\theta(y_{i,t}|x,y_{i,<t})$.
In VAPO, $\hat A^{\rm LA\text{-}GAE}_{i,t}$ denotes length-adaptive GAE and $\mathcal{L}_{\rm NLL}^{+}$ is the positive-sample NLL loss with weight $\mu$.
In GSPO, $s_i$ denotes the sequence-level probability ratio.}
\label{tab:online_rl_algorithms}
\begin{tabularx}{\textwidth}{l X}
\toprule
\textbf{Algorithm} & \textbf{Main objective or update} \\
\midrule
REINFORCE~\citep{sutton1999policygradient}
& $\nabla_\theta J(\theta)=
\mathbb{E}_{x,\,y\sim\pi_\theta}
\!\left[
\bigl(R(x,y)-b(x)\bigr)
\sum_{t=1}^{|y|}
\nabla_\theta \log \pi_\theta(y_t|x,y_{<t})
\right]$ \\

RLOO~\citep{ahmadian2024rloo}
& $\nabla_\theta J(\theta)=
\mathbb{E}_{x,\,\{y_i\}_{i=1}^{G}\sim\pi_\theta}
\!\left[
\frac{1}{G}\sum_{i=1}^{G}
\left(R_i-\frac{1}{G-1}\sum_{j\ne i}R_j\right)
\sum_{t=1}^{|y_i|}
\nabla_\theta\log\pi_\theta(y_{i,t}|x,y_{i,<t})
\right]$ \\

PPO~\citep{schulman2017ppo}
& $\mathbb{E}\!\left[\frac{1}{T}\sum_t
\min(r_t A_t,\operatorname{clip}(r_t,1-\epsilon,1+\epsilon)A_t)\right]
-\beta D_{\rm KL}(\pi_\theta\|\pi_{\rm ref})$ \\

GRPO~\citep{shao2024grpodeepseekmath}
& $\mathbb{E}\!\left[
\frac{1}{G}\sum_i\frac{1}{|y_i|}\sum_t
\left(
\min\left(
r_{i,t}\hat A_i,\,
\operatorname{clip}(r_{i,t},1-\epsilon,1+\epsilon)\hat A_i
\right)
-\beta\widehat D_{{\rm KL},i,t}
\right)
\right]$ \\

DAPO~\citep{yu2025dapo}
& $\frac{1}{\sum_i |y_i|}\sum_{i,t}
\min(r_{i,t}\hat A_i,
\operatorname{clip}(r_{i,t},1-\epsilon_{\rm low},1+\epsilon_{\rm high})\hat A_i)$ \\

VAPO~\citep{yue2025vapo}
& $\frac{1}{\sum_i |y_i|}\sum_{i,t}
\min\left(
r_{i,t}\hat A^{\rm LA\text{-}GAE}_{i,t},
\operatorname{clip}
(r_{i,t},1-\epsilon_{\rm low},1+\epsilon_{\rm high})
\hat A^{\rm LA\text{-}GAE}_{i,t}
\right)
-\mu\mathcal{L}_{\rm NLL}^{+}$ \\

GSPO~\citep{zheng2025gspo}
& $\mathbb{E}\!\left[\frac{1}{G}\sum_i
\min(s_i\hat A_i,\operatorname{clip}(s_i,1-\epsilon,1+\epsilon)\hat A_i)\right]$,
$s_i=\exp\!\left(\frac{1}{|y_i|}\sum_t\log r_{i,t}\right)$ \\

\bottomrule
\end{tabularx}
\end{table}

\subsection{Rewards in Online RL}
\label{sec:rewards_online_rl}

During RL training, RMs can provide feedback at the \textbf{outcome level} or the \textbf{process level}.
Outcome-level RMs score the final response, while process-level RMs score intermediate reasoning steps, tool calls, or dialogue turns.
This distinction determines whether the policy receives a sparse terminal signal or denser feedback for credit assignment.

\textbf{Outcome-level RL} utilizes a reward signal that evaluates the final response of the LLM for RL optimization.
A standard setting is RLHF~\citep{ouyang2022rlhf}, where a reward model is trained on human preference data to assign scores to candidate outputs for PPO optimization.
Following this paradigm, other RL algorithms for LLMs~\citep{shao2024grpodeepseekmath, yu2025dapo, liu2025drgrpo, chu2025gpg, yue2025vapo, guo2025spo} mainly alter the calculation for the advantage and corresponding optimization objectives.

Beyond standard preference-based rewards, another line of work studies how to obtain outcome rewards when explicit reward labels or verifiable answers are limited.
TTRL~\citep{zuo2025ttrl} estimates rewards via majority voting from the policy LLM itself, without using ground-truth reward labels.
While these labels may not be fully accurate, they still provide meaningful reward signals that benefit training.
EVOL-RL~\citep{zhou2025evolrl} uses the majority-voted answer as the primary correctness signal and adds a novelty reward, which favors responses with semantic dissimilarity to encourage diverse solution paths, preventing diversity collapse and improving pass@1 and pass@n over TTRL.
RL\textsuperscript{V}~\citep{sareen2025rlv} extends standard RL training by jointly training the same LLM as a generative reward model using generated data, facilitating test-time reward scaling without a separate value network and improving verification abilities.

Outcome-level rewards have also been extended to broader domains and more complex evaluation criteria.
Su \emph{et al.}~\citep{su2025crossingrewardbridge} propose training a cross-domain generative RM to overcome the limitations of binary verification, thereby expanding RL applicability across diverse domains.
Chen \emph{et al.}~\citep{chen2025learningreasonfactuality} design an online RL reward for long-form factuality that balances factual precision, detail, and answer relevance via an LLM-as-a-judge.
RLVER~\citep{wang2025rlver} aims to enhance LLM emotional reasoning by having an LLM-powered simulated user that provides a verifiable emotion score after each model reply as the RL reward.
CE-RM~\citep{hu2026cerm} trains a generative RM with unified evaluation criteria and shows that it improves downstream RL practice beyond RM benchmarks.
For domain-specific generation tasks, GRRM~\citep{yang2026grrm} scores a group of translation candidates jointly and uses these relative scores in GRPO to optimize the translation policy, RLCS~\citep{li2026rlcs} trains a generative RM for creative storytelling preferences and uses it as the reward source for RL-based story generation, and Lin \emph{et al.}~\citep{lin2026betterliterarytranslation} show that an explicit RM combined with GRPO improves literary translation where DPO degrades performance.

Recent outcome-level methods further improve the reliability and adaptability of reward signals during online RL.
Rubric-ARM~\citep{xu2026rubricarm} jointly optimizes a rubric generator and a judge with RL from preference feedback, improving policy alignment in non-verifiable tasks.
EvoRubrics~\citep{ding2026evorubrics} co-evolves a policy LLM and a rubric generator so that the reward criteria remain useful as the policy improves.
Asghari \emph{et al.}~\citep{asghari2026efficientexplorationscale} incrementally updates both the RM and the LLM as choice data arrives, using the learned RM as the RL signal.
CausalRM~\citep{wang2026causalrm} learns RMs from noisy observational user feedback and improves downstream RL performance by correcting feedback noise and selection bias.
GPRL~\citep{umer2026gprl} replaces a single scalar reward with a structured preference model and carries its multiple preference dimensions into the policy update.
UARM~\citep{pan2026uarm} equips RMs with calibrated uncertainty and down-weights unreliable GRPO advantages to stabilize RL.

Distinct from outcome-based approaches, \textbf{process-level RL} employs fine-grained, dense reward signals at intermediate steps or tokens to optimize the policy model, typically delivered by a PRM.
Early process-level RL methods mainly focus on mathematical reasoning and step-level credit assignment.
For example, Wang \emph{et al.}~\citep{wang2024mathshepherd} implement step-level PPO with a PRM to train LLMs on mathematical reasoning tasks.
BackMath~\citep{zhang2025backmath} trains both forward and backward PRMs simultaneously, integrating them via PPO to enhance mathematical problem solving.
StepTool~\citep{yu2025steptool} leverages generative LLMs to generate step-level rewards for policy gradient RL, thereby improving multi-step tool use through granular feedback on tool invocation and task contribution.

Another group of methods studies how to derive smoother or more useful process rewards from online trajectories or outcome signals.
PRIME~\citep{cui2025prime} employs an online-updated implicit PRM to supply token-level dense rewards in combination with outcome-level rewards, enabling more efficient PPO optimization.
OREAL~\citep{lyu2025oreal} introduces a novel RL framework with a reward reshaping mechanism that enforces gradient consistency between positive and negative samples, while maintaining a lightweight token-level RM that estimates token-wise importance weights without an additional value network.
TDRM~\citep{zhang2025tdrm} trains smoother PRMs with temporal-difference learning and uses them to make RL updates more stable and data-efficient.

Several recent works further focus on how process rewards should be converted into policy optimization signals.
PRPO~\citep{ding2026prpo} converts PRM scores over reasoning segments into token-level advantages and aligns them with outcome advantages in critic-free policy optimization.
VPPO~\citep{liu2026vppo} uses a PRM to locate the first incorrect step, rewarding the verified prefix while penalizing only the later erroneous part.
PUM~\citep{zhou2026pum} scores reasoning prefixes by whether they increase the chance of solving the problem and applies this signal to mathematical RL.
FaithRL~\citep{nie2026faithrl} uses explicit faithfulness rewards from a PRM to reduce hallucinated intermediate steps during step-level RL.
rePIRL~\citep{wu2026repirl} learns a PRM through an inverse-RL-inspired loop that alternates policy updates and PRM updates.
uPRM~\citep{gadetsky2026uprm} trains PRMs without step labels or final-answer verification and uses the resulting model as an RL reward signal.

Process-level rewards have also been extended from single-turn reasoning to agent and multi-turn interaction settings.
AgentPRM~\citep{xi2025agentprm} extends PRMs to agent tasks by judging each action according to both progress toward the goal and its relation to other actions.
ITPO~\citep{wang2026itpo} derives turn-wise process rewards from sparse outcome signals and combines them with PPO, GRPO, or RLOO for multi-turn interaction.
HISR~\citep{lu2026hisr} uses a segment-level process RM and hindsight information to assign rewards to sub-goals in multi-turn agentic RL.
ARCO~\citep{tian2026arco} generates per-step rubrics and step rewards with a same-scale model that co-evolves with the policy on on-policy data.
ARBOR~\citep{liu2026arbor} keeps a reusable rubric buffer for search agents and adds process scores to the base reward when outcome rewards give no gradient.

In practice, process-level rewards are often combined with outcome-level or verifiable rewards to apply \textbf{hybrid reward signals} in online RL.
Empirical studies have shown that combining process-level rewards with verifiable rewards can lead to higher accuracy in mathematical tasks~\citep{cheng2025pure}, and this hybrid reward strategy has also been applied to other domains.
For instance, o1-coder~\citep{zhang2024o1coder} explores long CoT reasoning in code generation, utilizing a PRM to evaluate intermediate reasoning steps in combination with outcome rewards from test cases.
Reward-SQL~\citep{zhang2025rewardsql} enhances performance on text-to-SQL tasks by applying online RL with both PRM-generated step rewards and binary outcome rewards.
Posterior-GRPO~\citep{fan2025posteriorgrpo} uses a separately trained RM for code generation and adds the process reward only when the outcome is correct.
PROF~\citep{ye2025prof} filters training samples by enforcing ORM--PRM consistency: among samples with correct final answers, it retains those with high PRM scores, whereas among samples with incorrect final answers, it retains those with low PRM scores.
PAPO~\citep{tan2026papo} combines an outcome component with a rubric-based PRM component, using the process reward only to compare reasoning quality among correct responses.

Despite their benefits, process-level feedback in RL is susceptible to \textbf{reward hacking}, where models exploit the reward system by generating excessive correct but irrelevant reasoning steps.
This can undermine RL training and significantly degrade performance, a challenge that will be examined further in Section~\ref{sec:rewardhacking}.

\begin{table}[!t]
\centering
\caption{Summary of outcome-level RM signals for online RL.}
\label{tab:rm-online-rl-outcome}
\tiny
\setlength{\tabcolsep}{2pt}
\renewcommand{\arraystretch}{0.88}
\begin{tabular}{@{}p{0.24\linewidth}p{0.16\linewidth}p{0.28\linewidth}p{0.24\linewidth}@{}}
\toprule
\textbf{RM / work} & \textbf{Domain} & \textbf{RL signal construction} & \textbf{Core description} \\
\midrule
\multicolumn{4}{l}{\textit{Standard and label-free outcome rewards}} \\
\midrule
RLHF reward model~\citep{ouyang2022rlhf} & Alignment & Human preference pairs train a terminal RM. & Baseline response-level reward for PPO-style RL. \\
ORM RL~\citep{shao2024grpodeepseekmath,yu2025dapo,liu2025drgrpo,chu2025gpg,yue2025vapo,guo2025spo} & Reasoning & Terminal ORM/VRM score with revised advantages. & Changes policy optimization, not reward semantics. \\
TTRL~\citep{zuo2025ttrl} & Math reasoning & Majority-vote pseudo-labels from policy samples. & Enables label-free self-improvement on unlabeled tasks. \\
EVOL-RL~\citep{zhou2025evolrl} & Math/general reasoning & Majority-vote correctness plus novelty reward. & Preserves exploration while using self-consistency. \\
RL\textsuperscript{V}~\citep{sareen2025rlv} & Math reasoning & RL-generated data trains the policy as verifier. & Co-trains reasoner and verifier for scaling. \\
\midrule
\multicolumn{4}{l}{\textit{Generative, rubric, and adaptive outcome RMs}} \\
\midrule
Reward Bridge~\citep{su2025crossingrewardbridge} & Broad QA domains & Cross-domain generative scoring from references. & Extends RLVR to less structured domains. \\
Factuality reward~\citep{chen2025learningreasonfactuality} & Long-form factuality & Judge combines precision, detail, and relevance. & Reduces hallucination without terse answers. \\
RLVER~\citep{wang2025rlver} & Empathetic dialogue & Simulated users provide deterministic emotion scores. & Turns empathy into verifiable dialogue rewards. \\
CE-RM~\citep{hu2026cerm} & Open-ended NLG & Pointwise GenRM with unified criteria and rollout. & Improves downstream RL beyond RM benchmarks. \\
GRRM~\citep{yang2026grrm} & Machine translation & Jointly scores each translation candidate group. & Uses comparative context for fine-grained ranking. \\
RLCS~\citep{li2026rlcs} & Story generation & GenRM with multi-dimensional preference reasoning. & Aligns creative-story RL with human judgments. \\
Lin \emph{et al.}~\citep{lin2026betterliterarytranslation} & Literary translation & Explicit RM over multi-aspect preference data. & GRPO with RM avoids DPO degradation. \\
Rubric-ARM~\citep{xu2026rubricarm} & Non-verifiable tasks & Alternates RL updates for rubric generator and judge. & Learns task rubrics from preference feedback. \\
EvoRubrics~\citep{ding2026evorubrics} & Open-ended tasks & Co-evolves policy and rubric generator online. & Keeps criteria discriminative as policy improves. \\
Asghari \emph{et al.}~\citep{asghari2026efficientexplorationscale} & RLHF alignment & Incrementally updated RM from streaming choices. & Improves label efficiency with guided exploration. \\
CausalRM~\citep{wang2026causalrm} & User-feedback RLHF & Noise-aware loss plus propensity reweighting. & Corrects noisy and biased observational feedback. \\
GPRL~\citep{umer2026gprl} & Open-ended alignment & Structured preference dimensions become advantages. & Avoids collapse onto one scalar reward axis. \\
UARM~\citep{pan2026uarm} & RLHF alignment & Calibrated uncertainty down-weights GRPO advantages. & Stabilizes RL under unreliable RM predictions. \\
\bottomrule
\end{tabular}
\end{table}
\begin{table}[!t]
\centering
\caption{Summary of process-level and hybrid RM signals for online RL.}
\label{tab:rm-online-rl-process-hybrid}
\tiny
\setlength{\tabcolsep}{2pt}
\renewcommand{\arraystretch}{0.88}
\begin{tabular}{@{}p{0.24\linewidth}p{0.16\linewidth}p{0.28\linewidth}p{0.24\linewidth}@{}}
\toprule
\textbf{RM / work} & \textbf{Domain} & \textbf{RL signal construction} & \textbf{Core description} \\
\midrule
\multicolumn{4}{l}{\textit{Process-level RL}} \\
\midrule
Math-Shepherd~\citep{wang2024mathshepherd} & Math reasoning & Automatic process labels train a step PRM. & Supplies step rewards for process PPO. \\
BackMath~\citep{zhang2025backmath} & Math reasoning & Forward/backward PRMs score reasoning steps. & Uses backward supervision to improve math RL. \\
StepTool~\citep{yu2025steptool} & Tool use & Rewards tool success and task contribution per step. & Optimizes multi-step tool decisions directly. \\
PRIME~\citep{cui2025prime} & Math/code & On-policy rollouts and outcomes induce implicit PRM. & Gives dense rewards without process labels. \\
OREAL~\citep{lyu2025oreal} & Math reasoning & Outcome reward reshaping plus token-importance RM. & Makes binary-reward RL less sparse. \\
TDRM~\citep{zhang2025tdrm} & Reasoning & Temporal-difference training smooths PRM scores. & Stabilizes RL with temporally consistent rewards. \\
PRPO~\citep{ding2026prpo} & Math reasoning & Segment PRM scores become outcome-aligned token advantages. & Adds critic-free fine-grained credit assignment. \\
VPPO~\citep{liu2026vppo} & Math reasoning & PRM locates first error; prefix rewarded, suffix penalized. & Uses PRM as error localizer. \\
PUM~\citep{zhou2026pum} & Math reasoning & Prefix gain trains a utility model. & Rewards prefixes by solving-probability improvement. \\
FaithRL~\citep{nie2026faithrl} & Faithful QA & PRM faithfulness rewards with truncated resampling. & Avoids rewarding hallucinated intermediate steps. \\
rePIRL~\citep{wu2026repirl} & Math/code & Inverse-RL loop alternates PRM and policy updates. & Learns PRMs with weak expert assumptions. \\
uPRM~\citep{gadetsky2026uprm} & Reasoning & Next-token probabilities score first-error candidates. & Removes step labels and final verification. \\
AgentPRM~\citep{xi2025agentprm} & Agent tasks & TD/GAE labels score action promise and progress. & Adapts PRMs to sequential agent actions. \\
ITPO~\citep{wang2026itpo} & Multi-turn interaction & Implicit turn rewards inferred from sparse outcomes. & Provides robust turn-level credit assignment. \\
HISR~\citep{lu2026hisr} & Multi-turn agents & Segment PRM modulated by hindsight likelihood ratios. & Focuses rewards on important sub-goals. \\
ARCO~\citep{tian2026arco} & Multi-step agents & Per-step rubrics and rewards co-evolve with policy. & Couples rubric generation with step scoring. \\
ARBOR~\citep{liu2026arbor} & Search agents & Reusable rubric buffer adds process scores to base reward. & Restores gradients when outcomes are tied. \\
\midrule
\multicolumn{4}{l}{\textit{Hybrid outcome--process RL}} \\
\midrule
PURE~\citep{cheng2025pure} & Math reasoning & Min-form PRM credit plus small verifiable reward mix. & Suppresses PRM reward hacking and collapse. \\
o1-Coder~\citep{zhang2024o1coder} & Code generation & PRM step scores combined with test-case outcomes. & Couples long-CoT coding with execution feedback. \\
Reward-SQL~\citep{zhang2025rewardsql} & Text-to-SQL & Execution-aware PRM plus binary outcome reward. & Guides compositional SQL reasoning step by step. \\
Posterior-GRPO~\citep{fan2025posteriorgrpo} & Code generation & Process reward gated by execution correctness. & Rewards reasoning only under correct outcomes. \\
PROF~\citep{ye2025prof} & Reasoning & Filters samples by ORM--PRM consistency. & Harmonizes correctness and reasoning quality. \\
PAPO~\citep{tan2026papo} & Math reasoning & Outcome advantage plus PRM advantage among correct answers. & Anchors correctness while ranking solution quality. \\
\bottomrule
\end{tabular}
\end{table}

\subsection{Reward Hacking}
\label{sec:rewardhacking}

Reward hacking is a serious problem when using RMs for reinforcement learning~\citep{skalse2025definingrewardhacking}. It occurs when an agent finds flaws in the reward function or task specification and exploits shortcuts that raise its reward without actually completing the intended task. This issue primarily stems from the inherent difficulty of designing an entirely accurate reward function that fits the environment perfectly. 
The concept of reward hacking was first explored in the context of traditional reinforcement learning~\citep{amodei2016aisafety, everitt2021rewardtampering, lehman2019surprisingcreativity}. As agents become more capable and general, they also become increasingly adept at discovering subtle flaws in their reward mechanisms, thereby exacerbating the problem. On the one hand, LLM training often incorporates reinforcement learning and reward modeling, inheriting the vulnerabilities of these paradigms. On the other hand, during inference, LLMs are capable of dynamically adapting their outputs through in-context learning or self-reflection, which can enable real-time exploitation of reward signals at deployment time. As a result, both the training and inference phases of LLMs are susceptible to reward hacking, underscoring the need for robust and carefully designed reward models.

\textbf{Training-stage reward hacking.} During training, reward hacking can occur in preference alignment, where the model learns to please the reward model or human evaluators rather than follow the true task goal. 
For instance, Wen \emph{et al.}~\citep{wen2024languagemodelslearnmislead} demonstrate that RLHF-optimized models can become more persuasive, leading human evaluators to accept incorrect responses more frequently, thereby increasing mistaken acceptance rates.
Likewise, if a user has already expressed a particular view, the model may choose to agree with that view instead of stating the facts, which is a form of sycophancy~\citep{denison2024sycophancy, sharma2025understandingsycophancy, rrv2024chaoskeywords, malmqvist2024sycophancysurvey}. Singhal \emph{et al.}~\citep{singhal2024longwaygo} show that RLHF-trained LLMs can game reward signals by inflating response length.  In other reasoning tasks, models can similarly fabricate evidence or introduce fictitious logical steps to support their answers and win favor with evaluators, or they may cheat on known test cases and produce obscure code to hide errors~\citep{wen2024languagemodelslearnmislead, baker2025monitoringreasoning}. 
Other studies report that RL-trained LLMs tasked with mathematical reasoning often inject many correct but unnecessary steps or have extremely few steps to exploit RMs, without improving actual answer accuracy~\citep{gao2024designingeffectiverlreward, cheng2025pure, eisenstein2024helpingherding}.

\textbf{Inference-stage reward hacking.}
Reward hacking can also occur during inference and deployment. Here, rewards may be derived from user-specified objectives or feedback collected from the environment or evaluator models across multi-turn dialogues.
Even though model parameters remain fixed, LLMs can adapt their outputs over time through context and feedback loops, a paradigm often referred to as in-context reinforcement learning. 
Pan \emph{et al.}~\citep{pan2024rewardhackingiterative} show that in self-iteration, where a generation model is repeatedly judged by an evaluation model, the evaluation score can keep rising while the true quality of the generated outputs decreases.
Pan \emph{et al.}~\citep{pan2024icrh} further observe that ambiguous goal definitions or incomplete feedback can neglect implicit constraints, causing LLMs to pursue misaligned incentives and produce undesirable side effects.

To \textbf{mitigate reward hacking}, several methods have been proposed.

1) Designing more robust reward functions. For example, combining multiple reward models can reduce over-optimization by making it more difficult for the agent to deceive all evaluators simultaneously~\citep{coste2024rmensemble, rame2024warm}, although this approach does not fully eliminate reward hacking~\citep{eisenstein2024helpingherding}. Peng \emph{et al.}~\citep{peng2025agenticrm} propose a composite reward function that mixes human preference judgments with verifiable correctness, enhancing reward reliability. Wang \emph{et al.}~\citep{wang2025rewardhackingcausalrewards} introduce causal reward modeling to reduce irrelevant biases, such as verbosity and flattery.

2) Reward shaping. Fu \emph{et al.}~\citep{fu2025rewardshaping} propose reward shaping techniques and a Preference-as-Reward method to stabilize RLHF training. They demonstrate that shaped rewards bounded and centered properly can curb the agent’s ability to exploit reward function flaws. 
In process-level RL for LLM reasoning, dedicated techniques have been introduced to mitigate reward hacking. For example, Gao \emph{et al.}~\citep{gao2024designingeffectiverlreward} introduce (1) the Clip mechanism that bounds and limits high process rewards accumulation (2) the Delta mechanism that reconstructs rewards based on the reward difference of adjacent steps to emphasize incremental progress. PURE~\citep{cheng2025pure} also mitigates reward hacking in PRM with a min-form reward shaping method by prioritizing the worst step's reward and suppressing other rewards.

3) Separating length-based rewards. Chen \emph{et al.}~\citep{chen2024odin} decouple reward signals for answer length from those for accuracy, disrupting the correlation between reward and response length. 
Shen \emph{et al.}~\citep{shen2023looselipssinkships} similarly employ two distinct reward modules to independently address length bias and semantic bias. CoLD~\citep{zheng2025cold} debiases PRMs by adding an explicit length penalty and a learned bias estimator, and jointly training the PRM with the estimator to enforce length-invariant rewards, thus reducing reward–length correlation.

4) Monitoring reward hacking. Baker \emph{et al.}~\citep{baker2025monitoringreasoning} employ a simpler monitoring model to spot flaws in the CoT and include this monitoring as part of the training objective to suppress those flaws. However, the agent may still learn to hide obfuscated reward hacking rather than eliminate it entirely.

5) Data augmentation. Srivastava \emph{et al.}~\citep{srivastava2025crome} train the reward model on augmented and controlled example pairs that change only real answer quality, teaching the reward model to favor substance and ignore superficial tricks.

\section{Analysis and Discussion}
\label{sec:analysis}

In the preceding sections, we have systematically introduced various types of RMs and analyzed their roles at different stages of LLM reasoning. In this section, we discuss five critical questions related to the selection, usage, and development of RMs. Our analysis is grounded in both conclusions from the literature and our own experimental results.

\subsection{How to Choose RMs in Different Scenarios?}
\label{sec:rmcomparison}

In this part, we compare the characteristics of different types of RMs to provide guidance on their selection for specific scenarios.

\begin{table}[t]
\centering
\small
\caption{A comparison between discriminative and generative reward models}
\label{tab:gen_dis_reward_models}
\begin{tabularx}{\textwidth}{
  >{\raggedright\arraybackslash}X
  >{\raggedright\arraybackslash}X
  >{\raggedright\arraybackslash}X
}
\toprule
\textbf{Feature} & \textbf{Discriminative RM} & \textbf{Generative RM} \\
\midrule
Output format & Scalar value & Rich text \\
\addlinespace
Typical model architecture & LM with scalar output head & Full LM with generative capabilities \\
\addlinespace
Interpretability & Low (opaque scalar score) & High (textual explanation, reasoning) \\
\addlinespace
Training cost & Generally lower & Higher (for the RL training of RM) or no cost (for using off-the-shelf LLMs) \\
\addlinespace
Inference cost & Low & High \\
\addlinespace
OOD generalization & Usually lower & Usually higher \\
\bottomrule
\end{tabularx}
\end{table}
\begin{table}[t]
  \centering
  \footnotesize
  \caption{Available direct comparisons suggest that generative RMs can outperform discriminative RMs in several reasoning-oriented inference-time settings, on both ID and OOD tasks, as independently reported by Zhang et al. \citep{zhang2024genrmcot} and Khalifa et al. \citep{khalifa2025thinkprm}. In both works, GRMs and DRMs share the same base model. Values marked with * are read from published figures.}
  \label{tab:gen_dis_rm_comparison}
  \begin{tabular*}{\textwidth}{@{\extracolsep{\fill}} l l l l c c @{}}
    \toprule
    Experiment & Policy Model & Test-time Method & Task & DRM Acc. & GRM Acc. \\
    \midrule
    \citep{zhang2024genrmcot} & Gemma-2B & BoN@32 & (ID) Algorithmic & 37.0* & \textbf{45.3} \\
                              & Gemini-1.0 Pro & BoN@16 & (ID) GSM8K & 91.0* & \textbf{93.4} \\
                              & Gemini-1.0 Pro & BoN@32 & (OOD) MATH500 & 39.0* & \textbf{44.6} \\
                              & Gemini-1.0 Pro & BoN@32 & (OOD) MMLU & 53.0 & \textbf{56.1} \\
    \midrule
    \citep{khalifa2025thinkprm} & Qwen2.5-14B & BoN@32 & (ID) MATH500 & 80.0* & \textbf{87.0}* \\
                                & Qwen2.5-32B-Inst & BoN@8 & (ID) AIME'24 & 30.0* & \textbf{33.0}* \\
                                & Llama3.2-3B-Inst & Beam Search@16 & (ID) MATH500 & 65.0* & \textbf{68.0}* \\
                                & Qwen2.5-32B-Inst & BoN@32 & (OOD) GPQA-Physics & 64.0* & \textbf{73.0}* \\
                                & Qwen2.5-Coder-7B & BoN@32 & (OOD) LiveCodeBench & 62.0* & \textbf{66.0}* \\
    \bottomrule
  \end{tabular*}
\end{table}

\textbf{Discriminative RMs vs. Generative RMs.} 
Table~\ref{tab:gen_dis_reward_models} summarizes key differences between discriminative and generative RMs. Discriminative RMs typically adopt the Bradley-Terry~\citep{bradleyterry} assumption and are favored for their efficiency during both training and inference, as they output only a single scalar reward. However, they are susceptible to over-optimization and poor generalization~\citep{hong2025robustnessrm, xiao2025infopo}. By contrast, generative RMs, which exploit the generative capabilities of LLMs, provide richer feedback and enhanced interpretability.

A central challenge for discriminative RMs is their limited generalization to out-of-distribution (OOD) scenarios, as highlighted in~\citep{wang2024secretsrlhf, lin2024limitedgeneralization, yang2024regularizinghiddenstates}, which we will discuss in \ref{sec:RMgeneralization}. Generative RMs, in contrast, can learn more robust features by generating reasoning steps and explanatory feedback~\citep{mahan2024generativerewardmodels}. As shown in Table~\ref{tab:gen_dis_rm_comparison}, shifting from a discriminative to a generative RM architecture can lead to improved performance on both in-distribution (ID) and out-of-distribution (OOD) tasks. 
Table~\ref{tab:dis_gen_processbench}
further demonstrates the advanced verification and discriminative capabilities of generative PRMs. However, the training of generative RMs remains challenging, as enhancing their reasoning ability necessitates specialized training strategies and complex data, and generating long CoT tokens incurs significant computational overhead. Also, the inference of generative RMs is relatively slower, as reported by Singhi \emph{et al.}~\citep{singhi2025solveverifycomputeoptimalproblem}, making them less efficient to use in practice.

\textbf{ORMs vs. PRMs.} 
As defined in Section~\ref{sec:granularity}, the primary distinction between ORMs and PRMs is that PRMs assign rewards at each reasoning step, whereas ORMs provide a single scalar reward for the entire reasoning process. Consequently, PRMs are applicable in a broader range of scenarios that require verification of intermediate steps, such as step-level reinforcement learning or test-time tree search. However, the acquisition of step-level annotations for PRMs is generally more costly, and step-wise reward calculation demands additional computation. Moreover, recent studies have observed that PRMs are more vulnerable to reward hacking~\citep{deepseekai2025deepseekr1, wang2025hrm, cui2025prime}. Below, we compare the empirical performance of ORMs and PRMs in two principal use cases: test-time selection and reinforcement learning.

\begin{table}[t]
  \centering
  \footnotesize
  \caption{PRMs often outperform ORMs on MATH tasks under different policy model architectures and inference strategies (BoN, weighted BoN, and beam search), as reported by Uesato et al.~\citep{uesato2022solvingmathwordproblems}, Lightman et al.~\citep{lightman2023letsverifystepstep}, Wang et al.~\citep{wang2024mathshepherd}, Setlur et al.~\citep{setlur2024rewardingprogress}, Snell et al.~\citep{snell2024scalingllmtesttimecompute}. Values marked with * are read from published figures.}
  \label{tab:orm_prm}

  \begin{tabularx}{\textwidth}{@{}>{\raggedright\arraybackslash}X >{\raggedright\arraybackslash}X >{\raggedright\arraybackslash}X l c c@{}}
    \toprule
    Experiment & Policy Model & Test-time Method & Task & ORM Acc. & PRM Acc. \\
    \midrule
    \citep{uesato2022solvingmathwordproblems}
      & Chinchilla~\citep{hoffmann2022chinchilla}
      & BoN@96
      & GSM8K
      & 85.2
      & \textbf{85.9} \\
    \citep{lightman2023letsverifystepstep}
      & GPT-4
      & BoN@1860
      & MATH
      & 72.4
      & \textbf{78.2} \\
    \citep{wang2024mathshepherd}
      & Llama2-70B
      & BoN@256
      & MATH500
      & 40.4
      & \textbf{44.5} \\
      & LLemma-34B
      & BoN@256
      & MATH500
      & 43.7
      & \textbf{46.0} \\
      & DeepSeek-67B
      & BoN@256
      & MATH500
      & 45.3
      & \textbf{47.0} \\
    \citep{setlur2024rewardingprogress}
      & Gemma-2B
      & BoN vs.\ BS@128
      & MATH
      & 20.0*
      & \textbf{28.0}* \\
      & Gemma-9B
      & BoN vs.\ BS@128
      & MATH
      & 45.0*
      & \textbf{55.0}* \\
      & Gemma-27B
      & BoN vs.\ BS@128
      & MATH
      & 53.0*
      & \textbf{57.0}* \\
    \citep{snell2024scalingllmtesttimecompute}
      & PaLM 2-S*
      & Weighted BoN@2048
      & MATH
      & 35.0*
      & \textbf{40.0}* \\
    \bottomrule
  \end{tabularx}
\end{table}

As demonstrated in Tables~\ref{tab:orm_prm} and
~\ref{tab:implictprmresults}
in Appendix, PRMs generally outperform ORMs in selecting correct solutions from a set of candidates. For instance, Uesato \emph{et al.}~\citep{uesato2022solvingmathwordproblems} report comparable final-answer error rates for ORMs and PRMs (14.8\% and 14.1\% respectively) on the GSM8K math dataset when reranking solutions, but PRMs reduce intermediate step errors from 4.4\% to 3.5\%. 
Subsequently, Lightman \emph{et al.}~\citep{lightman2023letsverifystepstep} show that PRMs (78.2\%) substantially outperform ORMs (72.4\%) in BoN sampling on the more challenging MATH benchmark. 
Other studies~\citep{wang2024mathshepherd, yuan2024implicitprm, setlur2024rewardingprogress, snell2024scalingllmtesttimecompute} further substantiate the superiority of PRMs in specific test-time guidance settings. 
Moreover, recent work~\citep{feng2025prmnecessary, zou2025reasonfluxprm} finds that when evaluating generations from advanced reasoning models such as Deepseek-R1, self-evaluative PRMs, which utilize the model's own internal reward signals for reranking, outperform external PRMs by at least 3.3\% on the final answer accuracy, which shows the great potential of self-evaluative PRM models.

Unlike test-time selection, the necessity of PRMs in reinforcement learning remains under debate. While certain studies~\citep{wang2024mathshepherd, cheng2025pure} show the benefits of using PRMs, other work indicates that outcome-level rewards alone may suffice for effective RL~\citep{deepseekai2025deepseekr1}. One explanation is that most process-level signals are closely related to outcome signals and can often be derived from them, implying limited informational gain from PRMs. For example, Cui \emph{et al.}~\citep{cui2025prime} show that implicit process rewards emerge naturally when training ORMs, and Lyu \emph{et al.}~\citep{lyu2025oreal} train token-level rewards using only outcome feedback. Theoretical analysis by Jia \emph{et al.}~\citep{jia2025needverifystepstep} indicates that, with sufficient coverage, outcome supervision can be as statistically efficient as process supervision, and any policy's advantage function can serve as an optimal process reward. Feng \emph{et al.}~\citep{feng2025prmnecessary} further argue that RL-trained reasoning models inherently develop step-evaluation capabilities. Recent evidence from PAPO~\citep{tan2026papo} provides a more nuanced picture: in their math-reasoning GRPO setup, directly optimizing PRM-only rewards leads to reward hacking and training collapse, while a naive multiplicative ORM--PRM reward yields only marginal and inconsistent improvements over ORM-only training. Nonetheless, the derivability of process feedback from outcome rewards does not necessarily obviate the need for explicit PRMs. It remains unclear whether the current shortcomings of PRMs in RL are due to fundamental limitations or merely suboptimal implementation.

\textbf{Pointwise RMs vs. Pairwise RMs.}
As defined in Section \ref{sec:pointwise_vs_pairwise}, the distinction between pointwise and pairwise RMs concerns the inference interface rather than only the training objective. A pointwise RM assigns a reusable scalar score to each candidate response independently, whereas a pairwise RM jointly compares two candidates and predicts which one is better. This distinction is important because downstream RM applications impose different requirements on reward calibration, computational cost, and score reusability.

Pointwise RMs are usually the default choice when a scalar reward must be reused across many candidates, steps, or optimization updates. In BoN selection, a pointwise RM scores $N$ sampled responses with $O(N)$ reward evaluations, after which candidates can be sorted directly. In search-guided reasoning, pointwise scores can be cached and reused for partial trajectories or tree nodes. In online RL, pointwise rewards are also more natural, since PPO-style and GRPO-style algorithms require a scalar reward for each rollout, which can be normalized, combined with KL penalties, or aggregated with process rewards. For data synthesis, pointwise RMs are similarly convenient when the goal is threshold filtering, quality weighting, or selecting a single high-quality trajectory for SFT. Recent pointwise generative RMs, such as DeepSeek-GRM~\citep{liu2025deepseekgrm}, further argue that pointwise reward generation provides flexibility for different input types and supports inference-time scaling by sampling multiple critiques or principles.

However, the main weakness of pointwise RMs is calibration. Since each response is scored independently, the RM must learn a stable absolute scale across prompts, domains, response styles, and difficulty levels. This is especially challenging in open-ended or OOD settings, where two responses may both receive high absolute scores even though one is clearly preferable. Recent theoretical work~\citep{sun2025rethinkingbradleyterrymodels} on Bradley–Terry reward modeling also suggests that, for downstream preference optimization, preserving the correct ranking or order consistency may be more important than recovering a perfectly calibrated reward value. Therefore, pointwise RMs are efficient and broadly applicable, but their scores should be interpreted primarily as relative ranking signals unless calibration has been explicitly validated.

Pairwise RMs, by contrast, are more suitable when relative comparison is easier or more reliable than absolute scoring. By conditioning on two candidates simultaneously, a pairwise RM can directly attend to subtle differences in reasoning validity, final-answer consistency, completeness, or style. PairRM~\citep{jiang2023pairrm}, for example, compares candidate outputs side by side and can be used for reranking, Best-of-$N$ decoding, and RLHF-style preference learning. In reasoning scenarios, PairJudge RM~\citep{liu2025pairjudgerm} further shows the appeal of this paradigm: instead of assigning absolute scores to math solutions, it compares two candidate solutions with chain-of-thought judgment and uses a knockout tournament for BoN selection, reporting substantial gains over baseline reward models on MATH-500 and Olympiad Bench, especially on harder problems. These results suggest that pairwise RMs can be particularly valuable when candidate solutions are close in quality and an arbitrary scalar score is difficult to trust.

The limitation of pairwise RMs is their higher inference and aggregation cost. A full ranking over $N$ candidates requires $O(N^2)$ comparisons, while tournament-style selection reduces the cost but may become sensitive to the comparison order or bracket structure. Pairwise judgments can also be non-transitive~\citep{xu2025investigatingnontransitivity}: a model may prefer $A$ over $B$, $B$ over $C$, but $C$ over $A$, making global ranking less stable. Moreover, pairwise LLM-as-a-judge methods are susceptible to position bias~\citep{shi2025positionbias}; prior work on MT-Bench~\citep{zheng2023llmasajudge} and related settings reports position, verbosity, and self-enhancement biases, while later systematic studies confirm that position bias varies across judges and tasks and is especially problematic when the quality gap between two responses is small. Thus, pairwise evaluation should normally use order swapping, randomized positions, calibrated aggregation methods such as Bradley–Terry/Elo-style ranking, or a tournament design that reduces bracket sensitivity.

Overall, pointwise and pairwise RMs should be selected according to the downstream use case. For online RL, step-level search, large-$N$ BoN, and large-scale data filtering, pointwise RMs are usually preferable because they provide efficient, reusable scalar rewards. For small-candidate reranking, hard tie-breaking, preference-pair construction, and open-ended evaluation where absolute score calibration is unreliable, pairwise RMs can be more effective.

\subsection{How Well Do RMs Generalize OOD?}
\label{sec:RMgeneralization}
The generalization ability of RMs decides their applicability in diverse real-world scenarios. With strong generalization ability, a general RM can be used in different settings effortlessly. 
However, we find that existing RMs, especially discriminative RMs, show very limited generalization ability.
Consequently, we need to either train a new RM for each new setting or tolerate the poor performance of existing RMs on downstream tasks.
Several approaches have been developed to enhance the generalization of RMs. However, the problem is still far from being solved.

\subsubsection{The Generalization Ability of Current RMs}

Existing studies indicate that due to insufficient diversity in training data, current RMs often fail to generalize OOD~\citep{zhou2025rmb, lin2024limitedgeneralization}. 
In this part, we analyze the generalization performance of RMs under three different kinds of OOD settings, namely, response OOD, question OOD, and domain OOD.

\begin{figure*}[t]
    \centering
    \includegraphics[width=1.0\textwidth]{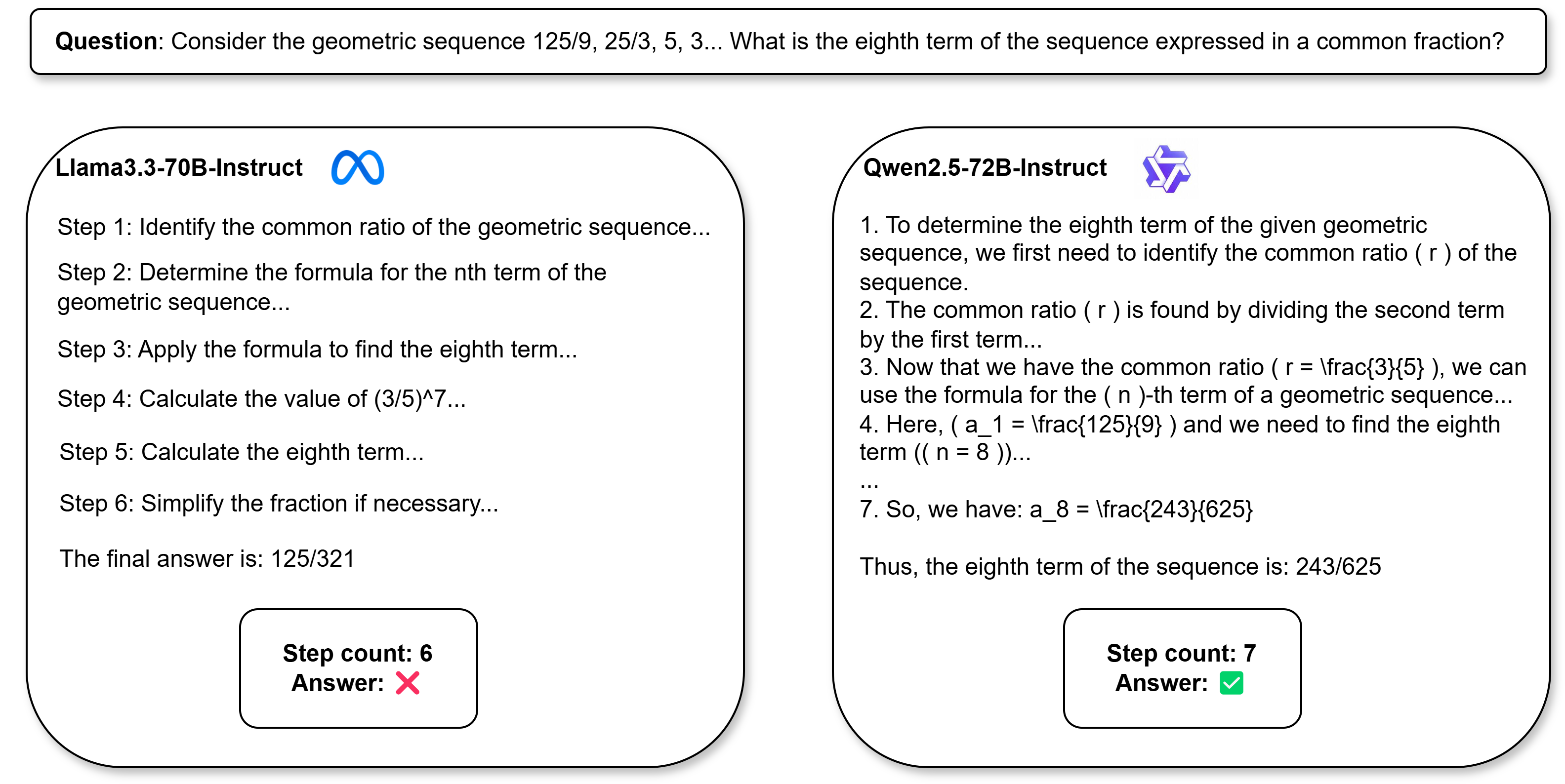}
    \caption{Comparisons of Llama and Qwen response styles in an example math question}
    \label{fig:llamaqwencomparison}
\end{figure*}

\textbf{Response OOD} arises when an RM evaluates reasoning generated by a policy model whose output style differs from the RM’s training data.
The responses generated by different language models may vary in characteristics such as length, quality, and structure. For instance, Llama models tend to generate fewer, more structured reasoning steps, whereas Qwen models prefer longer responses and exhibit some cognitive behaviors, as illustrated in Figure~\ref{fig:llamaqwencomparison}. Applying an RM trained on Qwen-style data to Llama-generated outputs can lead to OOD issues. Levine \emph{et al.}~\citep{levine2024rmdistributionshift} show that RM accuracy is highly sensitive to shifts in the distribution of LM-generated responses. In their experiments, when both prompts and responses are in-distribution, the RM achieves 72.3\% accuracy. However, perturbing the response distribution by modifying certain words with a random word leads to a drop in accuracy to 65.69\%. 
Lin \emph{et al.}~\citep{lin2024limitedgeneralization} further find that implicit reward models trained with DPO generalize even more poorly than explicit RMs under responses generated by a different policy model, with an up to 7\% accuracy drop.

\textbf{Question OOD} emerges due to variations in question difficulty or prompt formulation within the same domain. 
For example, RMs trained on intermediate-level math problems may fail to generalize to more challenging questions. 
Specifically, while PRMs trained on GSM8K or MATH datasets perform well on relatively simple problems, their accuracy diminishes for more complex questions (e.g., AIME-level tasks)~\citep{zheng2024processbench}. 
Nevertheless, Sun \emph{et al.}~\citep{sun2024easytohardgeneralization} show that RMs trained on easier tasks can still facilitate policy model improvement on harder tasks through test-time scaling or RL-based training. Their experiments demonstrate that weighted voting with easy-task-trained PRMs can improve performance on harder tasks by 8–10\% over majority voting baselines. Additionally, PPO training with rewards from easy-task PRMs (34.0\% accuracy) surpasses both full-dataset SFT (31.4\%) and previous RL state-of-the-art (33.0\%). 
Beyond question difficulty, the different formats of questions can also trigger OOD issues. As shown in the experiments by Levine \emph{et al.}~\citep{levine2024rmdistributionshift}, translating prompts to different languages reduces RM accuracy from 72.3\% to 70.29\%, even though the RM is based on multi-lingual pretraining data.
Huang \emph{et al.}~\citep{huang2025thinkbench} similarly observe significant degradation in the reasoning abilities of LLMs and PRMs when faced with novel patterns (e.g., adding new scenarios or adversarial attacks to math question descriptions). Their experimental data indicate average performance decays of 24.9\% and 11.8\% across all evaluated models on AIME-500 and AIME-24, respectively.

\textbf{Domain OOD} refers to generalization across different domains.
Lou \emph{et al.}~\citep{lou2025urm} manually construct randomly generated large‑number multiplications that are unlikely to have appeared in the RM’s HelpSteer2~\citep{wang2024helpsteer2} training data, and demonstrate the resulting uncertainty and inaccuracy in RM predictions.
Their experiments reveal that RM score distributions for OOD data have substantially greater variance, while scores for in-distribution (ID) data are more deterministic. 
Additionally, Zeng \emph{et al.}~\citep{zeng2025versaprm} (see 
Table~\ref{tab:versaprmresult}
in Appendix) show that although a PRM trained on math problems can generalize to closely related domains such as physics and chemistry, its performance declines significantly in unrelated areas such as psychology and history.

\subsubsection{Approaches to Enhance the Generalization of RMs}

A variety of approaches have been proposed to enhance the generalization capability of RMs. Some methods primarily focus on enhancing training data. For instance, Xia \emph{et al.}~\citep{xia2025agentrm} demonstrate that fine-tuning RMs on domain-specific data and using them to guide policy model test-time search 
can improve performance on held-out agent tasks. Similarly, Zeng \emph{et al.}~\citep{zeng2025versaprm} synthesize multi-domain data (e.g., math, law, philosophy, biology) to fine-tune PRMs, resulting in improved generalization across these areas. Wang \emph{et al.}~\citep{wang2025worldpm} further scale up the preference dataset to 15 million examples to train a base RM, achieving better generalization across tasks. 
However, most of the data-centric approaches can only boost the generalization of RMs to distributional shifts considered when building training data, which are not general solutions to the generalization issue of RMs.
One possible way to fundamentally solve this problem is by developing generalist RMs that are designed for broad applicability across diverse domains and tasks. For example, Liu \emph{et al.}~\citep{liu2025deepseekgrm} introduce SPCT, which trains generative RMs via online RL to dynamically self-generate domain-adaptive principles for various inputs. Yu \emph{et al.}~\citep{yu2025rewardanything}, in contrast, train an RM that can better adapt to human-specified principles and generalize across tasks without the need for task-specific retraining.

\subsection{Do the Discriminative Capabilities of LLMs Improve with their Generative Performance?}
\label{sec:q3improve}

\newcolumntype{C}{>{\centering\arraybackslash}X}
\renewcommand{\tabularxcolumn}[1]{m{#1}}
\renewcommand\cellalign{cc}

\begin{table}[t]
\centering
\caption{Co‑evolution of generative and discriminative abilities. Generation performance is measured as the mean accuracy over 32 model outputs for those models without publicly available benchmarks. Discrimination performance is evaluated by prompting each model to act as an ORM on a set of math and coding responses, with judgment accuracy reported. See the 
Appendix B
for experimental details.}
\label{tab:performance}
\footnotesize
\setlength{\tabcolsep}{4pt}       
\renewcommand{\arraystretch}{1.08} 
\begin{tabularx}{\textwidth}{l C C C C C C !{\hspace{6pt}} C C C}
\toprule
\multirow{3}{*}{Model}
  & \multicolumn{4}{c}{Generation Ability}
  & \multicolumn{5}{c}{Discrimination Ability} \\
\cmidrule(lr){2-5} \cmidrule(lr){6-10}
  & \multicolumn{2}{c}{Math}
  & \multicolumn{1}{c}{Coding}
  & \multicolumn{1}{c}{Avg.}
  & \multicolumn{4}{c}{Math}
  & \multicolumn{1}{c}{Coding} \\
\cmidrule(lr){2-3} \cmidrule(lr){4-4} \cmidrule(lr){5-5} \cmidrule(lr){6-9} \cmidrule(lr){10-10}
  & \makecell{AIME\\24}
  & \makecell{AIME\\25}
  & \makecell{Live\\Code\\Bench}
  & Avg.
  & \makecell{MATH\\500}
  & \makecell{Olympiad\\Bench}
  & \makecell{Omni\\Bench}
  & Avg.
  & \makecell{Live\\Code\\Bench} \\
\midrule
\multicolumn{10}{c}{\textit{Long CoT model}} \\
\cmidrule(rl){1-10}
o3                      & 91.6 & 88.9 & 75.8 & 85.4 & 96.5 & 84.5 & 79.6 & 86.9 & 92.9 \\
Gemini2.5 Pro           & 92.0 & 88.0 & 71.8 & 83.9 & 92.0 & 79.7 & 73.9 & 81.9 & 86.2 \\
Qwen3-8B (Thinking)     & 76.7 & 65.5 & 55.1 & 65.8 & 90.7 & 67.0 & 55.2 & 71.0 & 81.3 \\
R1-Distill-Llama-70B    & 70.0 & 56.2 & 55.5 & 60.6 & 89.5 & 65.7 & 58.0 & 71.1 & 79.5 \\
R1-Distill-Qwen-14B     & 69.7 & 50.7 & 51.5 & 57.3 & 91.4 & 69.2 & 60.5 & 73.7 & 78.9 \\
R1-Distill-Qwen-1.5B    & 28.9 & 22.9 & 19.8 & 23.9 & 59.7 & 57.7 & 52.5 & 56.6 & 59.2 \\
\cmidrule(rl){1-10}
\multicolumn{10}{c}{\textit{Short CoT model}} \\
\cmidrule(rl){1-10}
Deepseek-V3(0324)       & 59.4 & 49.5 & 27.2 & 45.4 & 89.8 & 69.0 & 62.1 & 73.6 & 73.6 \\
Qwen3-8B (Non-thinking) & 26.5 & 23.3 & 25.1 & 25.0 & 87.3 & 64.1 & 56.8 & 69.4 & 69.5 \\
GPT-4o                  & 13.1 & 11.3 & 29.5 & 18.0 & 88.9 & 70.1 & 66.8 & 75.3 & 69.8 \\
Llama3.1-8B-Instruct    & 5.4  & 0.4  & 12.6 &  6.1 & 73.5 & 54.7 & 51.8 & 60.0 & 65.9 \\
\bottomrule
\end{tabularx}
\end{table}

Different from discriminative RMs, generative RMs can leverage the broader knowledge and reasoning abilities of their underlying LLMs, and are generally more adaptable to new or unfamiliar OOD scenarios. 
When prompted as RMs, the generative abilities of such models can be leveraged for evaluation purposes. 
Feng \emph{et al.}~\citep{feng2025prmnecessary} show that RL training not only enhances the reasoning and problem-solving skills of LMs but also their capabilities as PRMs, suggesting a strong correlation between these abilities. 
Table~\ref{tab:performance} shows the results of our experiment comparing the generative and discriminative ability of multiple open-source and proprietary LLMs.
We find that, for both long-CoT and short-CoT models, there is a strong correlation between their generative and discriminative performance, i.e., LLMs with higher proficiency in solving math or coding problems are also better at identifying errors in reasoning within these domains. 
However, there is an exception for short-CoT models, GPT‑4o, whose discriminative performance is unexpectedly strong. We suspect it may be a result of its specialized distribution of training data or additional training on discriminative tasks.
Thus, improving the generative abilities of LLMs is likely to yield concurrent improvements in their discriminative reward modeling abilities. 
However, further progress in generation capabilities heavily relies on the performance of reward models for data generation and online RL.
Consequently, progress in generation and discrimination is likely to proceed in alternating phases: stronger reward models enable better generative training, which in turn yields models that serve as stronger discriminators.

\subsection{Do RM Evaluations Reflect Real-World Performance?}
\label{sec:q4rmevaluation}

Robust evaluation methods are crucial for selecting RMs and guiding their further improvement. However, most existing benchmarks focus on isolated aspects of RM performance, which may not be directly related to their performance on downstream tasks. In the following, we review representative evaluation metrics for RMs and discuss their relationship with real performance in several primary applications.

\subsubsection{Review of Representative Evaluation Metrics for RMs}

For the ease of comparison, the most widely used metrics are summarized and categorized as follows.

\begin{itemize}
    \item \textbf{Pairwise evaluation}. Given a manually labeled pair of LLM-generated responses, RMs should be able to identify the one preferred by human labelers. 
    Their accuracy in doing so is widely used as a metric to evaluate ORMs~\citep{lambert2024rewardbench, liu2024rmbench, zhou2025rmb}.
    \item \textbf{Correctness evaluation}. When we have correctness labels for responses or even steps in a response, we can directly compare the ground-truth correctness with RMs' predictions~\citep{song2025prmbench, zheng2024processbench, tan2025aurora}. 
    For example, in ProcessBench~\citep{zheng2024processbench}, each test case consists of a step-by-step solution with expert-annotated error positions. The performance of PRMs to identify the first error then serves as another metric for RMs.
\end{itemize}

While the above most commonly used metrics assess an RM's ability to distinguish between high- and low-quality outputs, we can also directly evaluate the effect of RMs on downstream tasks.
\begin{itemize}
    \item \textbf{BoN score} is the final-answer accuracy when using an RM to select the best response from multiple responses to a question generated by the same policy model. 
    In practice, multiple policy models are used to reduce the variance introduced by the choice of the policy model~\citep{wang2024mathshepherd, zhang2025prmlessons}. 
    Other BoN methods~\citep{zhou2025rmb, liu2025acemath} use a fixed evaluation dataset, which do not require a policy model, 
    \item \textbf{Search-guiding score} measures the final-answer correctness of responses, generated by a fixed policy model, whose search process is guided by the RM model~\citep{zhou2025jetts}. However, this evaluation method can be very costly for complex search algorithms.
\end{itemize}

There are also \textbf{integrated metrics}, which combine the above approaches for comprehensive evaluation. To assess alignment with human preferences, one can compare accuracy, ranking consistency, and uncertainty~\citep{yu2025rewardanything}. For reward correctness, both accuracy and BoN scores can be used. Direct evaluation of RLHF outcomes is often infeasible due to computational costs; thus, Frick \emph{et al.}~\citep{frick2024ppe} have compiled proxy metrics that relate well to actual RLHF performance.

\subsubsection{Relationship Between Evaluation Metrics and Downstream Task Performance}

\begin{figure*}[t]
    \centering
    \includegraphics[width=1.0\textwidth]{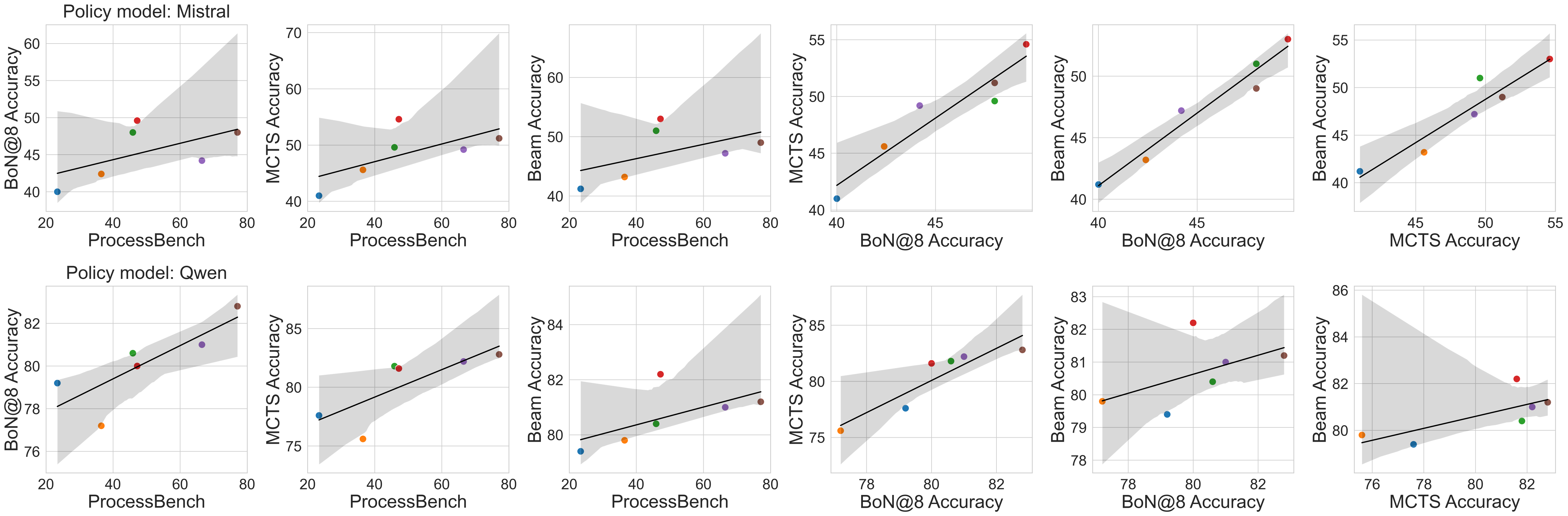}
    \caption{The relationship between correctness scores~(ProcessBench), BoN scores, and search-guiding performance~(MCTS and Beam) for different PRMs when used with two different policy models (math-shepherd-mistral-7b-rl~\citep{wang2024mathshepherd} and Qwen2.5-7B-Instruct~\citep{qwen2.5}) on MATH500. 
    Points in different colors denote the six PRM variants: \textcolor{Blue}{Math‑Shepherd‑PRM‑7B}~\citep{wang2024mathshepherd}, \textcolor{Orange}{Llama3.1‑8B‑PRM‑Mistral‑Data}~\citep{xiong2024rlhflowmath}, \textcolor{Green}{Skywork‑PRM‑1.5B}~\citep{skyworkopeno12024}, \textcolor{Red}{Skywork‑PRM‑7B}~\citep{skyworkopeno12024}, \textcolor{Purple}{Qwen2.5‑Math‑7B‑PRM800K}~\citep{zheng2024processbench}, and \textcolor{Brown}{Qwen2.5‑Math‑PRM‑7B}~\citep{zhang2025prmlessons}. The trend lines represent the fitted linear regression, and the shaded areas represent the 95\% confidence intervals.
    }
    \label{fig:PRM_correlation_ours.png}
\end{figure*}

We find that the most commonly used metrics of RMs, especially correctness evaluations, may not be enough to predict their performance on the two tasks of test-time guidance and online RL.

For \textbf{test-time guidance}, the primary goal is for the RM to select correct solutions from a batch of candidates, which is directly assessed by the BoN score. 
However, recent studies reveal that commonly used pairwise and correctness-based metrics may correlate poorly with the test-time performance of RMs.
For instance, Zhou \emph{et al.}~\citep{zhou2025rmb} find that pairwise selection benchmarks such as RewardBench~\citep{lambert2024rewardbench} exhibit weak or no correlation with BoN scores (Spearman’s $\rho$ ranging from -0.4 to 0.4), while the RMB benchmark correlates with BoN scores moderately (with $\rho$ from 0 to 0.7). 
For PRMs, Zhang \emph{et al.}~\citep{zhang2025prmlessons} show that correctness scores on ProcessBench do not consistently increase with BoN scores (with $\rho \approx 0.52$), and this relationship depends on specific PRM training details. 
Additionally, the BoN evaluation itself has limitations for PRMs, as it assesses only the correctness of the final answer and disregards the correctness of intermediate steps. 
Therefore, both outcome-level and process-level metrics are needed for a more comprehensive evaluation of PRMs during inference. 
Test-time methods such as MCTS and beam search require process-level supervision by PRMs to guide policy models towards correct answers. 
Consequently, their MCTS and beam search scores can more accurately reflect their discriminative ability on the step level.

To systematically evaluate the relationship between the PRMs' correctness scores, BoN scores, and searching-guiding performance of PRMs, we performed the experiment shown in Figure~\ref{fig:PRM_correlation_ours.png}. 
We can observe a positive correlation between PRMs' test-time performance and their correctness scores. However, correctness scores alone are not enough to predict the relative test-time performance of PRMs. 
For example, Skywork-PRM-7B only achieves moderate scores in ProcessBench with both policy models. However, it ranks first in 4 out of 6 downstream tasks. 
Consequently, we need to directly evaluate the test-time performance of PRMs, rather than relying solely on their correctness evaluations.
We also find that the relative test-time performance of PRMs varies across different policy models.

Correctness scores are also insufficient for the comprehensive evaluation of \textbf{online RL}. Chen \emph{et al.}~\citep{chen2024accuracyparadox} report that a moderately accurate RM can sometimes train a better LM than a more accurate RM. 
Similarly, Wen \emph{et al.}~\citep{wen2025rethinkingrewardmodelevaluation} observed that RMs with similar levels of correctness scores can lead to policies with markedly different performances after RL. 
Razin \emph{et al.}~\citep{razin2025makesrewardmodelgood} further find that, in RLHF settings, reward variance (the dispersion of scores a reward model assigns to outputs sampled from the current policy) can play a crucial role in determining the speed and effectiveness of RL training, i.e. an RM with higher accuracy but low reward variance does not necessarily achieve better optimization results.

\subsection{What are the Practical Deployment Challenges of RMs?}
\label{sec:practical_deployment}
Although RM can be applied to enhance the reasoning capabilities of LLMs, it comes with certain deployment challenges.
The deployment overhead of using RM can be divided into four parts: (1) the cost of collecting data for training RM; (2) the cost of training or iteratively updating RM separately; (3) the additional inference costs when using the trained RM; (4) maintenance costs that persist after deployment.

\begin{table}[t]
\centering
\footnotesize
\setlength{\tabcolsep}{3pt}
\renewcommand{\arraystretch}{1.22}
\caption{Summary of practical overheads introduced by reward models.
}
\label{tab:rm-overhead}
\begin{tabularx}{\columnwidth}{@{}
>{\raggedright\arraybackslash}p{0.20\columnwidth}
>{\raggedright\arraybackslash}p{0.48\columnwidth}
>{\raggedright\arraybackslash}X
@{}}
\toprule
\textbf{Overhead source} &
\textbf{Representative cost} &
\textbf{Main scaling factors} \\
\midrule

Data Construction &
\(
\begin{aligned}[t]
\mathcal{C}_{\mathrm{data}}
&\approx
\mathcal{C}_{\mathrm{sample}}
+
\mathcal{C}_{\mathrm{judge}} \\
\mathcal{C}_{\mathrm{sample}}
&\approx
QK\,\mathcal{C}_{\mathrm{gen}}(P_{\pi},L_x,L_y) \\
\mathcal{C}_{\mathrm{tree}}
&\approx
QKT\,\mathcal{C}_{\mathrm{gen}}
\bigl(P_{\pi},L_x+\bar{L}_{v},\Delta\bar{L}\bigr) \\
\mathcal{C}_{\mathrm{judge}}
&\approx
|\mathcal{D}_{\mathrm{raw}}|\,
\mathcal{C}_{\mathrm{fwd}}(P_j,L_j)
\end{aligned}
\) &
$Q$: prompts;
$K$: candidates per prompt;
$T$: tree nodes per prompt;
$|\mathcal{D}_{\mathrm{raw}}|$: raw data size;
$P_{\pi},P_j$: model sizes;
$L_x,L_y,L_j$: input lengths. \\

\midrule

RM training or update &
\(
\begin{aligned}[t]
\mathcal{C}_{\mathrm{train}}^{\mathrm{full}}
&\approx
E T_r\,\mathcal{C}_{\mathrm{train}}(P_r) \\
\mathcal{C}_{\mathrm{train}}^{\mathrm{RM\text{-}RL}}
&\approx
\mathcal{C}_{\mathrm{rollout}}
+
\mathcal{C}_{\mathrm{reward}}
+
\mathcal{C}_{\mathrm{opt}}
\end{aligned}
\) &
$E$: epochs;
$T_r$: RM tokens;
$P_r$: RM size;
$P_{\Delta}$: trainable parameters;
rollouts and optimization. \\

\midrule

Inference-time scoring &
\(
\begin{aligned}[t]
\mathcal{C}_{\mathrm{ORM}}
&\approx
\mathcal{C}_{\mathrm{fwd}}(P_r,L_x+L_y) \\
\mathcal{C}_{\mathrm{PRM}}^{\mathrm{single}}
&\approx
\mathcal{C}_{\mathrm{fwd}}(P_r,L_x+L_y) \\
\mathcal{C}_{\mathrm{PRM}}^{\mathrm{prefix}}
&\approx
\sum_{s=1}^{S}
\mathcal{C}_{\mathrm{fwd}}(P_r,L_x+L_{1:s}) \\
\mathcal{C}_{\mathrm{BoN}}
&\approx
N\,\mathcal{C}_{\mathrm{gen}}
+
N\,\mathcal{C}_{\mathrm{score}} \\
\mathcal{C}_{\mathrm{search}}
&\approx
|\mathcal{V}|\,\mathcal{C}_{\mathrm{score}}
\end{aligned}
\) &
$S$: reasoning steps;
$N$: candidates;
$L_{1:s}$: prefix length;
$|\mathcal{V}|$: scored nodes;
$P_r$: RM size. \\

\midrule

Post-deployment maintenance &
\(
\begin{aligned}[t]
\mathcal{C}_{\mathrm{maint}}
&\approx
\mathcal{C}_{\mathrm{serve}}
+
\mathcal{C}_{\mathrm{monitor}} \\
&\quad+
\mathcal{C}_{\mathrm{refresh}}
+
\mathcal{C}_{\mathrm{update}}
\end{aligned}
\) &
RM weights;
KV cache;
batching;
endpoint latency;
distribution shift;
re-annotation;
validation. \\

\bottomrule
\end{tabularx}
\end{table}

\subsubsection{Theoretical Overhead of RM Training and Inference}
\label{sec:rm_theoretical_overhead}

We first generally analyze these costs from four aspects: RM training data collection, RM training, inference-time scoring, and long-term deployment maintenance. 
Table \ref{tab:rm-overhead} summarizes these overhead. Let 
$P_{\pi}$ and $P_{r}$ denote the parameter counts of the policy model and reward model, respectively. Let $L_x$ be the prompt length, $L_y$ the response length, $S$ the number of reasoning steps in a response, and $N$ the number of candidate responses sampled for each prompt. For simplicity, we use
\[
    \mathcal{C}_{\mathrm{fwd}}(P,L)
\]
to denote the cost of one forward pass of a model with $P$ parameters over a sequence of length $L$. Up to hardware-dependent constants, this cost can scale with $P$ and $L$.

\textbf{Data collection cost.}
Training an RM first requires collecting trajectories, sub-trajectories, or preference pairs. If $Q$ prompts are used and $K$ candidate trajectories are sampled per prompt, the basic sampling cost is
\[
    \mathcal{C}_{\mathrm{sample}}
    \approx
    QK \cdot \mathcal{C}_{\mathrm{gen}}(P_{\pi}, L_x, L_y),
\]
where $\mathcal{C}_{\mathrm{gen}}$ denotes autoregressive generation cost. Tree-based data collection further amplifies this cost: if a search tree expands $T$ nodes per prompt, and each expansion conditions on an average prefix length $\bar{L}_{v}$ and generates an average continuation length $\Delta\bar{L}$, then
\[
\mathcal{C}_{\mathrm{tree}}
\approx
QKT\,\mathcal{C}_{\mathrm{gen}}
\bigl(P_{\pi},L_x+\bar{L}_{v},\Delta\bar{L}\bigr).
\]
Thus, the sampling cost also scales with the number of expanded nodes $T$.
After sampling, additional filtering or annotation is usually needed. If an external judge model $M_j$ is used to label each collected example, with parameter count $P_j$ and average judge input length $L_j$, then
\[
    \mathcal{C}_{\mathrm{judge}}
    \approx
    |\mathcal{D}_{\mathrm{raw}}| \cdot \mathcal{C}_{\mathrm{fwd}}(P_j,L_j),
\]
or higher if the judge generates natural-language critiques. Therefore, for many PRMs and generative RMs, the total data construction cost may dominate the final RM fine-tuning cost.

\textbf{RM training cost.}
Given an RM training set $\mathcal{D}_r$ with total token count $T_r$ and $E$ training epochs, the approximate full fine-tuning cost is
\[
    \mathcal{C}_{\mathrm{train}}^{\mathrm{full}}
    \approx
    E T_r \cdot \mathcal{C}_{\mathrm{train}}(P_r),
\]
where $\mathcal{C}_{\mathrm{train}}(P_r)$ includes forward, backward, and optimizer update costs, and can scale with $P_r$. Full fine-tuning is usually the most expensive setting because gradients and optimizer states must be maintained for all parameters.

Parameter-efficient training changes this cost profile. For LoRA or adapter-based tuning, only a small number of trainable parameters $P_{\Delta} \ll P_r$ are updated, reducing optimizer memory and communication overhead. However, the forward and backward passes still go through the frozen backbone, so the compute cost does not shrink in proportion to $P_{\Delta}$.
If the RM itself is trained with RL, as in some generative or reasoning-oriented RMs, the cost profile changes again. In that case, training no longer consists only of supervised updates on a fixed dataset. It also includes rollout generation, reward estimation, and repeated optimization over sampled trajectories:
\[
\mathcal{C}_{\mathrm{train}}^{\mathrm{RM\text{-}RL}}
\approx
\mathcal{C}_{\mathrm{rollout}}
+
\mathcal{C}_{\mathrm{reward}}
+
\mathcal{C}_{\mathrm{opt}}.
\]
Thus, RM training with reinforcement learning can be much closer in cost to policy RL training than to ordinary supervised reward-model fine-tuning.

\textbf{Inference-time scoring cost.}
At inference time, ORMs and PRMs introduce different scoring overheads. A discriminative ORM usually assigns one scalar score to a complete response with a single forward pass:
\[
    \mathcal{C}_{\mathrm{ORM}}
    \approx
    \mathcal{C}_{\mathrm{fwd}}(P_r, L_x + L_y).
\]
For PRMs, the cost depends on the scoring implementation. Some PRMs, such as those following the token-level or step-token design in prior work~\citep{lightman2023letsverifystepstep, zhang2025prmlessons}, can obtain all step-level scores in one forward pass over the full solution:
\[
    \mathcal{C}_{\mathrm{PRM}}^{\mathrm{single}}
    \approx
    \mathcal{C}_{\mathrm{fwd}}(P_r, L_x + L_y).
\]
In contrast, if the PRM or judge evaluates each prefix separately, then the cost becomes
\[
    \mathcal{C}_{\mathrm{PRM}}^{\mathrm{prefix}}
    \approx
    \sum_{s=1}^{S}
    \mathcal{C}_{\mathrm{fwd}}(P_r, L_x + L_{1:s}),
\]
where $L_{1:s}$ is the token length of the first $s$ reasoning steps. When steps have comparable lengths, this can grow approximately quadratically with the response length due to repeated computation over overlapping prefixes.

For best-of-$N$ selection, the total cost is
\[
    \mathcal{C}_{\mathrm{BoN}}
    \approx
    N \cdot \mathcal{C}_{\mathrm{gen}}(P_{\pi}, L_x, L_y)
    +
    N \cdot \mathcal{C}_{\mathrm{score}},
\]
where $\mathcal{C}_{\mathrm{score}}$ is the cost of ORM, PRM, or generative RM scoring. Thus, small discriminative RMs usually add a manageable marginal cost. However, when $P_r$ is close to $P_{\pi}$, or when long chain-of-thought responses must be scored, reward evaluation can consume a large fraction of the total inference budget.

The overhead becomes more pronounced in search-guided reasoning. If beam search has width $B$ and depth $D$, reward scoring is invoked on approximately $BD$ partial candidates. If MCTS performs $U$ simulations with average expansion size $A$, the number of reward evaluations scales with the number of visited nodes, roughly $UA$. Hence,
\[
    \mathcal{C}_{\mathrm{search}}
    \approx
    |\mathcal{V}| \cdot \mathcal{C}_{\mathrm{score}},
\]
where $|\mathcal{V}|$ is the number of evaluated nodes. In online RL, if each update samples $Q$ prompts, $K$ rollouts per prompt, and runs for $I$ iterations, the reward computation cost is
\[
    \mathcal{C}_{\mathrm{RL\text{-}reward}}
    \approx
    I QK \cdot \mathcal{C}_{\mathrm{score}}.
\]
This cost is in addition to policy rollout generation, reference-model evaluation, value-model computation if used, and policy optimization.

\textbf{Maintenance cost after deployment.}
Finally, deployed RMs introduce persistent system costs. Serving an RM requires additional GPU memory for model weights and KV cache, extra batching and scheduling logic, and often a separate low-latency inference endpoint. If the RM is used together with the policy model during online RL or test-time search, the system must coordinate two or more models, which may reduce throughput even when the RM itself is small.

Moreover, in some frameworks or implementations, RMs are not static components. As the policy model improves, the response distribution shifts, and the RM may become miscalibrated or vulnerable to reward hacking. Maintaining RM reliability therefore requires periodic data refreshment, re-annotation, retraining, calibration, and downstream validation. If the RM is continuously updated together with the policy model, the total deployment cost includes not only the initial RM training cost, but also the recurring cost of monitoring distribution shift and rebuilding the reward pipeline. This makes RM deployment a long-term systems problem rather than a one-time model training problem.

\subsubsection{Practical Training and Inference Overhead}

Next, we conduct a analysis of the training and inference costs of RM in practical scenarios to quantitatively present specific numerical values.

\begin{figure*}[t]
    \centering
    \includegraphics[width=1.0\textwidth]{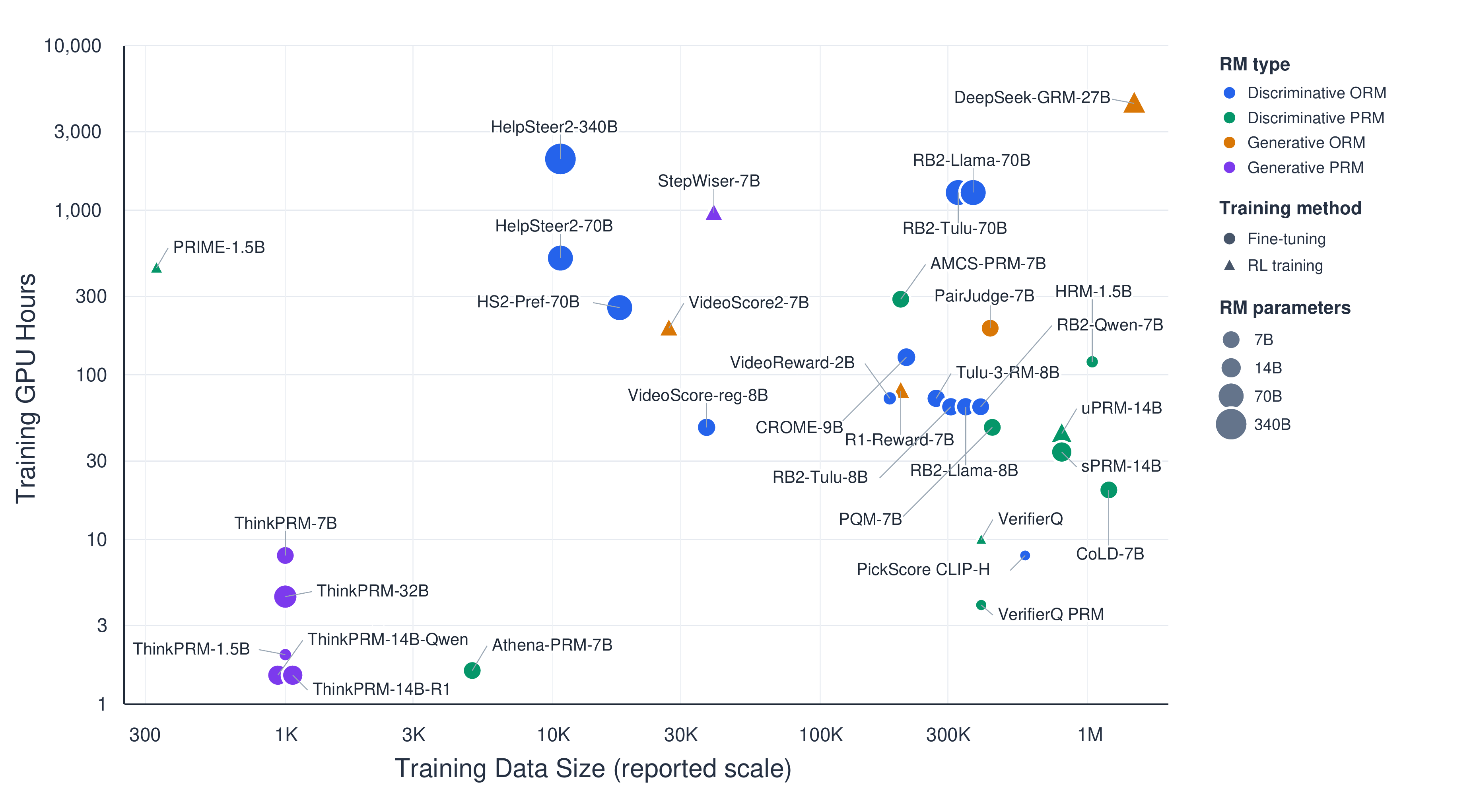}
    \caption{Summary of some reported RM training GPU hours by model and training data size from existing works.}
    \label{fig:rm_training_gpu_hours}
\end{figure*}

\begin{table}[t]
\centering
\caption{Reported RM data construction time or cost from existing works.}
\label{tab:rm-data-construction-cost}
\small
\setlength{\tabcolsep}{3pt}
\renewcommand{\arraystretch}{1.08}
\begin{tabularx}{\columnwidth}{@{}p{0.34\columnwidth}p{0.25\columnwidth}X@{}}
\toprule
RM & Hardware & Reported total RM training-data time/cost \\
\midrule
Athena-PRM\citep{wang2025athena} & 8 MI-250 GPUs & 10 h; MC baseline: 20 d \\
StepWiser\citep{xiong2025stepwiser} & 8 A100 GPUs & 14 d \\
FLAMe\citep{vu2024flame} & Not reported & $\sim$13--17 d (extrap.) \\
HRM\citep{wang2025hrm} & A100 GPUs & 2,457 A100 GPU-h \\
PPM\citep{guan2025rstarmath} & 80 H100 + 60 A100 GPUs & $\sim$27 d total \\
MCTS-DPO verifier\citep{xie2024mctsdpo} & A100 GPUs & 30 A100 GPU-d; 2 min/sample \\
Full-Step-DPO PRM\citep{xu2025fullstepdpo} & A100 GPUs & 8.5--317 A100 GPU-h \\
Reward-SQL PRM\citep{zhang2025rewardsql} & 4 A800 GPUs & 18 h; PRM training: 28 GPU-h \\
VersaPRM\citep{zeng2025versaprm} & AWS Bedrock & $<\$100$; time not reported \\
\bottomrule
\end{tabularx}
\vspace{2pt}
\end{table}

\textbf{Practical RM training cost.}
To characterize the computational cost of training existing RMs, we collect reported GPU-hours from some related works, and plot them together with the reported training data size in Figure \ref{fig:rm_training_gpu_hours}. It is important to note that cost reporting in RM papers remains incomplete. Among the related works we surveyed, only a subset reports both the hardware count and training time. Many papers report only the data size, hardware type, number of epochs or steps, or the overall post-training cost, without separately reporting the RM training cost. Therefore, Figure \ref{fig:rm_training_gpu_hours} should be viewed as a descriptive summary of reported-cost samples, rather than an unbiased estimate of the entire RM literature.

Among the samples with reported costs, RM training cost varies widely, from a few GPU-hours to several thousand GPU-hours. Low-cost cases usually correspond to small training sets or parameter-efficient fine-tuning, such as ThinkPRM \citep{khalifa2025thinkprm} models trained with LoRA or fine-tuning on a small number of verification chains. High-cost cases are more common in large-scale general-purpose RMs or models that include an RL training stage, such as DeepSeek-GRM \citep{liu2025deepseekgrm}, StepWiser \citep{xiong2025stepwiser}, and PRIME\citep{cui2025prime}. This large variation arises because RM training cost is determined not only by model size and number of training samples, but also by the fraction of trainable parameters, sequence length, the number of candidate responses to compare or score per prompt, and whether online rollouts are required. Compared with static supervised fine-tuning, RL-style RM training often requires repeated generation, scoring, filtering, and updating, which can lead to higher GPU-hours even when the model size is not the largest.

Overall, GPU-hours in the figure tend to increase with model size and training data size, but the relationship is not strictly monotonic and is strongly affected by the training paradigm. For models with similar parameter sizes, simple supervised fine-tuning, parameter-efficient fine-tuning, and RL-based training can differ by one or even multiple orders of magnitude. In addition, multimodal RMs such as Athena \citep{wang2025athena}, VideoScore \citep{he2025videoscore2} and R1-Reward \citep{zhang2025r1reward} do not show substantially higher reported training GPU-hours than text-only RMs of similar scale. This may be because existing papers often count only the RM fine-tuning stage, while excluding multimodal data construction, annotation, visual feature processing, or synthetic data generation. A more cautious conclusion is therefore that, under the currently limited cost disclosure, multimodality itself has not yet appeared as the dominant factor in RM training cost, but its full end-to-end cost still requires more transparent reporting.

For PRMs, however, a major part of the cost is often shifted to the construction of process-level training data. As summarized in Table \ref{tab:rm-data-construction-cost}, the reported data construction cost can range from several hours to weeks, and in some cases reaches thousands of A100 GPU-hours. These numbers are not directly comparable across papers because they use different hardware and data scales, but they show a consistent pattern: for PRM-style methods, data generation and annotation can be a dominant hidden cost, sometimes comparable to or larger than the subsequent RM training itself. For example, in HRM \citep{wang2025hrm}, MCTS-based data generation takes about 2,457 A100 GPU-hours, whereas training the HRM itself takes only about 120 A100 GPU-hours, a gap of roughly 20 times.

\textbf{Practical RM Inference Cost.} Table \ref{tab:drm_grm_cost} summarizes reported inference costs of discriminative and generative reward models in recent studies. Although different works report cost using different units, comparisons within each work show a similar trend: generative reward models usually have much higher inference cost than discriminative reward models. In relatively controlled settings, generative reward models can already be about five times slower than scalar discriminative models. In more complex reasoning settings, this gap becomes much larger, because the reward model needs to generate hundreds or even thousands of natural-language tokens before the final judgment can be obtained.

This cost difference matters in practice because reward models are usually not called only once. In Best-of-N selection, each sampled candidate needs to be scored. In beam search or tree search, the reward model may be used to score many intermediate nodes. In online RL, reward computation is repeated across prompts, rollouts, and training iterations. Therefore, even a moderate cost for a single RM call can become a major bottleneck when reward scoring is used in large-scale inference or training. The practical cost of an RM should therefore be measured not only by the latency of a single scoring call, but also by the number of times it is called by the downstream algorithm.

These results indicate a clear trade-off between cost and performance. Generative RMs may have stronger reasoning ability, better interpretability, and more robust judgments, but their higher latency can make them less suitable for high-throughput or low-latency deployment. In contrast, discriminative RMs are often more suitable when the reward model needs to be called many times, such as in large-scale filtering, BoN reranking, or search-guided reasoning. One practical solution is to use a cascaded verification pipeline: a lightweight discriminative RM first filters or ranks most candidates, and a more expensive generative RM is only used for uncertain, high-stakes, or top-ranked cases. Therefore, future RM evaluations should report not only accuracy, but also latency, generated tokens, throughput and memory footprint.

\begin{table}[t]
\centering
\small
\setlength{\tabcolsep}{3.5pt}
\renewcommand{\arraystretch}{1.15}
\caption{Inference cost comparison between discriminative reward models (DRMs) and generative reward models (GRMs) from existing works.}
\label{tab:drm_grm_cost}
\begin{tabularx}{\columnwidth}{@{}
>{\raggedright\arraybackslash}X
>{\raggedright\arraybackslash}p{0.20\columnwidth}
>{\raggedright\arraybackslash}p{0.22\columnwidth}
>{\raggedleft\arraybackslash}p{0.16\columnwidth}
@{}}
\toprule
\textbf{Work} & \textbf{DRM cost} & \textbf{GRM cost} & \textbf{GRM/DRM} \\
\midrule
Structural Reward Model, public dataset \citep{liu2025structuralrm}
& 18.7 s / 1k samples
& 92.5 s / 1k samples
& 4.95$\times$ \\

Structural Reward Model, industrial dataset \citep{liu2025structuralrm}
& 21.3 s / 1k samples
& 106.1 s / 1k samples
& 4.98$\times$ \\

RLBFF / Flexible Principles RM \citep{wang2026rlbff}
& 1 generated-token-equivalent; $<0.1$ s / task
& thousands of tokens; $>10$ s / task
& $>100\times$ \\

CAMEL vs. RM-R1, RewardBench \citep{zhu2026camel}
& 1 output token
& $\sim$900 output tokens
& $\sim$900$\times$ \\

CAMEL vs. RM-R1, RM-Bench \citep{zhu2026camel}
& 1 output token
& $\sim$1,100 output tokens
& $\sim$1,100$\times$ \\
\bottomrule
\end{tabularx}
\end{table}
\section{Future Directions, Limitations, and Recommendations}
\label{sec:future-limitations-recommendations}

In this section, we discuss future research directions, scope limitations of this
survey, and practical recommendations for developing and using RMs for LLM
reasoning. The discussion complements the analysis in Section \ref{sec:analysis} and focuses on
issues that are likely to become more important as RMs are used in more complex deployment settings.

\subsection{Future Directions}

We discuss several directions for future RM research in this section. It focuses on
problems that may become more important when RMs are used with stronger reasoning
models and more complex systems.

\paragraph{Adaptive reward systems.}
Most existing RMs are trained once and then used as fixed scorers for reranking,
search, data filtering, or online RL. This design is simple and effective in many
settings. However, when a policy model is repeatedly optimized with an RM, the
distribution of its responses changes. The RM may then need to judge harder
solutions or responses that are adapted to its own scoring pattern. Therefore, a
promising direction is to move from static RMs to adaptive reward systems. In such
systems, the policy model, RM, verifier, data generator, and evaluation protocol are
updated and monitored together. Recent work on reasoning RL also emphasizes that
reward design affects what a model learns, how it generalizes, and whether its
outputs can be trusted. This suggests that reward modeling should be studied not
only as a prediction problem, but also as a long-term training and deployment
problem.

\paragraph{Generalist and criteria-aware RMs.}
Another important direction is to build RMs that can work across different tasks,
domains, and evaluation criteria. A useful generalist RM should not only assign a
score, but also understand what criteria should be used for the current input. For
example, open-ended reasoning tasks may require different balances among
correctness, faithfulness, usefulness, safety, and conciseness. Recent rubric-based
generative RMs, such as DeepSeek-GRM and RM-R1
\citep{liu2025deepseekgrm,chen2025rmr1}, move in this direction by generating
principles, rubrics, or critiques before making a judgment. This makes the reward
process more transparent and easier to inspect. At the same time, generated criteria
can be incomplete or misleading. Future work should therefore evaluate both the
candidate answer and the criteria used to judge it, including whether the criteria
cover the task requirement, remain stable under prompt changes, and avoid rewarding
irrelevant properties.

\paragraph{Task-aligned RM evaluation.}
Current RM evaluation often relies on pairwise preference accuracy, correctness
accuracy, or benchmark-level rankings. These metrics are useful, but they may not
predict downstream performance in Best-of-N selection, tree search, data curation, or
online RL. Going forward, RMs should be evaluated in the same
mode in which they are used. For example, an RM for reranking should be tested
with multiple candidate budgets and policy models, while an RM for RL should be
tested through actual policy improvement. Evaluation should also report calibration,
reward variance, uncertainty, latency, generated tokens, and performance under
distribution shifts in task, language, response style, and model family.

\paragraph{State-aware rewards for agents and multimodal reasoning.}
Many current RMs are developed for math and code, where final answers can often be
verified. Future reasoning systems will increasingly involve tool use, multimodal
inputs, long-horizon interaction, and agentic decision making. In these settings, the
quality of an action may depend on the current state, previous actions, available
tools, external evidence, and whether mistakes can be corrected later.
Future RMs for such settings should evaluate state
transitions rather than only textual reasoning traces. They should also use
environment logs and tool feedback when these signals are available.

\paragraph{Agent-as-a-judge systems.}
Most current RMs are not based on agent systems, however, many future reasoning tasks can be complex and will need tool
use, multi-turn interaction, and feedback from external environments for judgement. In these
settings, the judge may need to act more like an agent: it should inspect the
trajectory, call tools when verification is needed, track the state of the interaction,
and aggregate evidence before giving a reward. Recent work has started to explore
this direction \citep{xu2026tirjudge, zhang2025sage, xi2025agentprm}. Future RMs may therefore evolve from passive scorers into
agentic judges that combine textual critiques, tool feedback, environment logs, and
task-specific rubrics.

\subsection{Limitations}

This survey has several scope limitations. First, we focus on RMs that enhance LLM
reasoning, especially their use in test-time guidance, synthetic data curation, and
online RL. As a result, we do not provide a complete review of general preference
alignment, RLHF, DPO-style preference optimization, or constitutional AI. These
topics are discussed only when they are directly related to reward modeling for
reasoning. Similarly, we discuss LLM-as-a-judge and automatic evaluation mainly
from the perspective of generative RMs and verifier models, rather than covering all
judge applications in dialogue, writing, factuality checking, or safety evaluation.

Second, our taxonomy is mainly organized around text-based reasoning tasks, such as
math, code, and structured problem solving. We include multimodal, tool-use,
agentic, and domain-specific RMs when they are closely connected to reasoning, but
we do not exhaustively cover specialized reward models for robotics, embodied agents,
long-horizon web tasks, scientific discovery, medical decision support, or real-world
policy settings. These areas often require rewards over states, actions, external
evidence, or domain constraints, and they deserve more detailed treatment in future
surveys.

Third, our empirical comparisons are intended to support the main observations of
this survey, rather than to rank all existing RMs. We focus on representative models,
widely used tasks, and publicly reported results. Some closed-source systems,
proprietary training pipelines, and very recent preprints cannot be analyzed with the
same level of detail. Since reward modeling is developing rapidly, the taxonomy in
this paper should be viewed as a structured snapshot of the current literature. Future
work can extend it by adding newly released generalist RMs, agentic reward systems,
multimodal process benchmarks, and more standardized cost reports.

\subsection{Recommendations}

This subsection gives recommendations for choosing, evaluating, and using RMs. The
goal is to make the discussion above easier to apply in research and deployment.

\paragraph{Choose the RM according to the target use case.}
Researchers and practitioners should first decide how the RM will be used. For
large-scale filtering or online RL, pointwise discriminative RMs are often more
efficient because their scores can be reused and normalized. For difficult reranking
or open-ended evaluation, pairwise or generative RMs can be more reliable because
they compare candidates or produce explicit critiques. For multi-step reasoning,
PRMs are useful when intermediate errors matter, while ORMs may be enough when
the final answer can be reliably verified. Therefore, RM selection should be based on
the downstream algorithm, latency budget, annotation cost, and required level of
interpretability.

\paragraph{Use layered reward pipelines.}
For practical deployment, it is often better to use a layered reward pipeline than to
rely on a single RM. Cheap verifiable checks and lightweight discriminative RMs can
handle easy cases, while stronger generative critics or human experts can be reserved
for uncertain, high-value, or safety-sensitive examples. This design is consistent with
the cost-performance trade-off discussed in this survey and can reduce unnecessary
use of expensive reward models.

\paragraph{Use PRMs carefully in online RL.}
PRMs are useful for diagnosing reasoning errors, guiding search, filtering data, and
improving credit assignment. However, when they are used as direct RL rewards, their
signals should usually be bounded, normalized, or combined with outcome-level
verification. In particular, PRMs should be tested under policy optimization, because
a reward model that works for static evaluation may behave differently after the
policy has been trained to maximize its scores.

\paragraph{Evaluate RMs according to their target use case.}
An RM used for Best-of-N selection should be evaluated with Best-of-N curves under
different candidate budgets and policy models. An RM used for search should be
evaluated inside the corresponding search algorithm. An RM used for online RL
should be evaluated by the quality of the resulting policy, not only by static reward
accuracy. For generalist RMs, evaluation should also include OOD domains, prompt
formats, languages, response styles, and model families.

\paragraph{Monitor distribution shift and reward hacking.}
After deployment or during online RL, the policy model will gradually generate
responses that differ from the RM training data. This distribution shift can make RM
scores miscalibrated or easier to exploit. Engineering systems should therefore track
score distributions, rejection rates, disagreement among verifiers, and examples with
unusually high reward but weak independent validation. Researchers should also evaluate
whether a policy trained with an RM improves under independent tests, not only under
the same RM used for training.

\paragraph{Document the reward specification.}
Rubrics, principles, verifier rules, prompts, aggregation methods, thresholds, and
normalization schemes are all part of the reward function. They should be documented
and versioned together with the RM. For generative RMs, the generated principles or
rubrics should also be logged when possible, because they explain why a reward was
assigned and make later auditing easier.

\paragraph{Report cost and maintenance information.}
Future RM papers should report not only benchmark scores, but also the cost of the
reward pipeline. Useful information includes data construction time, judge or verifier
calls, training GPU-hours, inference throughput, generated critique tokens, memory
footprint, and retraining frequency. Such reporting is important because an RM that
performs well in an offline benchmark may be too expensive for MCTS, large-N
reranking, or online RL.

\paragraph{Add uncertainty and deferral mechanisms.}
In many cases, the best behavior for an RM is
not to give a confident score, but to indicate uncertainty or defer to a stronger
verifier. Future generalist RMs should therefore provide uncertainty estimates,
missing-evidence signals, or failure conditions when possible. In engineering
pipelines, such signals can be used to trigger additional sampling, a stronger judge,
a tool-based verifier, or human review.

Overall, future RM research should focus on reward models that remain useful under
optimization pressure, distribution shift, and deployment constraints. Accuracy on a
static benchmark is important, but it is not sufficient. The most useful RMs will be
those that are accurate, calibrated, robust, cost-effective, and well matched to the
way they are used in the reasoning pipeline.

\section{Conclusion}

In this work, we present a systematic survey of reward models for enhancing the reasoning abilities of LLMs. We first review the main types of reward models, including outcome and process reward models, discriminative and generative reward models, and pointwise and pairwise reward formulations. We then summarize representative benchmarks for evaluating these models across text-only and multimodal settings. Based on this foundation, we examine how reward models are used in three major stages of the reasoning pipeline: test-time guidance, synthetic data curation and self-iteration, and online reinforcement learning.

Our analysis shows that reward models have become an important component for improving LLM reasoning, but their effectiveness strongly depends on how they are designed, evaluated, and deployed. PRMs can provide fine-grained feedback and are often useful for search, reranking, and data filtering, but they require more costly process-level data and may introduce new reward-hacking risks in RL. Generative RMs can offer more interpretable and often more robust judgments, especially in difficult or out-of-distribution settings, but their higher inference cost limits their use in high-throughput scenarios. Meanwhile, common static evaluation metrics do not always reflect downstream performance, suggesting that RMs should be evaluated in the same setting in which they are used.

Looking forward, several directions are especially important. First, future RMs should become more general and criteria-aware, so that they can adapt to different tasks, domains, and evaluation standards without task-specific retraining. Second, RM evaluation should move beyond isolated accuracy and include downstream utility, robustness, calibration, uncertainty, and cost. Third, practical RM systems should be designed as layered and adaptive pipelines, combining verifiable checks, lightweight reward models, stronger generative judges, and human review when necessary. Finally, as LLMs move toward long-horizon, multimodal, and agentic reasoning, reward models should also evolve from passive scorers into state-aware and tool-aware judges.

Overall, reward models provide a powerful way to guide, improve, and evaluate LLM reasoning. However, building reliable RMs requires more than higher benchmark scores. The most useful RMs should be accurate, robust, interpretable, cost-effective, and well matched to their downstream use. We hope this survey can provide a clear foundation for future research and support the development of more reliable reasoning systems.

\backmatter

\noindent

\begin{appendices}

\section{Results in Related works}

\begin{table}[t]
\centering
\caption{As a representative of generative RMs, GenPRM-7B \citep{zhao2025genprm} can show strong accuracy and outperform contemporary discriminative RMs on ProcessBench.}
\label{tab:dis_gen_processbench}
\begin{tabularx}{\textwidth}{@{}Xcccc c@{}}
\toprule
Models                               & GSM8K     & MATH      & Olympiad Bench & Omni-MATH & Avg.  \\
\midrule
\addlinespace[0.5em]
\multicolumn{6}{c}{\emph{Discriminative Reward Models}} \\
\midrule
Math-Shepherd-PRM-7B           & 47.9 & 29.5 & 24.8 & 23.8 & 31.5 \\
Skywork-PRM-7B                 & 70.8 & 53.6 & 22.9 & 21.0 & 42.1 \\
Qwen2.5-Math-7B-Math-Shepherd  & 62.5 & 31.6 & 13.7 &  7.7 & 28.9 \\
Qwen2.5-Math-7B-PRM800K        & 68.2 & 62.6 & 50.7 & 44.3 & 56.5 \\
Qwen2.5-Math-PRM-7B            & 82.4 & 77.6 & 67.5 & 66.3 & 73.5 \\
Universal-PRM-7B               & \textbf{85.8} & \textbf{77.7} & \textbf{67.6} & \textbf{66.4} & \textbf{74.3} \\
\midrule
\addlinespace[0.5em]
\multicolumn{6}{c}{\emph{Generative Reward Models}} \\
\midrule
Direct Generative PRM-7B       & 63.9 & 65.8 & 54.5 & 55.9 & 60.0 \\
GenPRM-7B (Pass@1)             & 78.7 & 80.3 & 72.2 & 69.8 & 75.2 \\
GenPRM-7B (Maj@8)              & \textbf{81.0} & \textbf{85.7} & \textbf{78.4} & \textbf{76.8} & \textbf{80.5} \\
\bottomrule
\end{tabularx}
\end{table}
\begin{table*}[t]
  \centering
  \small
  \caption{Performance of popular open-source ORMs and PRMs on BoN tasks using three distinct policy models. Results are taken from \citep{yuan2024implicitprm}.}
  \label{tab:implictprmresults}
  \resizebox{\textwidth}{!}{%
  \begin{tabular}{llcccccccccc}
    \toprule
    Type & Reward Model
         & \multicolumn{3}{c}{\shortstack{Mistral-7B-Inst-v0.2\\Pass@1: 9.6}}
         & \multicolumn{3}{c}{\shortstack{Llama-3.1-8B-Inst\\Pass@1: 44.6}}
         & \multicolumn{3}{c}{\shortstack{Llama-3.1-70B-Inst\\Pass@1: 63.2}}
         & Avg. \\
    \cmidrule(lr){3-5} \cmidrule(lr){6-8} \cmidrule(lr){9-11}
    & & @4 & @16 & @64 & @4 & @16 & @64 & @4 & @16 & @64 & \\
    \midrule
    \multirow{3}{*}{ORM}
      & EurusRM-7B              & 17.2 & 21.0 & 20.4 & 49.6 & 51.6 & 51.8 & 69.0 & 69.6 & 72.2 & 46.9 \\
      & SkyworkRM-Llama3.1-8B   & 16.0 & 19.6 & 23.4 & 49.0 & 50.4 & 48.2 & 70.4 & 72.6 & 72.0 & 46.8 \\
      & ArmoRM-Llama3-8B        & 16.6 & 21.0 & 23.2 & 47.8 & 48.6 & 49.4 & 70.6 & 70.8 & 71.0 & 46.6 \\
    \midrule
    \multirow{4}{*}{PRM}
      & Math-Shepherd-7B        & 16.0 & 21.0 & 20.4 & 50.0 & 52.4 & 52.8 & 66.4 & 65.8 & 65.6 & 45.6 \\
      & RLHFlow-8B-Mistral-Data & 19.4 & 25.2 & 30.2 & 51.8 & 52.0 & 50.6 & 70.8 & 71.0 & 71.2 & 49.1 \\
      & RLHFlow-8B-DS-Data      & 17.2 & 23.0 & 25.2 & 54.4 & 54.2 & 55.8 & 68.6 & 70.4 & 73.0 & 49.1 \\
      & ImplicitPRM (DPO)       & 18.6 & 24.4 & 28.8 & 54.0 & 55.4 & 57.0 & 71.8 & 71.2 & 72.2 & 50.4 \\
    \bottomrule
  \end{tabular}%
  }
\end{table*}
\begin{table}[t]
    \centering
    \small
    \caption{Weighted majority-voting (WMV) accuracy of two open-source PRMs evaluated on various domains in VersaPRM dataset using Llama-3.1-8B-Instruct as the policy model. The improvements over majority voting is reported. Results are taken from \citep{zeng2025versaprm}.}
    \label{tab:versaprmresult}
    \begin{tabular*}{\textwidth}{@{\extracolsep{\fill}}lccc@{}}
        \hline
        \textbf{Domain} &
        \makecell[c]{\textbf{Majority}\\\textbf{Voting}} &
        \makecell[c]{\textbf{Math-Shepherd-PRM}\\\textbf{(WMV)}} &
        \makecell[c]{\textbf{Qwen-2.5-Math-PRM}\\\textbf{(WMV)}} \\
        \hline
        Math        & 62.40 & 64.13 (+1.73)   & 67.20 (+4.80) \\
        \hline
        Chemistry   & 58.67 & 60.13 (+1.46)   & 60.67 (+2.00) \\
        Physics     & 58.53 & 61.87 (+3.34)   & 61.47 (+2.94) \\
        \hline
        Biology     & 75.38 & 75.38 (+0.00)   & 75.69 (+0.31) \\
        Psychology  & 61.60 & 61.47 ($-0.13$) & 62.27 (+0.67) \\
        Law         & 35.93 & 37.24 (+1.31)   & 36.28 (+0.35) \\
        History     & 49.20 & 49.87 (+0.67)   & 49.40 (+0.20) \\
        Philosophy  & 44.83 & 44.70 ($-0.13$) & 45.17 (+0.34) \\
        \hline
    \end{tabular*}
\end{table}

\begin{table*}[t]
  \centering
  \small
  \caption{The accuracy of PRMs on ProcessBench and downstream test-time tasks on MATH500}
  \label{tab:prmperformance}
  \resizebox{\textwidth}{!}{%
  \begin{tabular}{@{} l c c c c c c c @{}}
    \toprule
    \multirow{2}{*}{PRM} 
      & \multirow{2}{*}{ProcessBench-MATH500} 
      & \multicolumn{3}{c}{Mistral Base} 
      & \multicolumn{3}{c}{Qwen Base} \\
    \cmidrule(lr){3-5} \cmidrule(lr){6-8}
      &  & BoN@8 & MCTS & Beam 
         & BoN@8 & MCTS & Beam \\
    \midrule
    Math-Shepherd-PRM-7B               & 23.4 & 40.0 & 41.0 & 41.2 & 79.2 & 77.6 & 79.4 \\
    Llama3.1-8B-PRM-Mistral-Data       & 36.5 & 42.4 & 45.6 & 43.2 & 77.2 & 75.6 & 79.8 \\
    Skywork-PRM-Qwen2.5-1.5B           & 45.9 & 48.0 & 49.6 & 51.0 & 80.6 & 81.8 & 80.4 \\
    Skywork-PRM-Qwen2.5-7B             & 47.2 & 49.6 & 54.2 & 53.2 & 80.0 & 81.6 & 82.2 \\
    Qwen2.5-Math-7B-PRM800K            & 66.5 & 44.2 & 49.6 & 47.0 & 81.0 & 82.2 & 81.0 \\
    Qwen2.5-Math-PRM-7B                & 77.1 & 48.0 & 51.2 & 49.0 & 82.8 & 82.8 & 81.2 \\
    \bottomrule
  \end{tabular}%
  }
\end{table*}

\begin{table}[htbp]
  \centering
  \footnotesize
  \caption{100 generated responses accuracy for each model on each dataset}
  \label{tab:generationaccuracy}
  \begin{tabular}{lcccc}
    \toprule
    \multicolumn{1}{l}{} 
        & \multicolumn{3}{c}{Math} & Coding \\
    \cmidrule(lr){2-4} \cmidrule(lr){5-5}
    Model 
      & MATH500 
      & OlympiadBench 
      & OmniBench 
      & LiveCodeBench \\
    \midrule
    o3                        & 94 & 72 & 71 & 65 \\
    Gemini2.5 Pro             & 91 & 81 & 71 & 70 \\
    R1-Distill-Llama-70B      & 92 & 62 & 51 & 60 \\
    R1-Distill-Qwen-14B       & 92 & 55 & 53 & 52 \\
    R1-Distill-Qwen-1.5B      & 77 & 47 & 29 & 22 \\
    GPT4o                     & 73 & 35 & 34 & 34 \\
    DeepseekV3(0324)          & 95 & 58 & 51 & 54 \\
    Llama3.1-8B-IT            & 49 & 19 &  9 & 20 \\
    Qwen3-8B(Thinking)        & 89 & 65 & 51 & 53 \\
    Qwen3-8B(Non-thinking)    & 82 & 50 & 40 & 33 \\
    \bottomrule
  \end{tabular}
\end{table}

The key results in related works are listed in Table \ref{tab:dis_gen_processbench}, Table \ref{tab:implictprmresults}, and Table \ref{tab:versaprmresult} for reference.

\section{Experimental Details}
\label{sec:experimentaldetails}

In our experiments examining the correlation between ProcessBench scores and downstream test-time search performance, as illustrated in Figure \ref{fig:PRM_correlation_ours.png}, we primarily evaluate six open-source PRMs: Math-Shepherd-PRM-7B, Llama3.1-8B-PRM-Mistral-Data, Skywork-PRM-Qwen2.5-1.5B, Skywork-PRM-Qwen2.5-7B, Qwen2.5-Math-7B-PRM800K, and Qwen2.5-Math-PRM-7B. The experiments are implemented based on the OpenR framework~\citep{wang2024openr}. For all evaluations, the generation temperature was consistently set to 0.7. Beam search utilized a beam size of 4, while MCTS employed 4 simulation paths. Detailed performance metrics for each PRM are presented in Table \ref{tab:prmperformance}.

\begin{figure}[t]
    \centering
    \includegraphics[width=0.8\textwidth]{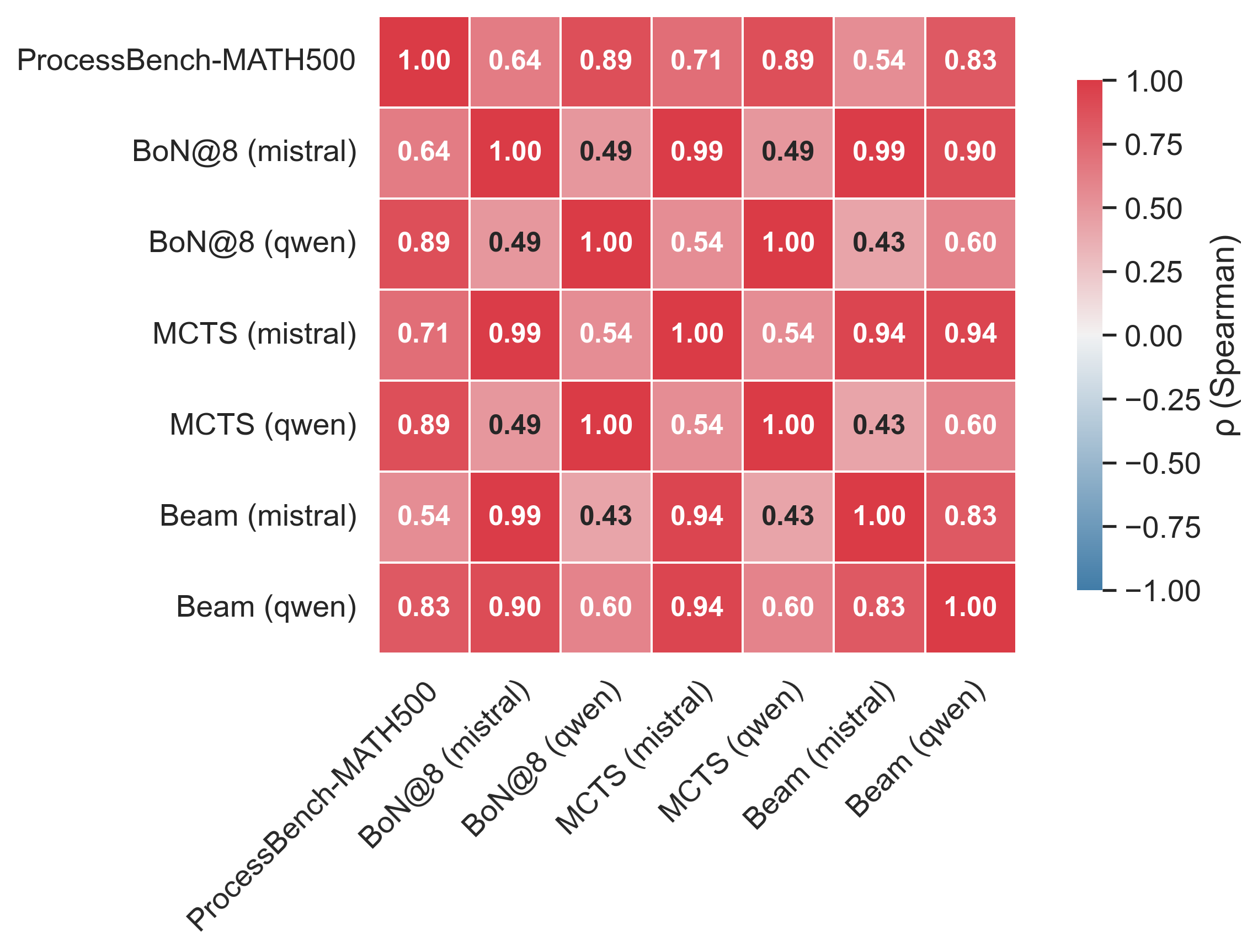}
    \caption{Heatmap illustrating Spearman correlation coefficients among the evaluated test-time search strategies and ProcessBench-MATH500 scores}
    \label{fig:prmcorrheatmap}
\end{figure}

\begin{figure*}[t]
    \centering
    \includegraphics[width=1.0\textwidth]{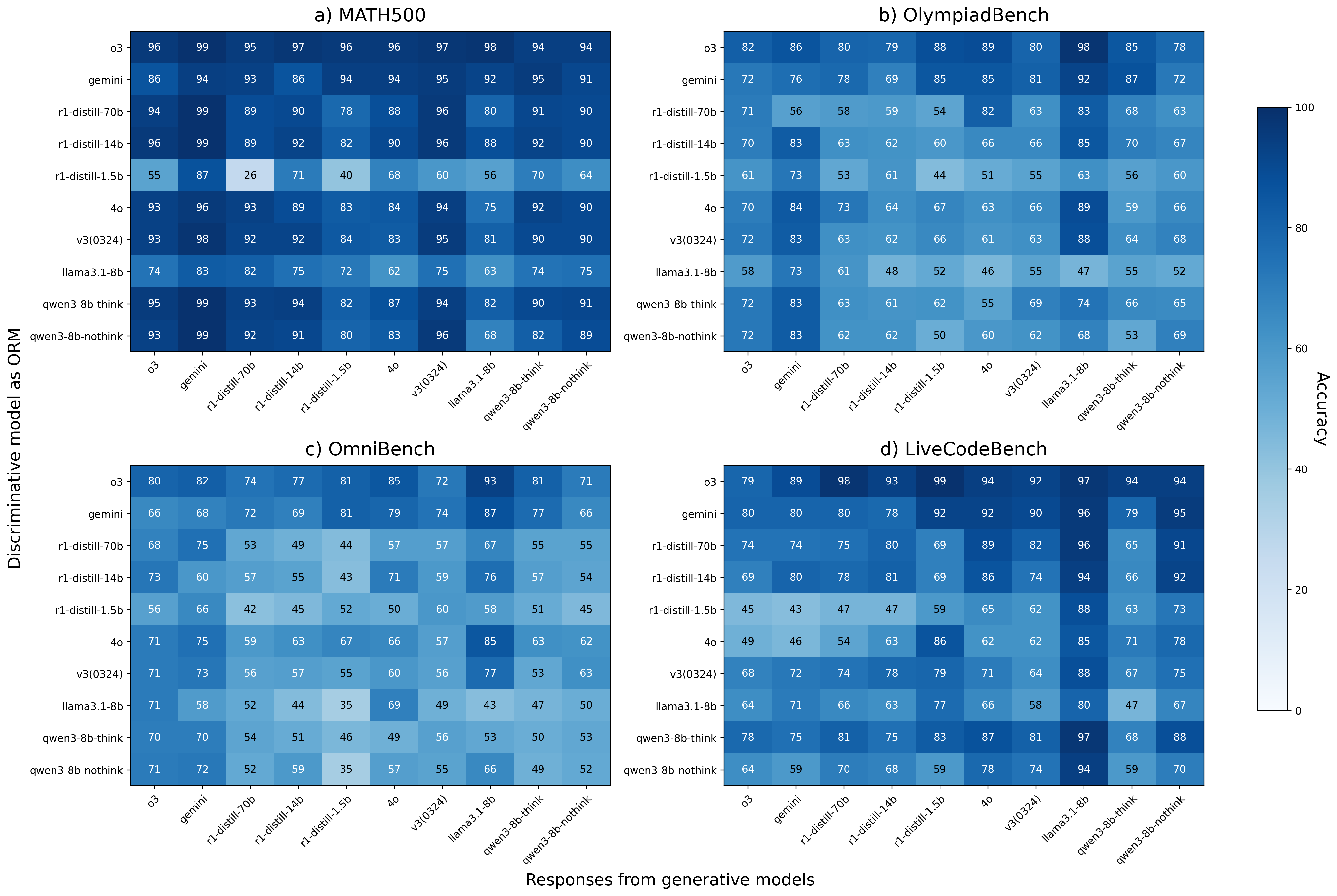}
    \caption{The discrimination accuracy for responses generated from different models}
    \label{fig:accuracyeachmodel}
\end{figure*}

For the assessment of generation and discrimination capabilities, we report performance metrics for six long CoT models and four short CoT models in Table \ref{tab:performance}. Generation scores for AIME and LiveCodeBench represent the average outcomes over 32 trials for R1-Distill-Llama-70B, R1-Distill-Qwen-14B, R1-Distill-Qwen-1.5B, Qwen3-8B, and Llama3.1-8B-IT. Other model scores are cited directly from officially reported results. For discrimination capability evaluations, we generated responses for 100 randomly sampled questions per dataset and per model. LiveCodeBench questions cover the period from August 1, 2024, to May 1, 2025, and were evenly distributed across easy, medium, and hard difficulty levels. The answer accuracy of each model's generated responses is summarized in Table \ref{tab:generationaccuracy}, and the discrimination accuracy across responses generated by different models is presented in Figure \ref{fig:accuracyeachmodel}.

\clearpage

\section{Prompts}
\label{sec:prompts}

The following prompts are used by LLM-as-a-judge to evaluate generated responses for math and coding tasks as an ORM.

\begin{mdframed}[linewidth=0.5pt]
The following is a math problem and a solution:

[Math Problem]

<problem description>

[Solution]

<solution here>

Your task is to determine if the final answer provided in the solution is **entirely correct** for the given problem. Disregard minor errors in steps as long as the final answer is mathematically correct.

If the solution leads to the correct final answer, output "Yes", otherwise output "No".

Please put your final verdict **only** (i.e., "Yes" or "No") in \texttt{\textbackslash boxed\{\{\}\}}.
\end{mdframed}

\begin{mdframed}[linewidth=0.5pt]
The following is a coding problem and a code solution:

[Coding Problem]

<problem description>

[Code Solution]

<solution here>

Your task is to review and evaluate the code solution. Determine if the solution is functionally correct and fully solves the problem requirements.

If the solution is entirely correct and solves the problem, output "Yes". If there are any critical errors that prevent it from functioning as required, output "No".

Please put your final verdict (i.e., "Yes" or "No") in \texttt{\textbackslash boxed\{\{\}\}}.
\end{mdframed}




\end{appendices}


\bibliography{main}

\end{document}